%% file: arxiv.tex
\pdfoutput=1

\documentclass[10pt,twocolumn,letterpaper]{article}

\usepackage[pagenumbers]{cvpr}

\input{preamble}
\usepackage{stfloats}

\definecolor{cvprblue}{rgb}{0.21,0.49,0.74}
\usepackage[pagebackref,breaklinks,colorlinks,allcolors=cvprblue]{hyperref}
\hypersetup{pdftitle={The Cost of Conversation: How Self-Generated History Impairs Video Understanding},pdfauthor={Ananth K Suresh, Yadynesh D Sonale, Anuj Lalla, Vineeth N Balasubramanian}}

\def\confName{CVPR}
\def\confYear{2027}

\title{The Cost of Conversation: How Self-Generated History Impairs Video Understanding}

\author{%
  Ananth K Suresh\qquad Yadynesh D Sonale\qquad Anuj Lalla\qquad Vineeth N Balasubramanian\\[0.45em]
  Microsoft Research, India\\[0.3em]
  {\small\ttfamily \{ksananth4424, yadyneshsonale, anujlalla11\}@gmail.com\qquad vineeth.nb@microsoft.com}%
}

\begin{document}
\addtocontents{toc}{\protect\setcounter{tocdepth}{-1}}
\maketitle
\input{fig/figure1_excalidraw_v5_bars_tex}
\input{sec/0_abstract}
\input{sec/1_intro}
\input{sec/2_related}
\input{sec/3_testbed}
\input{sec/4_drop}
\input{sec/5_history}
\input{sec/5_lookuse}
\input{sec/6_repair}
\input{sec/8_conclusion}
\input{sec/9_llm_usage}
{
    \small
    \bibliographystyle{ieeenat_fullname}
    \bibliography{main}
}

\addtocontents{toc}{\protect\setcounter{tocdepth}{2}}
\input{sec/X_suppl}

\end{document}

%% file: preamble.tex
\usepackage{cuted} %

\usepackage{currfile} %

\usepackage{caption} %

\usepackage{microtype}

\usepackage[T1]{fontenc} %
\usepackage{multirow}
\usepackage{array}
\usepackage{tikz}
\usetikzlibrary{arrows.meta,positioning,calc}
\usepackage{xspace}

\definecolor{condfull}{HTML}{5A35B8}
\definecolor{condconcat}{HTML}{0D7864}
\definecolor{condsharded}{HTML}{B85A00}
\definecolor{condstrip}{HTML}{1B5C9E}
\DeclareRobustCommand{\condtag}[3]{%
  \ifmmode
    \text{\textcolor{#2}{\normalfont\textsc{#3}}}%
  \else
    \begingroup\setlength{\fboxsep}{0.9pt}\colorbox{#1}{\vphantom{Ay}\textcolor{#2}{\normalfont\textsc{#3}}}\endgroup
  \fi
}
\newcommand{\full}{\condtag{plotpurple!14}{condfull}{full}\xspace}
\newcommand{\concat}{\condtag{plotteal!15}{condconcat}{concat}\xspace}
\newcommand{\sharded}{\condtag{plotamber!22}{condsharded}{sharded}\xspace}
\newcommand{\replystripped}{\condtag{plotblue!16}{condstrip}{strip}\xspace}
\newcommand{\noted}{``Noted.''\xspace}
\newcommand{\cleanturn}{CleanTurn\xspace}

\newcommand{\modelmark}[1]{%
  \makebox[8pt][c]{\raisebox{-0.7pt}{%
    \includegraphics[width=8pt,height=7pt,keepaspectratio]{fig/logos/#1.pdf}}}\hspace{2pt}}

\usepackage{colortbl}
\usetikzlibrary{patterns}
\definecolor{okblue}{HTML}{0072B2}
\definecolor{okverm}{HTML}{D55E00}
\definecolor{okgreen}{HTML}{009E73}
\definecolor{okpurple}{HTML}{CC79A7}
\definecolor{okorange}{HTML}{E69F00}
\definecolor{oksky}{HTML}{56B4E9}
\definecolor{okyellow}{HTML}{F0E442}
\definecolor{plotpurple}{HTML}{7851D5}
\definecolor{plotteal}{HTML}{20A58A}
\definecolor{plotblue}{HTML}{4A86C5}
\definecolor{plotamber}{HTML}{DFA415}
\newcommand{\pend}{\textcolor{black!35}{$\cdot$}}

%% file: fig/figure1_excalidraw_v5_bars_tex.tex
\begin{figure*}[t]
  \centering
  \includegraphics[trim=0 2bp 0 1bp,clip,width=1\textwidth]{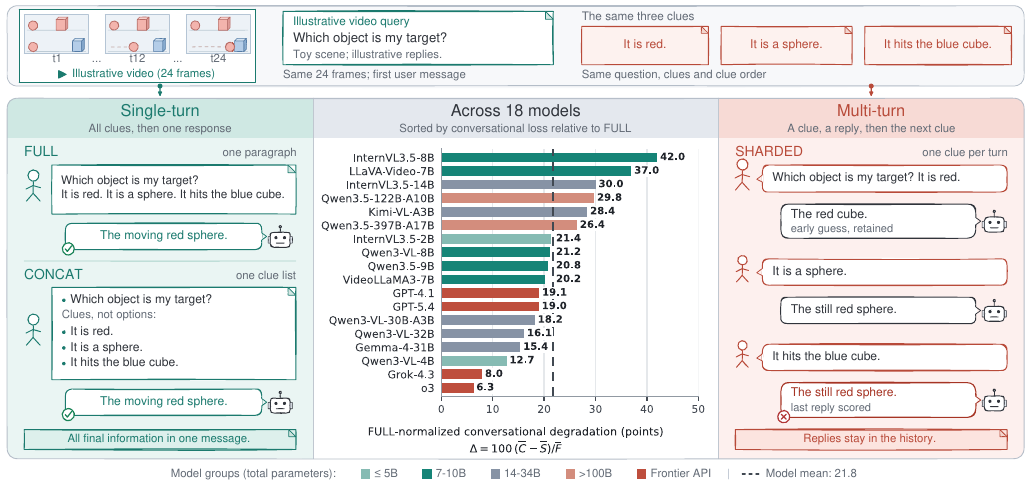}%
  \caption{\textbf{Same video and clues, different delivery.}
  Single-turn prose (\full, \emph{left, top}) and clue-list (\concat, \emph{left, bottom}) versus multi-turn \sharded (\emph{right}), flanking 18-model \full-normalized conversational degradation $\Delta$ (\emph{centre}).}
  \label{fig:teaser}%
  \label{fig:delivery_radar}%
\end{figure*}

%% file: sec/0_abstract.tex
\begin{abstract}
Video-language models often answer a question accurately when all clues arrive at once, yet fail when those same clues arrive across dialogue turns even though the video frames remain visible throughout the conversation.
Across 10,023 annotation-verified queries over 6,822 videos, five synthetic and real-world tasks, and 18 open-weight and proprietary models, single-turn clue lists (\concat) retain $97.0\%$ of single-turn prose (\full) accuracy whereas multi-turn delivery (\sharded) retains $75.2\%$, a \textbf{$\mathbf{21.8}$-percentage-point conversational drop} beyond list formatting.
Controlled history interventions show that retaining the model's own earlier replies accounts for a substantial share of this loss alongside turn-level factors such as user-turn fragmentation, while wrong early commitments actively bias subsequent answers.
Moreover, under equal-budget greying of annotated supporting versus non-evidence frames, retained replies dampen behavioural sensitivity to supporting visual evidence, whereas removing those replies restores visual sensitivity across all seven tested models.
Motivated by this diagnosis, omitting or position-masking prior assistant replies (\emph{Reply Masking}) provides a simple, zero-extra-pass primary intervention at inference time, while instance-specific \emph{Clean-State Steering} (\cleanturn) serves as a model-dependent extension when residual turn-fragmentation loss remains.
\end{abstract}

%% file: sec/1_intro.tex
\section{Introduction}
\label{sec:intro}

Given all clues to a video question in one prompt, a video-language model often answers accurately, yet fails when those same clues arrive across turns (\cref{fig:teaser}).
Unlike text-only chat, the video frames stay visible throughout the dialogue, so the model keeps the visual evidence needed to check or revise an early guess once the final clue arrives.
We ask: \textbf{when visual evidence stays in context, how does a model's own reply history affect video-QA accuracy and sensitivity to the video?}

Existing benchmarks leave this open.
Single-turn video QA presents the full query at once~\cite{clevrer2020,nextgqa2024,perceptiontest2023}, whereas video-dialogue benchmarks change questions and turn dependencies across turns~\cite{mtvideobench,svbench,dvd2021,sama}.
Text-only sharding studies have no persistent visual signal~\cite{laban2025lost,ownwords2026}, and multimodal snowballing work plants curated false histories or adversarial premises on static images~\cite{mmhalsnowball2024,mmsnowball2026} rather than delivering truthful clues on video.
In our four linked findings, we show that conversational delivery degrades video QA (\cref{sec:drop}), retained assistant replies contribute to this loss alongside turn factors (\cref{sec:cause}), those replies dampen behavioural dependence on supporting video frames (\cref{sec:lookuse}), and omitting or masking them provides a simple primary repair (\cref{sec:fix}).

We construct 10,023 queries across 6,822 videos spanning five object, timing and temporal-grounding tasks from CLEVRER~\cite{clevrer2020}, Charades~\cite{charades2016} and Perception Test~\cite{perceptiontest2023} (\cref{sec:testbed}).
Each query pairs 24 timestamped frames and an underspecified question with two to seven verbatim clues that identify one target under the source annotations.
Clues arrive in a fixed order as single-turn prose (\full), a single-turn list (\concat), or one clue per turn with replies kept in context (\sharded).
Across 18 open-weight and API models, \concat retains $97.0\%$ of five-task \full accuracy whereas \sharded retains $75.2\%$, a $21.8$-point conversational drop beyond list formatting (\cref{fig:delivery_radar}, centre; \cref{tab:drop}).
Every model drops, from $\Delta=6.3$ retention points on o3 to $42.0$ on InternVL3.5-8B, with wide variation across tasks (\cref{sec:drop}).

On a 3,218-query (2,942-video) subset, we separate prior-reply effects from the overall delivery drop (\cref{sec:cause,tab:history_scale}).
Retaining earlier assistant turns accounts for part of the loss, alongside final-turn query restatement (\emph{Recap}).
Deleting prior replies (\replystripped) gains $2.4$ to $12.1$ points across six models, and position-preserving \emph{Reply Masking} gains $2.2$ to $12.6$ across five open-weight models, ruling out rotary position shifts alone even though the active sequence still changes.
Separately, wrong prior replies actively bias later answers: planting a wrong guess cuts accuracy by $2.2$ to $28.9$ points, whereas neutral \noted replies help four of six models while collapsing Kimi-VL-A3B.

On 3,214 queries, greying annotated evidence frames rather than an equal number of non-evidence frames hurts \sharded far less than \full or \concat, whereas deleting prior replies (\replystripped) increases behavioural sensitivity across all seven tested models (\cref{sec:lookuse,fig:greying_summary,fig:greying_dose_by_task}).
Retained history thus makes final answers rely less on visual input, without implying that visual encoding itself fails.

Finally, we test how much loss \emph{Reply Masking} recovers alone and when adding \emph{Clean-State Steering} from a consolidated clean pass (\textbf{\cleanturn}) helps further (\cref{sec:fix,tab:repair_scale}).
Masking drives most of the gain on three of four open-weight models; steering adds $+8.2$ points beyond masking on LLaVA-Video-7B and $+1.9$ on Qwen3-VL-8B, with no resolved gain on InternVL3.5-8B or -14B, where \emph{Recap}, score mixing (\emph{Mix}) or direct \full answering match or beat \cleanturn.

Our contributions are threefold:
\begin{enumerate}
  \setlength{\itemsep}{0pt}
  \item \textbf{Benchmark:} 10,023 queries (6,822 videos, five tasks, 18 models) separating multi-turn video-QA loss from list formatting; we will release the query dataset and evaluation codebase upon acceptance.
  \item \textbf{Diagnostics:} Reply edits and matched frame greying separating retained-reply loss, wrong-answer bias and reduced dependence on supporting video frames.
  \item \textbf{Masking-first repair analysis:} Ablations of \emph{Reply Masking} as a zero-pass primary intervention and \cleanturn as a model-dependent extension against \emph{Recap}, \emph{Mix} and \full on four open-weight models.
\end{enumerate}

%% file: sec/2_related.tex
\section{Related Work}
\label{sec:related}

\paragraph{Progressive disclosure vs.\ video dialogue.}
Laban \etal~\cite{laban2025lost} show that sharding task constraints across turns degrades text LLMs relative to single-turn prose and list prompts, using an LLM user simulator and a final-turn recap baseline.
Multi-turn video benchmarks~\cite{dvd2021,svbench,sama,mtvideobench} instead vary questions and turn dependencies across turns, Layered-VQA~\cite{composition2026} splits static images across RGBA layers, and interactive grounding~\cite{seeing2026} studies adaptive clarification.
By holding the video intact, target and verbatim annotation-verified clues fixed across \full, \concat and \sharded, we isolate conversational loss without simulator paraphrasing while complete visual evidence remains in context.

\paragraph{Self-generated history and visual sensitivity.}
In text dialogue, early errors snowball~\cite{zhang2024snowball} and omitting prior assistant turns can improve accuracy~\cite{ownwords2026}; in static-image chat, prior studies induce snowballing via curated hallucinatory histories~\cite{mmhalsnowball2024} or adversarial user premises~\cite{mmsnowball2026}.
Separately, single-turn video-QA work shows that high accuracy need not imply reliance on temporal visual evidence~\cite{buch2022revisiting,nextgqa2024}.
However, text-only studies lack persistent sensory input, and image-snowballing benchmarks inject false context rather than truthful clues.
Neither establishes how a model's own replies under truthful clues alter sensitivity to supporting video frames, or separates position shifts (\replystripped) from position-preserving \emph{Reply Masking} and wrong-answer bias.

\paragraph{Inference-time mitigation and steering.}
Prior mitigations distill history-cleaned policies (MAIGO~\cite{maigo2026}) or adjust multimodal decoding via distorted-image logit contrast (VCD~\cite{leng2024vcd}), residual visual decoding (RVD~\cite{mmhalsnowball2024}) or representation and logit rectification (CAVR~\cite{mmsnowball2026}).
In activation steering, CAA and function vectors average behavior or task directions across examples~\cite{rimsky2024caa,todd2024functionvectors}.
We instead evaluate position-preserving \emph{Reply Masking} as our primary history intervention and test when a per-query clean-state residual difference (\cleanturn) adds further recovery against masking, \emph{Recap}, clean-context score mixing (\emph{Mix}) and direct \full answering, without claiming superiority over the training or decoding mitigations above.

%% file: sec/3_testbed.tex
\section{Same Video, Different Route}
\label{sec:testbed}

To separate conversational effects from task difficulty or wording drift, we build annotation-verified video queries whose clues arrive either all at once or turn by turn (\cref{fig:teaser,fig:conditions}).
Standard open-ended video-QA and video-chat benchmarks rarely support this isolation: single-fact queries cannot be split into independent necessary clues, multi-turn chats change sub-questions across turns, and open-world scenes lack closed candidate sets for verification (Supp.~A).

\subsection{Annotation-verified clue sets}
\label{sec:requests}

\begin{table}[t!]
  \caption{Benchmark tasks and video coverage. Each task draws one query per video; cross-task overlap yields at most two queries per video across 6,822 unique videos (2,942 in the diagnostic subset).}
  \label{tab:tasks}
  \centering\scriptsize\setlength{\tabcolsep}{4pt}
  \begin{tabular}{@{}lrrcl@{}}
    \toprule
    Task & Pool & \shortstack{Diagnostic\\subset} & Clues & Answer \\
    \midrule
    CLEVRER object & 2,859 & 858 & 7 & object \\
    CLEVRER timing & 3,811 & 1,144 & 7 & timestamp \\
    Charades object & 1,738 & 522 & 2--3 & class \\
    Charades grounding & 1,317 & 396 & 7 & span \\
    Perception Test object & 298 & 298 & 2--4 & object \\
    \midrule
    Total queries & 10,023 & 3,218 & & \\
    Unique videos & 6,822 & 2,942 & & \\
    \bottomrule
  \end{tabular}
\end{table}

Each query pairs an underspecified question (\eg, \emph{``Which object in this video am I thinking of?''}) with two to seven true clues (\eg, \emph{``It collides with something blue.''}) that jointly select one target in the scene's annotation-defined candidate set (\cref{tab:tasks}; Supp.~A).
Each task draws at most one query per video, giving 10,023 queries across 6,822 videos (4,358 CLEVRER~\cite{clevrer2020}, 2,166 Charades~\cite{charades2016} and 298 Perception Test~\cite{perceptiontest2023}; at most two queries per video across tasks).
To exclude ambiguous targets and label noise, CLEVRER uses debounced visible objects and unique event anchors; Charades restricts to verified videos ($\le 40$\,s), actions $\ge 1.0$\,s, unique anchors, $1.0$\,s ordering margins, target spans $\le 25\%$ of duration, and merges $\text{IoU}\ge 0.5$ overlaps; Perception Test applies three-valued logic over tracked boxes and action slots, dropping unreliable or $>1.6$\,s-gapped tracks and role-ambiguous templates, plus an independent verifier.
After uniform 24-frame sampling (step $\le 1.67$\,s on Charades and $\le 1.46$\,s on Perception Test), $100\%$ of the 10,023 queries retain at least one sampled target-evidence frame (mean $4.7\text{ to }9.5$ frames per task), and $94.1\%$ ($9{,}432/10{,}023$; $99.9\text{--}100\%$ CLEVRER, $94.0\%$ Perception Test, $91.4\%$ Charades grounding and $73.7\%$ Charades object) are solved from those 24 frames by at least one single-turn model ($88.7\%$ by at least two models under \full; Supp.~A).
Clues arrive verbatim on a fixed schedule independent of model replies, isolating delivery from adaptive clarification or topic shifts.

\subsection{Delivery routes and evaluation protocol}
\label{sec:conditions}

\input{fig/sharded_simulation_excalidraw_v2_tex}

Every query places the same 24 chronological, timestamped frames in the first user message and holds one seeded random clue order fixed across three routes (\cref{fig:teaser}):
\begin{itemize}
  \setlength{\itemsep}{1pt}
  \item \full presents the question and clues in one paragraph.
  \item \concat presents them as one bulleted list with a note that lines are clues rather than answer options, isolating list formatting from turn-taking.
  \item \sharded reveals one clue per turn with free assistant replies kept in context; only the final reply is scored (\cref{fig:sharded_sim}).
\end{itemize}
Prompts request answers in task format without forbidding clarification; intermediate non-answers are unpenalized.

\input{fig/figure2_excalidraw_v3_tex}

\paragraph{Models.}
We test 18 models across Qwen3-VL~\cite{qwen3vl2025}, Qwen3.5~\cite{qwen35}, InternVL3.5~\cite{internvl35_2025}, Gemma~4~\cite{gemma4}, Kimi-VL~\cite{kimivl2025}, LLaVA-Video~\cite{llavavideo2024} and VideoLLaMA3~\cite{videollama3_2025}, plus GPT-4.1~\cite{gpt41}, GPT-5.4~\cite{gpt54}, o3~\cite{o3} and Grok-4.3~\cite{grok43} (\cref{tab:drop}).
All models use 24 frames ($\le 640$\,px; low-detail API images), temperature $1.0$ (omitted on o3) and a $1{,}000$-token cap ($16{,}384$ for o3 and Grok-4.3); pinned Hugging Face commits, API snapshots (Nov 2026) and rerun stability ($\le 0.9$\,pp shift across local and API reruns) appear in Supp.~C--D.

\paragraph{Scoring and statistics.}
Diagnostics use a fixed 3,218-query (2,942-video) subset: a seeded 30\% sample of each larger task plus all 298 Perception Test queries; repair settings are tuned on case-disjoint queries (video-overlap check in Supp.~I).
All replies use GPT-4.1 extraction~\cite{gpt41} ($0.04\text{ to }0.10\%$ extraction errors; $99\%$ binary-score agreement on a 100-reply stratified manual check, Supp.~B) before exact object resolution, $\pm 0.5$\,s timestamp matching or $\text{IoU}\ge 0.5$ span matching, with unresolved outputs scored zero.
Model means weight the five tasks equally (per-task scores in \cref{tab:drop}, Supp.~E to I).
Paired 95\% confidence intervals in \cref{tab:history_scale,tab:repair_scale} use a dataset-stratified video-clustered bootstrap (4,000 draws) that resamples unique videos per dataset, shares weights across same-video queries and conditions, and recomputes task accuracies before averaging (Supp.~C).

%% file: fig/sharded_simulation_excalidraw_v2_tex.tex
\begin{figure}[t!]
  \centering
  \includegraphics[width=0.97\linewidth]{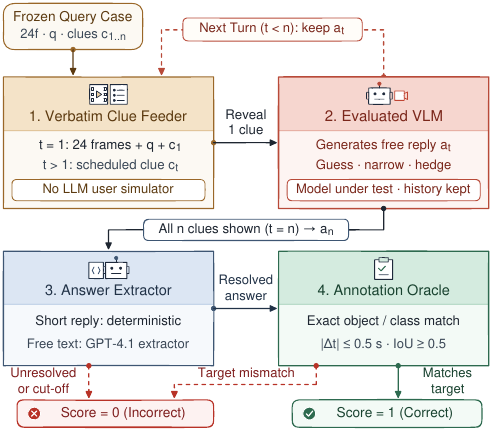}%
  \caption{\textbf{Generating and scoring multi-turn answers.} Verified clues arrive one per turn, with the video provided initially. Earlier model replies remain in context; the final reply is parsed and scored against annotations.}
  \label{fig:sharded_sim}
\end{figure}

%% file: fig/figure2_excalidraw_v3_tex.tex
\begin{figure*}[t]
  \centering
  \includegraphics[trim=0 1bp 0 1bp,clip,width=0.97\textwidth]{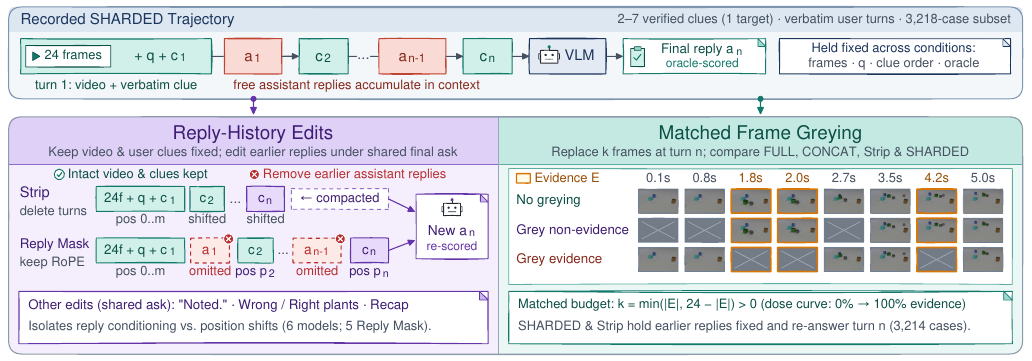}%
  \caption{\textbf{Controlled conversation edits.} Top: the original conversation. Left: earlier replies are removed, masked or replaced before regenerating the final answer. Right: equal numbers of evidence or non-evidence frames are greyed.}
  \label{fig:conditions}
\end{figure*}

%% file: sec/4_drop.tex
\section{Does Conversational Delivery Degrade Video QA?}
\label{sec:drop}

\textbf{Finding 1: Conversational delivery drops \full{}-normalized retention by $21.8$ percentage points beyond single-turn list formatting.}
Averaged equally across 18 models on 10,023 queries (6,822 videos), \concat retains $97.0\%$ of five-task \full accuracy whereas \sharded retains $75.2\%$ (\cref{tab:drop}; paired video-clustered 95\% CIs in Supp.~D).
List formatting thus costs $3.0$ retention points, versus a further $21.8$-point conversational drop ($10.7$ raw points, from $50.0\%$ to $39.3\%$).

\paragraph{Model and format variation.}
Following Laban \etal~\cite{laban2025lost}, we normalize each model's five-task means by \full:
\begin{equation}
  \Delta = 100\,\frac{\overline C-\overline S}{\overline F},
  \label{eq:delivery_gaps}
\end{equation}
where $\overline F$, $\overline C$ and $\overline S$ are unweighted five-task accuracies under \full, \concat and \sharded.
Every model has a positive conversational drop whose 95\% CI excludes zero (Supp.~D), spanning $\Delta=6.3$ on o3 and $8.0$ on Grok-4.3 up to $42.0$ on InternVL3.5-8B, with $>100$B Qwen3.5 models still losing $26.4$ to $29.8$ points.
Format sensitivity also varies before turns begin: Kimi-VL-A3B loses $8.2$ raw points from \full to \concat ($C/F=74.2\%$), whereas ten models retain $C/F\ge 98.5\%$, motivating the \concat control.
Because $\Delta$ divides by $\overline F$, raw drops ($3.9$ points on o3 to $20.6$ on InternVL3.5-8B) should be read alongside $\Delta$ on low-baseline tasks (\eg, $0.9\%$ LLaVA Charades grounding).

\input{tables/main_results}

\paragraph{Task-level, real-world and clue-order checks.}
Conversational loss is positive in $87$ of $90$ model-task cells ($82$ with 95\% CIs above zero), concentrating on multi-clue temporal and object reasoning (\cref{tab:drop}): 18-model mean $\overline C-\overline S$ is $16.4$ points on 7-clue CLEVRER timing, $13.3$ on Perception Test and $12.1$ on 7-clue CLEVRER object, versus $5.0$ on 2 to 3 clue Charades object and $6.6$ on Charades grounding ($31.1\%$ of its $21.1\%$ \full mean; o3 almost remains the same CLEVRER timing yet $12.9$ on Charades grounding).
Crucially, across the three real-world tasks alone ($3{,}353$ queries over $2{,}464$ Charades and Perception Test videos), all 18 models still degrade, with mean $\Delta=22.1$ points matching synthetic CLEVRER ($\Delta=21.4$; Supp.~D).
Testing three independent random clue orders across $250$ balanced queries ($50$ per task) on Qwen3-VL-8B and InternVL3.5-8B also confirms positive conversational drops under every order (all six 95\% CIs above zero; Supp.~D).
Because \sharded bundles turn splitting, absent query restatement and self-generated replies, \cref{sec:cause} edits recorded histories to separate them.

%% file: tables/main_results.tex
\begin{table*}[t]
  \caption{\textbf{Same video, three routes.}
  Accuracy (\%) across all five tasks under \full, \concat and \sharded (task query counts in \cref{tab:tasks}).
  Darker amber marks larger losses relative to \full in the same task.
  Relative scores use equal-task means ($C/F$, $S/F$); $\Delta=100(C/F-S/F)$ is their difference in percentage points (purple).
  Group means average per-model $\Delta$ equally; the bottom row weights all 18 models equally.}
  \label{tab:drop}
  \centering\footnotesize
  \setlength{\tabcolsep}{2.5pt}
  \setlength{\arrayrulewidth}{0.3pt}
  \renewcommand{\arraystretch}{1.08}
  \begin{tabular*}{\textwidth}{@{\extracolsep{\fill}}>{\scriptsize}l*{5}{c}!{\vrule width 0.8pt}*{5}{c}!{\vrule width 0.8pt}*{5}{c}!{\vrule width 0.8pt}ccc!{\vrule width 0.8pt}c@{}}
    \toprule
    & \multicolumn{5}{c!{\vrule width 0.8pt}}{\textbf{\full} ($F$)} & \multicolumn{5}{c!{\vrule width 0.8pt}}{\textbf{\concat} ($C$)} & \multicolumn{5}{c!{\vrule width 0.8pt}}{\textbf{\sharded} ($S$)} & \multicolumn{3}{c!{\vrule width 0.8pt}}{Relative to \full} & Group \\
    \cmidrule(lr){2-6}\cmidrule(lr){7-11}\cmidrule(lr){12-16}\cmidrule(lr){17-19}\cmidrule(l){20-20}
    & \multicolumn{2}{c}{CLEVRER} & \multicolumn{2}{c}{Charades} & PT & \multicolumn{2}{c}{CLEVRER} & \multicolumn{2}{c}{Charades} & PT & \multicolumn{2}{c}{CLEVRER} & \multicolumn{2}{c}{Charades} & PT & $C/F$ & $S/F$ & $\Delta$ & Mean $\Delta$ \\
    \cmidrule(lr){2-3}\cmidrule(lr){4-5}\cmidrule(lr){7-8}\cmidrule(lr){9-10}\cmidrule(lr){12-13}\cmidrule(lr){14-15}
    \multicolumn{1}{c}{\footnotesize Model} & Obj & Tim & Obj & Gnd & Obj & Obj & Tim & Obj & Gnd & Obj & Obj & Tim & Obj & Gnd & Obj & (\%) & (\%) & (pp) & (pp) \\
    \specialrule{\lightrulewidth}{0pt}{0pt}
    \rowcolor{plotteal!9}[0pt][0pt]
    \multicolumn{20}{@{}c@{}}{\strut \textbf{Small models}\quad $\leq 5$B total parameters\quad \textit{(2 models)}} \\
    \arrayrulecolor{black!25}\specialrule{0.3pt}{0pt}{0pt}\arrayrulecolor{black}
    \modelmark{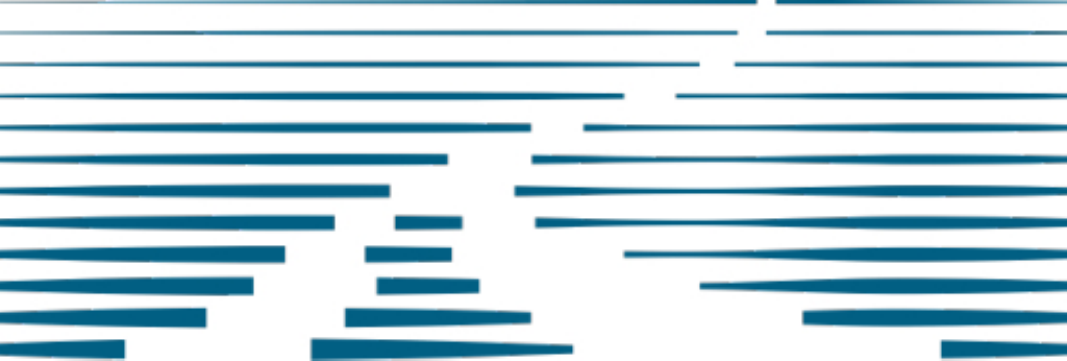}InternVL3.5-2B & 44.2 & 27.3 & 20.9 & 11.6 & 39.6 & \cellcolor{plotamber!20}40.2 & \cellcolor{plotamber!15}25.0 & 21.3 & \cellcolor{plotamber!23}6.5 & \cellcolor{plotamber!13}37.9 & \cellcolor{plotamber!35}32.7 & \cellcolor{plotamber!26}20.8 & \cellcolor{plotamber!23}15.6 & \cellcolor{plotamber!27}4.6 & \cellcolor{plotamber!37}26.5 & 91.1 & 69.8 & \cellcolor{plotpurple!36}21.4 & \multirow{2}{*}{17.0} \\
    \modelmark{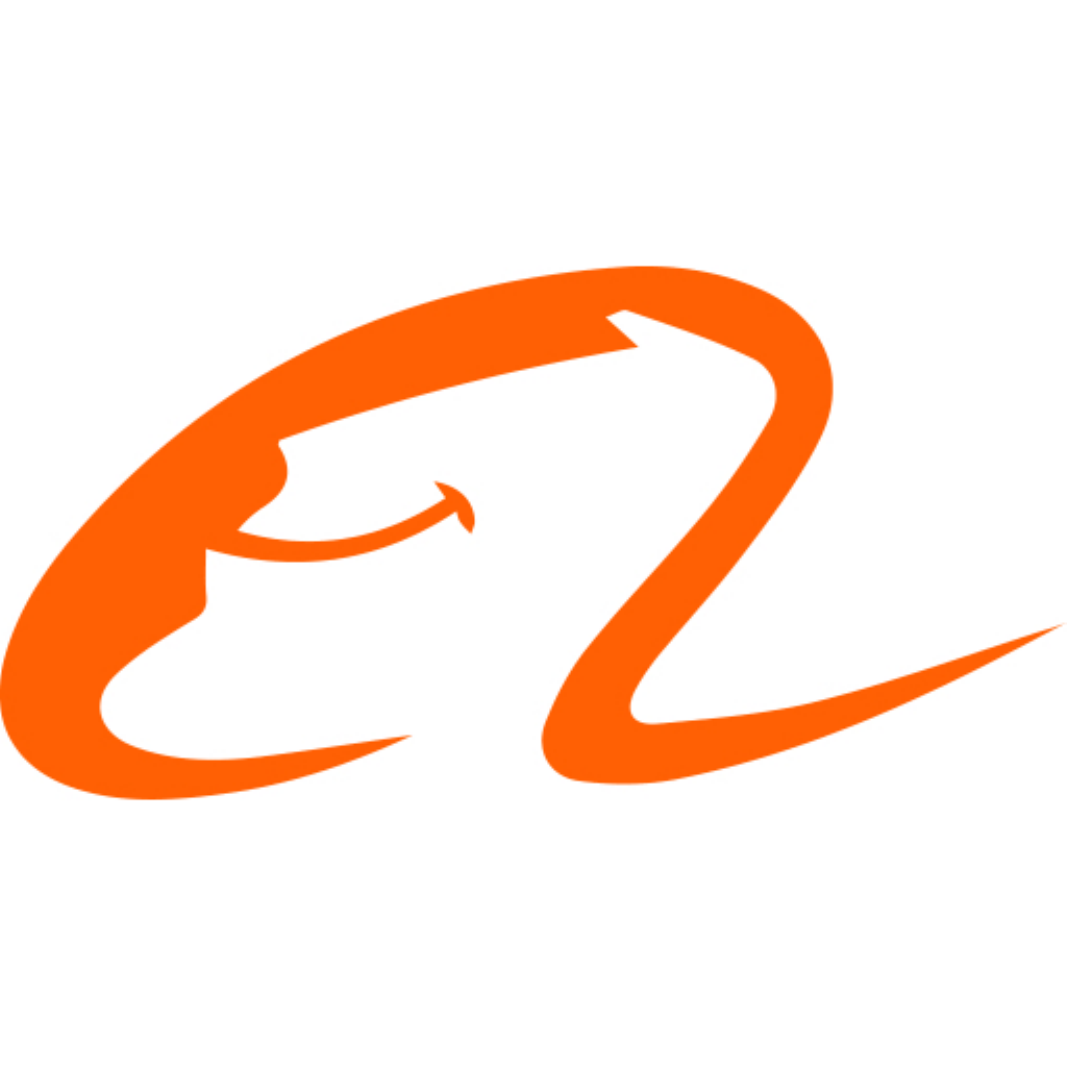}Qwen3-VL-4B & 70.1 & 57.7 & 31.6 & 23.5 & 59.4 & \cellcolor{plotamber!13}68.4 & \cellcolor{plotamber!29}49.8 & \cellcolor{plotamber!13}30.0 & \cellcolor{plotamber!6}23.2 & \cellcolor{plotamber!16}57.0 & \cellcolor{plotamber!35}58.2 & \cellcolor{plotamber!43}39.7 & \cellcolor{plotamber!11}30.4 & \cellcolor{plotamber!28}15.7 & \cellcolor{plotamber!24}53.7 & 94.3 & 81.6 & \cellcolor{plotpurple!28}12.7 &  \\
    \specialrule{\lightrulewidth}{0pt}{0pt}
    \rowcolor{plotteal!9}[0pt][0pt]
    \multicolumn{20}{@{}c@{}}{\strut \textbf{Mid-sized models}\quad 7--10B total parameters\quad \textit{(5 models)}} \\
    \arrayrulecolor{black!25}\specialrule{0.3pt}{0pt}{0pt}\arrayrulecolor{black}
    \hspace{10pt}LLaVA-Video-7B & 44.6 & 24.1 & 30.1 & 0.9 & 57.4 & 45.6 & \cellcolor{plotamber!11}23.0 & \cellcolor{plotamber!18}27.1 & \cellcolor{plotamber!0}0.9 & \cellcolor{plotamber!30}48.7 & \cellcolor{plotamber!43}26.7 & \cellcolor{plotamber!39}9.7 & \cellcolor{plotamber!26}23.8 & 1.1 & \cellcolor{plotamber!57}25.8 & 92.5 & 55.5 & \cellcolor{plotpurple!47}37.0 & \multirow{5}{*}{28.2} \\
    \modelmark{alibaba}VideoLLaMA3-7B & 50.7 & 29.5 & 28.8 & 18.4 & 52.0 & \cellcolor{plotamber!15}48.6 & 33.1 & 29.0 & \cellcolor{plotamber!21}14.0 & 53.0 & \cellcolor{plotamber!16}48.2 & \cellcolor{plotamber!32}19.9 & \cellcolor{plotamber!17}26.1 & \cellcolor{plotamber!31}9.1 & \cellcolor{plotamber!38}38.3 & 99.1 & 78.9 & \cellcolor{plotpurple!35}20.2 &  \\
    \modelmark{alibaba}Qwen3-VL-8B & 74.4 & 69.6 & 36.4 & 23.6 & 59.4 & \cellcolor{plotamber!18}71.2 & \cellcolor{plotamber!15}67.4 & \cellcolor{plotamber!7}36.0 & 24.2 & 61.4 & \cellcolor{plotamber!39}59.8 & \cellcolor{plotamber!49}46.7 & \cellcolor{plotamber!15}34.1 & \cellcolor{plotamber!27}16.5 & \cellcolor{plotamber!35}47.3 & 98.8 & 77.6 & \cellcolor{plotpurple!36}21.2 &  \\
    \modelmark{shanghai-ai-lab}InternVL3.5-8B & 86.6 & 72.6 & 25.9 & 13.1 & 47.0 & \cellcolor{plotamber!12}85.2 & 75.6 & 27.6 & 14.2 & \cellcolor{plotamber!6}46.6 & \cellcolor{plotamber!55}57.1 & \cellcolor{plotamber!68}27.4 & \cellcolor{plotamber!23}20.7 & \cellcolor{plotamber!20}9.2 & \cellcolor{plotamber!40}31.9 & 101.7 & 59.7 & \cellcolor{plotpurple!50}42.0 &  \\
    \modelmark{alibaba}Qwen3.5-9B & 67.0 & 66.8 & 30.1 & 22.2 & 62.1 & \cellcolor{plotamber!22}62.5 & \cellcolor{plotamber!25}61.0 & \cellcolor{plotamber!11}29.0 & 23.2 & \cellcolor{plotamber!28}54.4 & \cellcolor{plotamber!36}54.3 & \cellcolor{plotamber!51}41.7 & \cellcolor{plotamber!19}26.5 & \cellcolor{plotamber!26}15.8 & \cellcolor{plotamber!48}40.3 & 92.7 & 72.0 & \cellcolor{plotpurple!35}20.8 &  \\
    \specialrule{\lightrulewidth}{0pt}{0pt}
    \rowcolor{plotteal!9}[0pt][0pt]
    \multicolumn{20}{@{}c@{}}{\strut \textbf{Large models}\quad 14--34B total parameters\quad \textit{(5 models)}} \\
    \arrayrulecolor{black!25}\specialrule{0.3pt}{0pt}{0pt}\arrayrulecolor{black}
    \modelmark{shanghai-ai-lab}InternVL3.5-14B & 86.0 & 88.3 & 29.8 & 23.5 & 59.7 & \cellcolor{plotamber!20}82.3 & \cellcolor{plotamber!7}87.9 & \cellcolor{plotamber!17}27.0 & \cellcolor{plotamber!16}21.0 & 60.7 & \cellcolor{plotamber!52}59.8 & \cellcolor{plotamber!56}58.4 & \cellcolor{plotamber!24}24.2 & \cellcolor{plotamber!35}12.0 & \cellcolor{plotamber!47}38.3 & 97.1 & 67.0 & \cellcolor{plotpurple!43}30.0 & \multirow{5}{*}{21.6} \\
    \modelmark{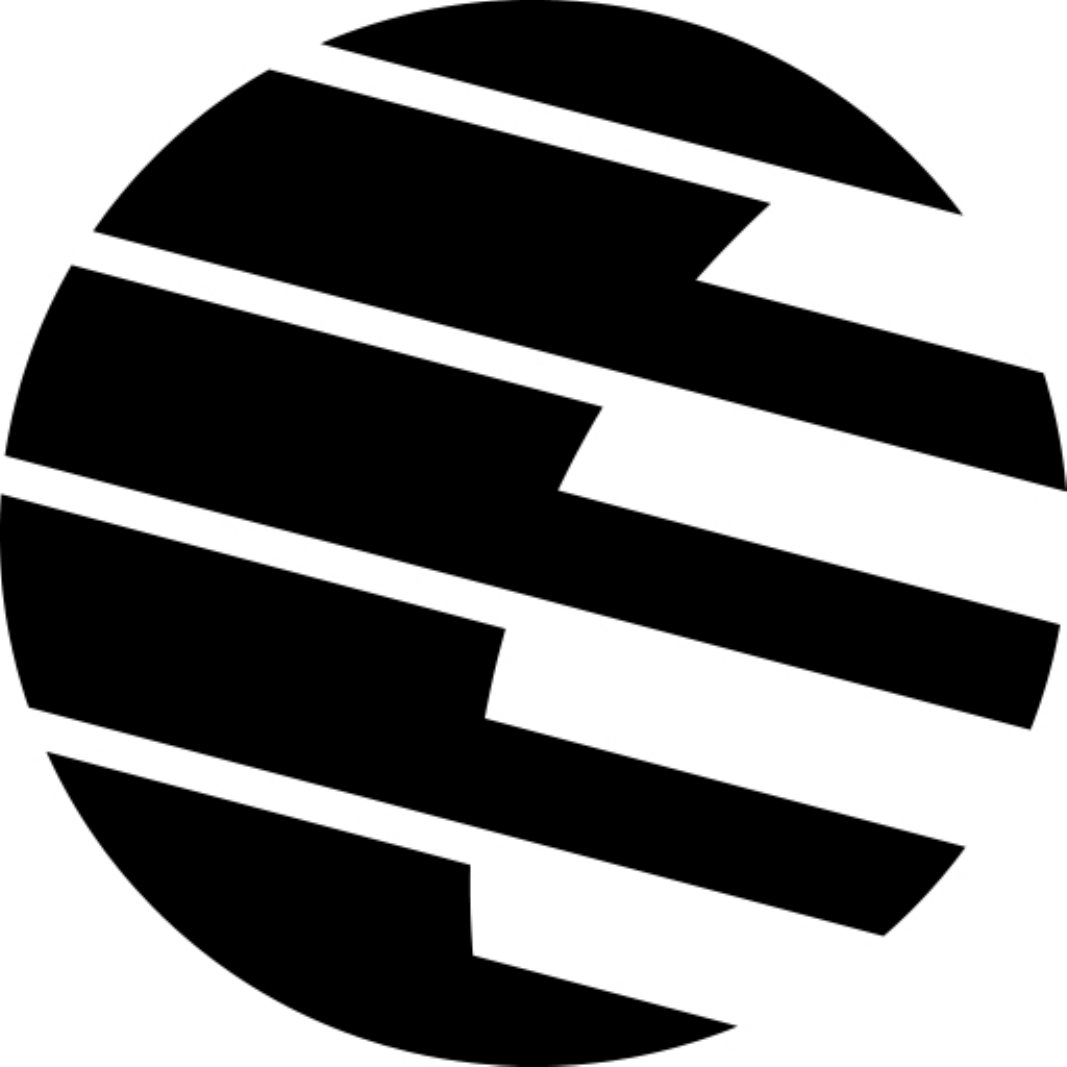}Kimi-VL-A3B & 38.4 & 40.1 & 26.4 & 5.4 & 49.0 & \cellcolor{plotamber!32}28.4 & \cellcolor{plotamber!28}32.3 & \cellcolor{plotamber!28}19.0 & \cellcolor{plotamber!7}4.9 & \cellcolor{plotamber!40}33.6 & \cellcolor{plotamber!37}25.4 & \cellcolor{plotamber!49}17.4 & \cellcolor{plotamber!44}7.3 & \cellcolor{plotamber!18}2.4 & \cellcolor{plotamber!54}20.5 & 74.2 & 45.8 & \cellcolor{plotpurple!41}28.4 &  \\
    \modelmark{alibaba}Qwen3-VL-30B-A3B & 64.2 & 45.0 & 39.3 & 28.8 & 65.8 & 66.1 & 47.5 & \cellcolor{plotamber!19}36.0 & \cellcolor{plotamber!17}26.0 & \cellcolor{plotamber!15}63.8 & \cellcolor{plotamber!31}54.8 & \cellcolor{plotamber!28}37.2 & \cellcolor{plotamber!32}29.2 & \cellcolor{plotamber!31}19.7 & \cellcolor{plotamber!34}54.4 & 98.5 & 80.3 & \cellcolor{plotpurple!33}18.2 &  \\
    \modelmark{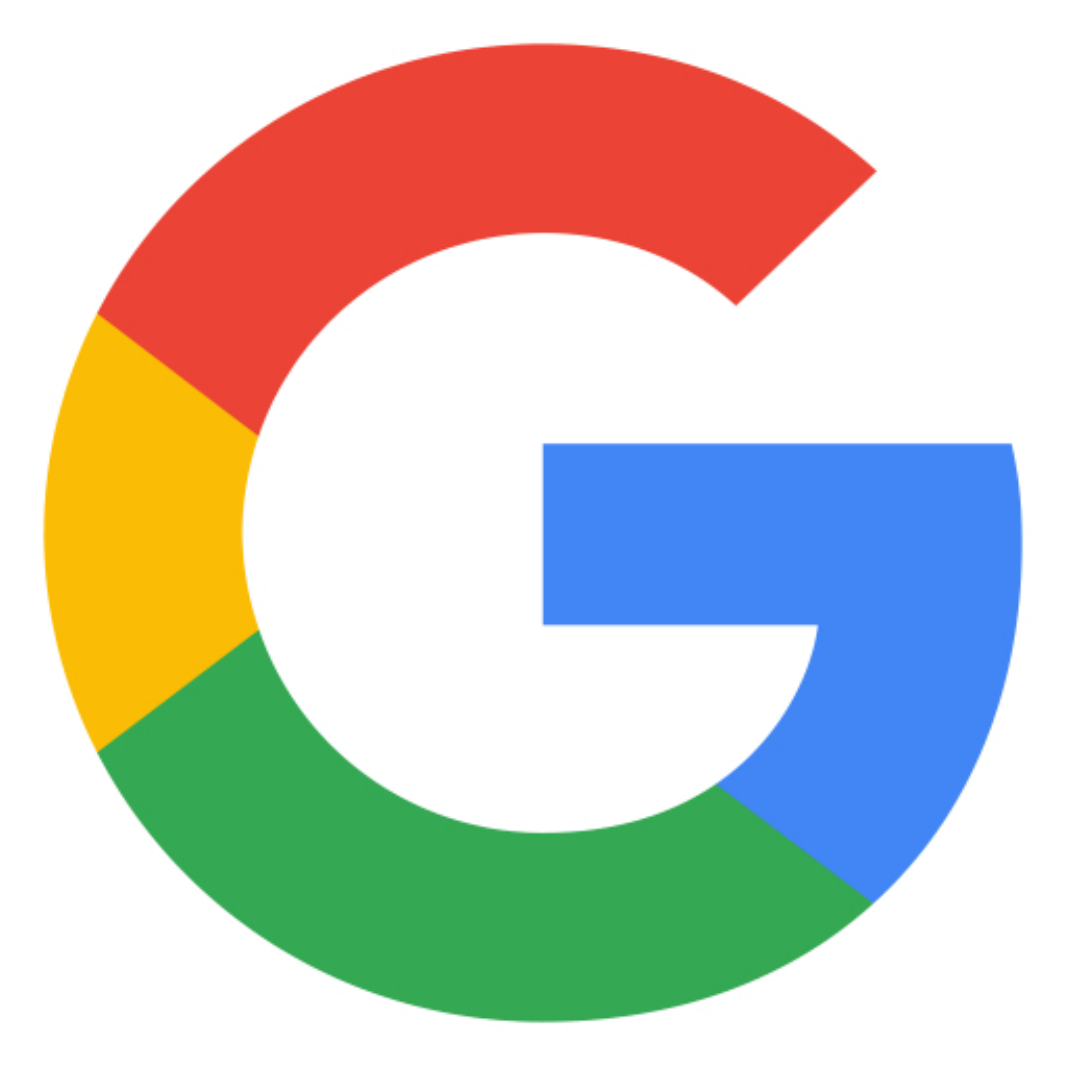}Gemma-4-31B & 78.6 & 93.1 & 40.2 & 23.2 & 74.5 & 79.4 & 94.3 & \cellcolor{plotamber!6}39.9 & 24.8 & \cellcolor{plotamber!16}72.1 & \cellcolor{plotamber!36}65.9 & \cellcolor{plotamber!29}85.1 & \cellcolor{plotamber!22}35.6 & \cellcolor{plotamber!21}18.8 & \cellcolor{plotamber!42}57.4 & 100.3 & 84.9 & \cellcolor{plotpurple!31}15.4 &  \\
    \modelmark{alibaba}Qwen3-VL-32B & 87.1 & 84.5 & 38.1 & 29.7 & 70.5 & \cellcolor{plotamber!12}85.7 & \cellcolor{plotamber!10}83.5 & 38.1 & \cellcolor{plotamber!14}27.8 & \cellcolor{plotamber!10}69.5 & \cellcolor{plotamber!33}76.4 & \cellcolor{plotamber!33}74.3 & \cellcolor{plotamber!21}33.7 & \cellcolor{plotamber!34}18.3 & \cellcolor{plotamber!44}52.0 & 98.3 & 82.2 & \cellcolor{plotpurple!31}16.1 &  \\
    \specialrule{\lightrulewidth}{0pt}{0pt}
    \rowcolor{plotteal!9}[0pt][0pt]
    \multicolumn{20}{@{}c@{}}{\strut \textbf{Very large models}\quad $>100$B total parameters\quad \textit{(2 models)}} \\
    \arrayrulecolor{black!25}\specialrule{0.3pt}{0pt}{0pt}\arrayrulecolor{black}
    \modelmark{alibaba}Qwen3.5-122B-A10B & 78.5 & 73.9 & 37.9 & 26.3 & 69.5 & 79.2 & 75.0 & 38.8 & 28.5 & \cellcolor{plotamber!8}68.8 & \cellcolor{plotamber!34}67.3 & \cellcolor{plotamber!57}42.1 & \cellcolor{plotamber!31}28.5 & \cellcolor{plotamber!21}21.9 & \cellcolor{plotamber!50}45.3 & 101.5 & 71.7 & \cellcolor{plotpurple!42}29.8 & \multirow{2}{*}{28.1} \\
    \modelmark{alibaba}Qwen3.5-397B-A17B & 85.0 & 78.1 & 37.9 & 36.5 & 62.8 & 85.1 & 83.6 & 39.0 & \cellcolor{plotamber!14}34.6 & 68.5 & \cellcolor{plotamber!46}64.7 & \cellcolor{plotamber!47}56.7 & \cellcolor{plotamber!25}31.7 & \cellcolor{plotamber!30}27.6 & \cellcolor{plotamber!35}50.7 & 103.5 & 77.1 & \cellcolor{plotpurple!40}26.4 &  \\
    \specialrule{\lightrulewidth}{0pt}{0pt}
    \rowcolor{plotteal!9}[0pt][0pt]
    \multicolumn{20}{@{}c@{}}{\strut \textbf{Frontier API models}\quad \textit{(4 models)}} \\
    \arrayrulecolor{black!25}\specialrule{0.3pt}{0pt}{0pt}\arrayrulecolor{black}
    \modelmark{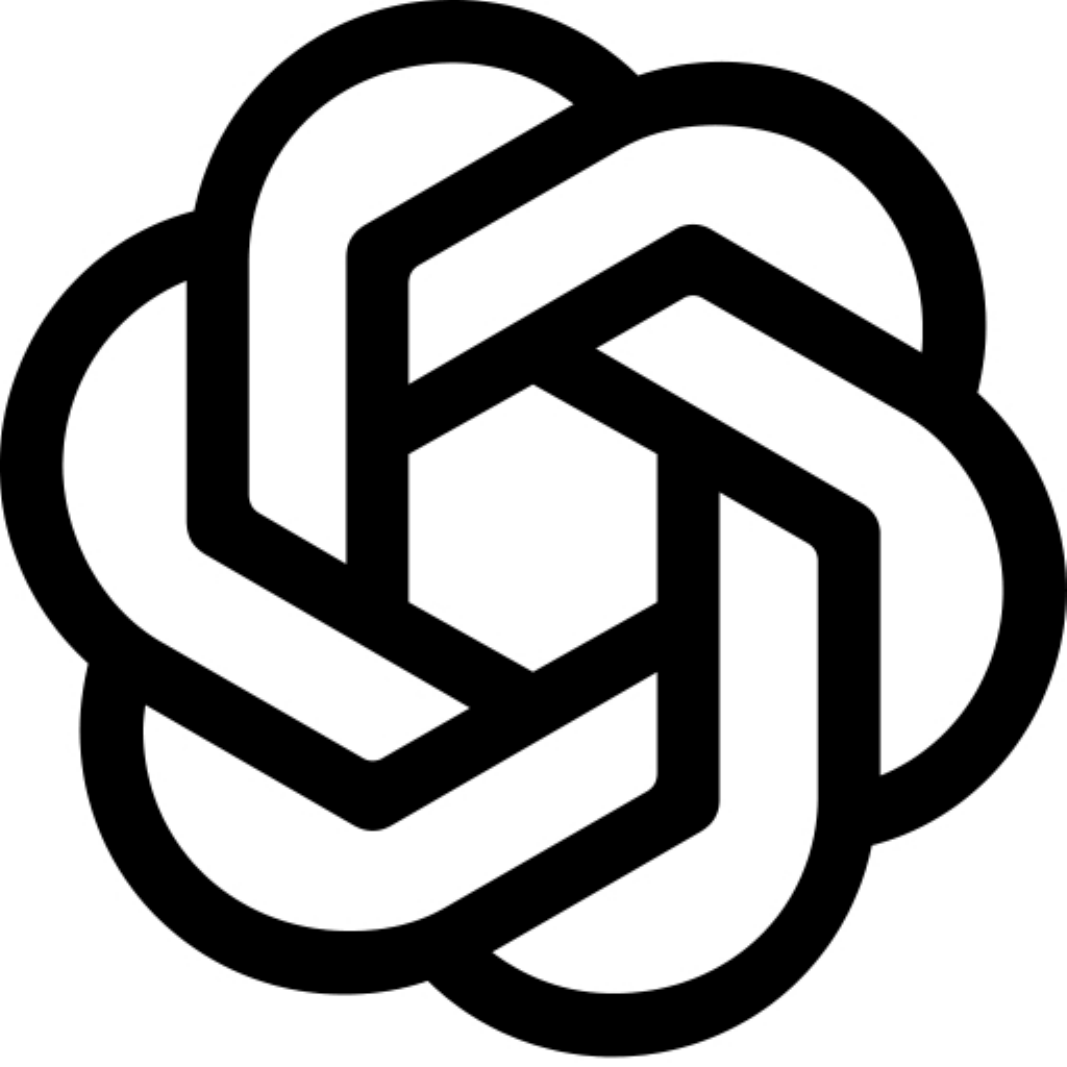}GPT-4.1 & 70.2 & 79.2 & 41.3 & 21.4 & 69.8 & 71.4 & 80.2 & 41.8 & 22.5 & \cellcolor{plotamber!8}69.1 & \cellcolor{plotamber!38}56.3 & \cellcolor{plotamber!34}68.0 & \cellcolor{plotamber!22}36.6 & \cellcolor{plotamber!28}14.0 & \cellcolor{plotamber!37}56.4 & 101.1 & 82.1 & \cellcolor{plotpurple!34}19.1 & \multirow{4}{*}{13.1} \\
    \modelmark{openai}GPT-5.4 & 77.0 & 95.5 & 40.1 & 20.4 & 70.8 & 78.1 & \cellcolor{plotamber!3}95.4 & 41.2 & 22.0 & \cellcolor{plotamber!8}70.1 & \cellcolor{plotamber!44}58.2 & \cellcolor{plotamber!31}86.5 & \cellcolor{plotamber!32}30.3 & \cellcolor{plotamber!28}13.1 & \cellcolor{plotamber!32}61.1 & 101.0 & 82.0 & \cellcolor{plotpurple!34}19.0 &  \\
    \modelmark{openai}o3 & 80.0 & 83.5 & 39.9 & 29.8 & 75.2 & 80.1 & 84.0 & 41.8 & \cellcolor{plotamber!13}28.3 & 75.8 & \cellcolor{plotamber!12}78.6 & 84.2 & \cellcolor{plotamber!10}38.8 & \cellcolor{plotamber!39}15.4 & \cellcolor{plotamber!12}73.8 & 100.6 & 94.3 & \cellcolor{plotpurple!19}6.3 &  \\
    \modelmark{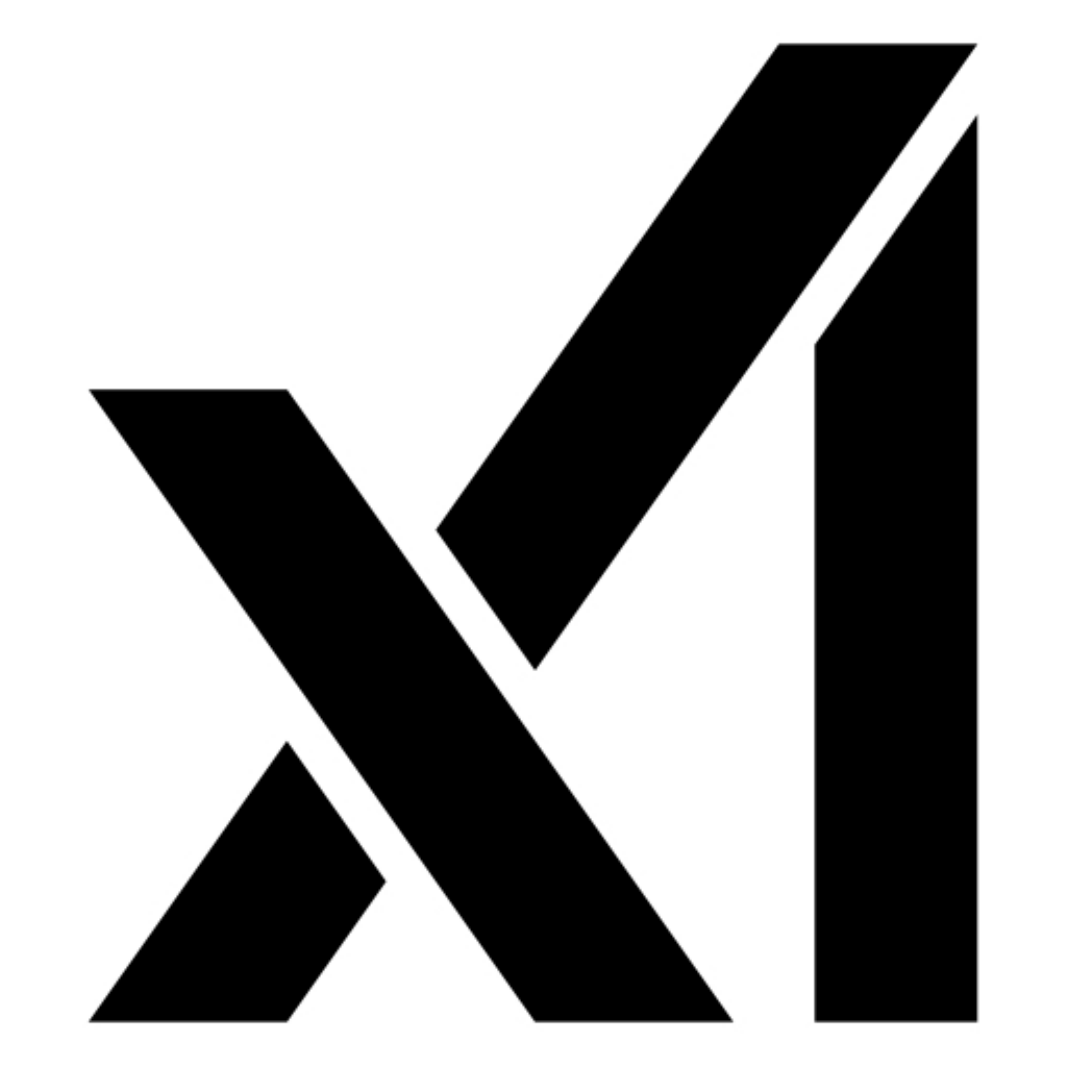}Grok-4.3 & 96.5 & 94.5 & 38.6 & 21.0 & 74.8 & \cellcolor{plotamber!10}95.5 & 94.5 & 38.8 & 22.6 & \cellcolor{plotamber!13}73.2 & \cellcolor{plotamber!24}91.0 & \cellcolor{plotamber!34}83.1 & \cellcolor{plotamber!7}38.1 & \cellcolor{plotamber!23}16.0 & \cellcolor{plotamber!21}70.5 & 99.8 & 91.8 & \cellcolor{plotpurple!22}8.0 &  \\
    \midrule
    \emph{Mean (18 models)} & 71.1 & 66.8 & 34.1 & 21.1 & 62.2 & \cellcolor{plotamber!12}69.6 & \cellcolor{plotamber!8}66.3 & \cellcolor{plotamber!8}33.4 & \cellcolor{plotamber!7}20.5 & \cellcolor{plotamber!14}60.2 & \cellcolor{plotamber!38}57.5 & \cellcolor{plotamber!42}49.9 & \cellcolor{plotamber!24}28.4 & \cellcolor{plotamber!27}14.0 & \cellcolor{plotamber!40}46.9 & 97.0 & 75.2 & \cellcolor{plotpurple!36}21.8 & - \\
    \bottomrule
  \end{tabular*}
\end{table*}

%% file: sec/5_history.tex
\section{How Do Earlier Replies Contribute to the Drop?}
\label{sec:cause}

\textbf{Finding 2: Removing or masking prior replies recovers part of the conversational loss, while wrong prior replies actively bias later answers.}
To interpret edits to recorded dialogues on the 3,218-query (2,942-video) subset (\cref{tab:history_scale}), we distinguish three complementary findings with differing levels of evidential support: (i)~conversational delivery degrades accuracy overall (\cref{sec:drop}), (ii)~retaining earlier assistant replies contributes to part of that degradation alongside non-reply turn factors, and (iii)~incorrect prior replies actively bias subsequent answers.

\subsection{Reply removal, position masking and turn factors}

Post-hoc deletion of earlier assistant turns (\replystripped, keeping only user turns before regenerating the final answer) improves five-task accuracy over the matched \sharded baseline in all six models, from $+2.4$ points on LLaVA-Video-7B to $+10.0$ and $+12.1$ on InternVL3.5-8B and -14B (\cref{tab:history_scale}).
Because \replystripped compacts consecutive user turns and shifts rotary positions, we test white-box \emph{Reply Masking} across all five open-weight models, omitting assistant spans while keeping surviving rotary position IDs fixed.
Reply Masking still gains $+2.2$ to $+12.6$ points over each model's matched white-box baseline (Supp.~F), ruling out rotary position shifts alone, though it still shortens the active sequence and key-value cache rather than isolating pure semantics.
Prior replies also explain only part of the single-turn gap (side-by-side summary of \full, \concat, \sharded, \replystripped, Reply Masking and \emph{Recap} on the 3,218 subset in Supp.~F): final-turn restatement (\emph{Recap}) adds $+1.6$ to $+11.7$ points (\cref{tab:history_scale}), and \replystripped leaves residual turn-fragmentation loss on models such as LLaVA-Video-7B and InternVL3.5-8B (Supp.~E); these distinct interventions do not sum additively.

\input{tables/history_results}

\input{fig/greying_summary}

\subsection{Wrong-answer bias and neutral acknowledgments}

Content replacements test how strongly final answers track supplied prior commitments (\cref{tab:history_scale}).
Planting a consistent wrong candidate (\emph{Wrong}) degrades all six models (95\% CIs below zero), from $-2.2$ points on GPT-4.1 to $-28.9$ on Qwen3-VL-8B, whereas planting the target (\emph{Right}; $59.1\text{ to }92.7\%$) is an answer-leaking upper check rather than a repair or an estimate of natural error share.
Fixed \noted replacements improve four of six models ($+2.5$ on Qwen3-VL-8B, $+6.9$ on GPT-4.1, $+7.4$ to $+8.8$ on InternVL3.5), leave LLaVA unresolved ($-0.7$), and collapse Kimi-VL-A3B ($20.8\%$ to $7.9\%$) via final-turn \noted copying on $2{,}126$ of $3{,}218$ queries (post-hoc edits, not a wait-to-answer prompt; Supp.~E).
Because retained replies and wrong commitments shift final predictions, \cref{sec:lookuse} tests how strongly final answers still depend on the video frames.

%% file: tables/history_results.tex
\begin{table*}[t]
\caption{\textbf{Removing earlier replies improves all six models.} Five-task mean accuracy (\%) on 3,218 queries per model. Blue highlights \replystripped; amber marks losses from the matched \sharded{} baseline; purple shows paired \replystripped gains (pp) with pointwise video-clustered 95\% bootstrap CIs. Shading indicates magnitude, not significance. Right reveals the correct answer (not a repair); Recap repeats full query.}
\label{tab:history_scale}
\centering\footnotesize\setlength{\tabcolsep}{3pt}
\renewcommand{\arraystretch}{1.2}
\renewcommand{\arraystretch}{1.08}
\setlength{\arrayrulewidth}{0.3pt}
\begin{tabular}{>{\centering\arraybackslash}p{\dimexpr 0.17\textwidth - 2\tabcolsep - 0.3pt\relax}|>{\centering\arraybackslash}p{\dimexpr 0.105\textwidth - 2\tabcolsep\relax}>{\centering\arraybackslash}p{\dimexpr 0.105\textwidth - 2\tabcolsep\relax}>{\centering\arraybackslash}p{\dimexpr 0.105\textwidth - 2\tabcolsep\relax}>{\centering\arraybackslash}p{\dimexpr 0.105\textwidth - 2\tabcolsep - 0.8pt\relax}!{\vrule width 0.8pt}>{\centering\arraybackslash}p{\dimexpr 0.105\textwidth - 2\tabcolsep\relax}>{\centering\arraybackslash}p{\dimexpr 0.105\textwidth - 2\tabcolsep - 0.8pt\relax}!{\vrule width 0.8pt}>{\centering\arraybackslash}p{\dimexpr 0.20\textwidth - 2\tabcolsep\relax}}
\specialrule{\heavyrulewidth}{0pt}{0pt}
\rowcolor{plotteal!9}
\strut & \multicolumn{4}{c!{\vrule width 0.8pt}}{\strut\textbf{Reply-history edits}} & \multicolumn{2}{c!{\vrule width 0.8pt}}{\strut\textbf{Answer plants}} & \strut\textbf{Key contrast} \\
\arrayrulecolor{black!25}\specialrule{0.3pt}{0pt}{0pt}\arrayrulecolor{black}
\strut Model & \sharded{}\allowbreak & \replystripped & Noted & Recap & Wrong & Right & \shortstack{\replystripped{} $-$ \sharded{}} \\
\specialrule{\lightrulewidth}{0pt}{0pt}
\strut Qwen3-VL-8B & 44.6 & \cellcolor{plotblue!7}50.8 & 47.1 & 46.2 & \cellcolor{plotamber!42}15.7 & 92.7 & \cellcolor{plotpurple!19}\textbf{+6.2} [4.7, 7.8] \\
\strut InternVL3.5-8B & 35.1 & \cellcolor{plotblue!7}45.1 & 42.5 & 46.8 & \cellcolor{plotamber!33}17.6 & 78.8 & \cellcolor{plotpurple!25}\textbf{+10.0} [8.0, 12.0] \\
\strut InternVL3.5-14B & 40.4 & \cellcolor{plotblue!7}52.5 & 49.2 & 48.9 & \cellcolor{plotamber!32}23.2 & 70.7 & \cellcolor{plotpurple!27}\textbf{+12.1} [10.2, 14.1] \\
\strut LLaVA-Video-7B & 22.4 & \cellcolor{plotblue!7}24.8 & \cellcolor{plotamber!7}21.7 & 25.8 & \cellcolor{plotamber!32}5.9 & 90.5 & \cellcolor{plotpurple!12}\textbf{+2.4} [0.7, 4.1] \\
\strut Kimi-VL-A3B & 20.8 & \cellcolor{plotblue!7}27.1 & \cellcolor{plotamber!28}7.9 & 26.9 & \cellcolor{plotamber!27}8.9 & 76.9 & \cellcolor{plotpurple!20}\textbf{+6.3} [4.4, 8.2] \\
\strut GPT-4.1 & 48.2 & \cellcolor{plotblue!7}53.8 & 55.1 & 50.4 & \cellcolor{plotamber!11}46.0 & 59.1 & \cellcolor{plotpurple!18}\textbf{+5.6} [4.0, 7.2] \\
\specialrule{\heavyrulewidth}{0pt}{0pt}
\end{tabular}
\end{table*}

%% file: fig/greying_summary.tex
\begin{figure*}[t]
  \centering
  \includegraphics[trim=0 0bp 0 10bp,clip,width=0.98\linewidth]{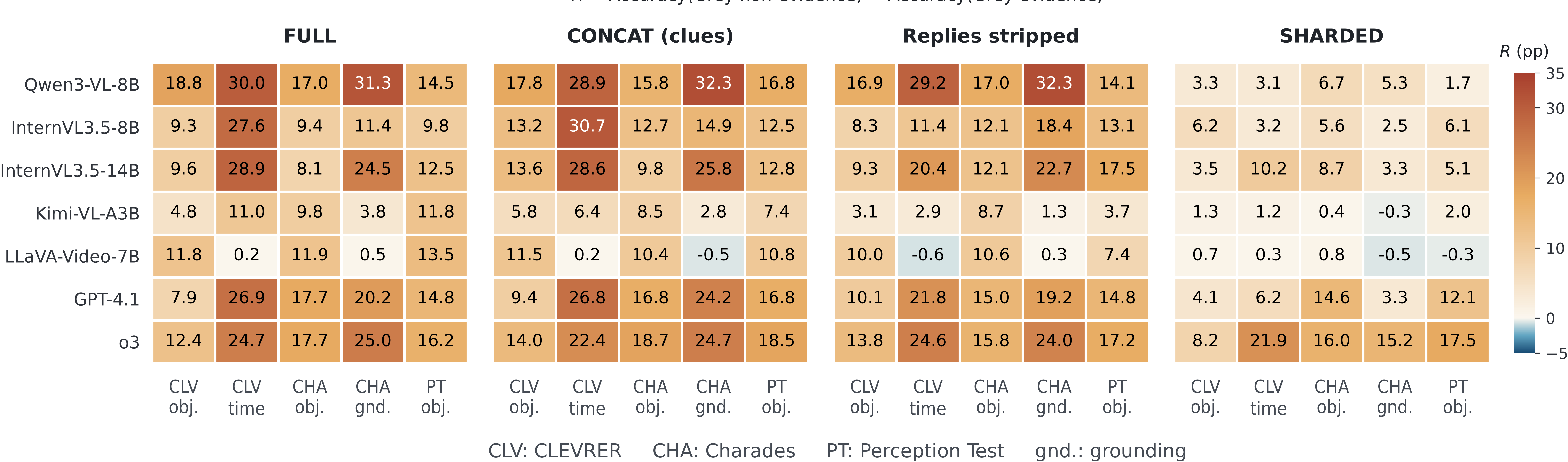}%
  \caption{\textbf{Final-answer visual sensitivity across models and tasks.} Cells report $R$, defined as non-evidence-greyed minus evidence-greyed accuracy (pp); warmer colours indicate greater sensitivity to evidence frames under equal-count greying.}
  \label{fig:greying_summary}
\end{figure*}

%% file: sec/5_lookuse.tex
\section{How Do Earlier Replies Alter Dependence on Visual Input?}
\label{sec:lookuse}

\begin{figure*}[t]
  \centering
  \includegraphics[trim=0 4bp 0 1bp,clip,width=0.98\linewidth]{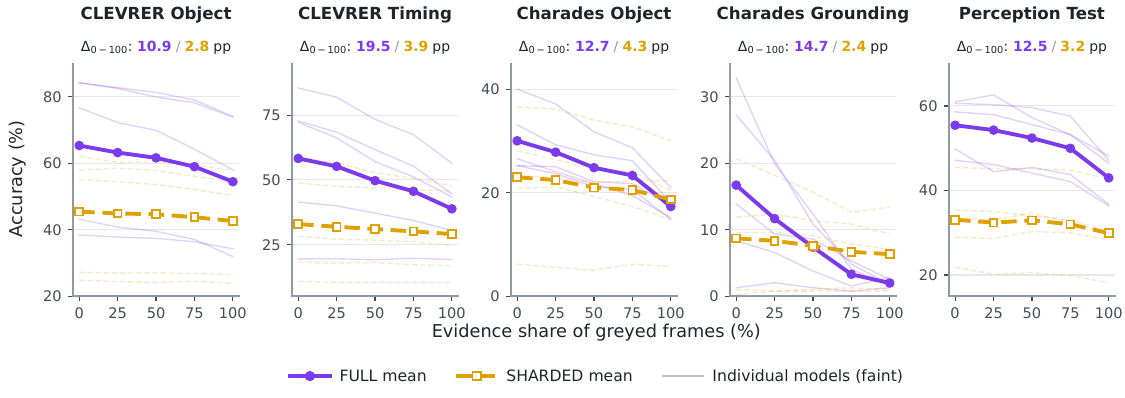}%
  \caption{\textbf{Increasing evidence greying affects \full more than \sharded.} The number of greyed frames stays fixed while their evidence share increases. Bold curves average five models equally; faint curves show individual models. Headers report accuracy at 0\% minus 100\% evidence share (pp). Panels use different y-axis ranges.}
  \label{fig:greying_dose_by_task}
\end{figure*}

\textbf{Finding 3: Retaining earlier assistant replies reduces final-answer dependence on supporting video frames, and removing them increases sensitivity in all seven tested models.}
Having shown in \cref{sec:cause} that retained replies shift final predictions, we test how strongly final answers rely on current visual input versus accumulated text across 3,214 eligible queries (2,938 videos) spanning all five tasks.
For each query with annotated evidence frames $E$, we compare three matched visual inputs (\cref{fig:conditions}, right): no greying, $k=\min(|E|,24-|E|)$ evidence frames greyed, and $k$ random non-evidence frames greyed, evaluated across \full, \concat, \replystripped and \sharded on identical frame sequences.

\subsection{Model- and task-level sensitivity across contexts}

\Cref{fig:greying_summary} reports visual sensitivity $R$ (non-evidence-greyed minus evidence-greyed accuracy, in percentage points) across all seven models and five tasks (raw accuracies in Supp.~G).
In single-turn prompts, greying evidence rather than $k$ non-evidence frames causes large drops across tasks (seven-model five-task mean $R=15.0$ points in \full and $15.5$ in \concat).
In \sharded (\cref{fig:greying_summary}, right), sensitivity drops in every model (mean $R=5.7$ points; paired $R_{\sharded}-R_{\full}$ spans $-3.5$ on o3 to $-18.3$ on Qwen3-VL-8B, all seven 95\% CIs below zero).
Removing prior assistant replies (\replystripped; third block) lifts mean sensitivity back to $R=13.7$ points, with paired $R_{\replystripped}-R_{\sharded}$ positive in all seven models ($+3.0$ on Kimi-VL-A3B and $+3.3$ on o3 up to $+10.3$ on InternVL3.5-14B and $+17.9$ on Qwen3-VL-8B).

Across individual cells (\cref{fig:greying_summary}), Qwen3-VL-8B collapses from $R=14.5\text{ to }31.3$ points across all five tasks in \full to $1.7\text{ to }6.7$ in \sharded, then rebounds to $14.1\text{ to }32.3$ in \replystripped.
Reasoning model o3 retains strong sensitivity even in \sharded (mean $R=15.7$, including $21.9$ on CLEVRER timing and $17.5$ on Perception Test), whereas near-zero cells on LLaVA-Video-7B ($0.2\text{ to }0.5$ on CLEVRER timing and Charades grounding even in \full) and Kimi-VL-A3B reflect floor task accuracy rather than visual independence.
Crucially, because frames in $E$ vary in importance and non-evidence frames can still carry scene context (\eg, object appearance), $R$ specifically measures behavioural sensitivity of the final answer to annotated supporting event frames relative to text history (\eg, prior commitments or verbalized observations), rather than visual encoding or visual grounding in general.

\subsection{Graded evidence removal at a fixed greying budget}

Fixing the greyed budget at $k$ frames and varying the share drawn from $E$ from $0\%$ to $100\%$ across the five open-weight models (\cref{fig:greying_dose_by_task}; Supp.~G) yields a steep, monotonic decline in \full across all five tasks ($10.9$ to $19.5$ points), versus $2.4$ to $4.3$ points in \sharded.
Together with \cref{sec:cause}, these diagnostics show that retained replies both lower accuracy and reduce reliance on video frames; \cref{sec:fix} tests how much recovery comes from masking those replies versus adding clean-state steering.

%% file: sec/6_repair.tex
\section{Repairing Interference: Masking and Steering}
\label{sec:fix}

\input{tables/repair_results}

Retained assistant replies lower accuracy and dampen reliance on supporting frames (\cref{sec:cause,sec:lookuse}), yet omitting them does not always close the gap to a consolidated single-turn prompt.
Across 3,218 queries (2,942 videos) on four open-weight models (\cref{tab:repair_scale}), we therefore evaluate position-preserving \emph{Reply Masking} as our primary history intervention and instance-specific \emph{Clean-State Steering} (\textbf{\cleanturn}; \cref{fig:method}) as an incremental test alongside \emph{Recap}, token-score mixing (\emph{Mix}) and \full answering.

\subsection{Reply Masking and Clean-State Steering}
\label{sec:method}

\input{fig/figure5_excalidraw}

\emph{Reply Masking} omits earlier assistant spans from attention and the key-value cache while preserving the rotary position IDs of all video and user tokens, adding no auxiliary forward pass.
To test whether clean-prompt states add further recovery, let $h^\ell(x)$ denote the residual stream at the final prompt token in decoder layer $\ell$, sharing the system prompt and 24 frames across views: $S$ deterministically joins the question and revealed clues in recorded order into one \full message (no target label or learned rewriter), $B$ keeps only the question, and $C$ is the reply-masked (\cleanturn) or unmasked (Steering-only) \sharded context.
\emph{Clean-State Steering} computes one per-query residual difference,
\begin{equation}
  v_{\mathrm{shadow}}^\ell = h^\ell(S)-h^\ell(B)
  \quad\text{or}\quad
  v_{\mathrm{delta}}^\ell = h^\ell(S)-h^\ell(C),
  \label{eq:vector}
\end{equation}
and injects it across a frozen layer band $\Lambda$ at every token $p$ of the final user turn and during generation:
\begin{equation}
  h_p^\ell \leftarrow h_p^\ell+\alpha v^\ell,\qquad \ell\in\Lambda.
  \label{eq:inject}
\end{equation}
Model weights remain frozen, while difference type, $\Lambda$, and fixed versus per-query KL-calibrated $\alpha\in[0,1]$ (against a 24-token greedy reference from $S$) are selected per model on case-disjoint tuning queries (600 cases, or 400 on InternVL3.5-8B; Supp.~H).
Because $v^\ell$ contrasts whole prompts, it carries both clue-content and format signal rather than a purely visual direction.

\subsection{Masking-driven recovery vs.\ incremental steering}
\label{sec:fix_results}
\label{sec:fix_scale}

\textbf{Finding 4: Reply Masking drives baseline recovery on three of four models, while Clean-State Steering adds resolved gains on two models.}
In \cref{tab:repair_scale}, Reply Masking alone improves over \sharded across all four models without auxiliary passes: $+12.6$ points on InternVL3.5-14B, $+9.0$ on InternVL3.5-8B, $+5.2$ on Qwen3-VL-8B and $+2.2$ on LLaVA-Video-7B.
Adding Clean-State Steering (\cleanturn; $+7.1$ to $+12.4$ points overall, Supp.~I) yields a large additional gain over masking on LLaVA-Video-7B ($+8.2$ points, where user-turn fragmentation leaves masking weak) and a smaller gain on Qwen3-VL-8B ($+1.9$), with both intervals remaining positive after excluding $126$ cross-task tuning-video overlaps (Supp.~I).
In contrast, steering adds no resolved gain beyond masking on InternVL3.5-14B ($-0.2$) or InternVL3.5-8B ($+0.6$).

\subsection{Comparison with Recap, Mix and \full}

Comparing against prompt and decoding baselines clarifies practical trade-offs (\cref{tab:repair_scale}).
When the consolidated prompt $S$ can be answered directly (\full), it uses one clean pass, works on black-box APIs, is close to \cleanturn on LLaVA ($31.3\%$ vs.\ $32.7\%$, $+1.4$) and Qwen ($52.1\%$ vs.\ $51.6\%$, $-0.6$), and beats \cleanturn by $4.4$ and $6.0$ points on InternVL3.5-14B and -8B (Supp.~I; white-box \replystripped in Supp.~F likewise tracks masking).
Among baselines that keep the multi-turn dialogue, \cleanturn outperforms \emph{Recap} ($+7.0$ and $+5.4$ points) and fixed 0.5-weight score mixing (\emph{Mix}; $+5.8$ and $+4.4$) on LLaVA and Qwen; however, on InternVL3.5-14B \emph{Mix} ($53.7\%$) matches \cleanturn ($52.6\%$; $-1.1$), and on InternVL3.5-8B both \emph{Recap} ($48.1\%$) and \emph{Mix} ($46.0\%$) beat masking ($43.6\%$) and \cleanturn ($44.2\%$).

\paragraph{Specificity and scope.}
Swapping in another query's steering vector lowers accuracy atop masking by $6.0$ points on LLaVA and $1.8$ on Qwen, confirming query-specific signal across video and clues.
In white-box serving, Reply Masking adds zero auxiliary forward passes, whereas \cleanturn trades additional final-turn passes over $S$ (plus reference scoring under auto-$\alpha$) for steering in place rather than restarting the session as \full (Supp.~H).
Reply Masking thus serves as a lightweight default, while steering adds targeted, model-dependent gains under residual turn-fragmentation loss.

%% file: tables/repair_results.tex
\begin{table*}[t]
\caption{\textbf{Reply Masking drives baseline recovery; Clean-State Steering adds model-dependent gains on two models.} Five-task mean accuracy (\%) on 3,218 queries (2,942 videos) per condition. \full answers from the consolidated clean prompt; amber marks \sharded{} losses from \full. The right two columns decompose \cleanturn recovery into paired Reply Masking gains over \sharded{} and incremental Clean-State Steering gains over Reply Masking (pp, video-clustered 95\% CIs).}
\label{tab:repair_scale}
\centering\footnotesize\setlength{\tabcolsep}{2.2pt}
\renewcommand{\arraystretch}{1.2}
\renewcommand{\arraystretch}{1.08}
\setlength{\arrayrulewidth}{0.3pt}
\begin{tabular}{@{}>{\centering\arraybackslash}p{\dimexpr 0.15\textwidth - 2\tabcolsep - 0.3pt\relax}!{\vrule width 0.3pt}>{\centering\arraybackslash}p{\dimexpr 0.065\textwidth - 2\tabcolsep\relax}>{\centering\arraybackslash}p{\dimexpr 0.085\textwidth - 2\tabcolsep - 0.8pt\relax}!{\vrule width 0.8pt}>{\centering\arraybackslash}p{\dimexpr 0.06\textwidth - 2\tabcolsep\relax}>{\centering\arraybackslash}p{\dimexpr 0.06\textwidth - 2\tabcolsep - 0.8pt\relax}!{\vrule width 0.8pt}>{\centering\arraybackslash}p{\dimexpr 0.075\textwidth - 2\tabcolsep\relax}>{\centering\arraybackslash}p{\dimexpr 0.07\textwidth - 2\tabcolsep\relax}>{\centering\arraybackslash}p{\dimexpr 0.085\textwidth - 2\tabcolsep - 0.8pt\relax}!{\vrule width 0.8pt}>{\centering\arraybackslash}p{\dimexpr 0.175\textwidth - 2\tabcolsep - 0.3pt\relax}!{\vrule width 0.3pt}>{\centering\arraybackslash}p{\dimexpr 0.175\textwidth - 2\tabcolsep\relax}@{}}
\specialrule{\heavyrulewidth}{0pt}{0pt}
\rowcolor{plotteal!9}
\strut & \multicolumn{2}{c!{\vrule width 0.8pt}}{\strut\textbf{Reference}} & \multicolumn{2}{c!{\vrule width 0.8pt}}{\strut\textbf{Baselines}} & \multicolumn{3}{c!{\vrule width 0.8pt}}{\strut\textbf{Masking \& steering}} & \multicolumn{2}{c@{}}{\strut\textbf{Decomposed recovery (pp)}} \\
\arrayrulecolor{black!25}\specialrule{0.3pt}{0pt}{0pt}\arrayrulecolor{black}
\strut Model & \full & \shortstack{\sharded{}} & Recap & Mix & \textbf{\shortstack{Reply\\Mask}} & Steering & \textbf{\cleanturn} & \textbf{\shortstack{Mask $-$\\\sharded{}}} & \textbf{\shortstack{\cleanturn $-$\\Reply Mask}} \\
\specialrule{\lightrulewidth}{0pt}{0pt}
\strut LLaVA-Video-7B & 31.3 & \cellcolor{plotamber!23}22.4 & 25.8 & 26.9 & \cellcolor{plotblue!7}24.6 & 31.2 & \cellcolor{plotblue!12}32.7 & \cellcolor{plotblue!11}\textbf{+2.2} [0.3, 3.9] & \cellcolor{plotpurple!22}\textbf{+8.2} [6.5, 10.0] \\
\strut Qwen3-VL-8B & 52.1 & \cellcolor{plotamber!22}44.5 & 46.2 & 47.2 & \cellcolor{plotblue!7}49.7 & 48.6 & \cellcolor{plotblue!12}51.6 & \cellcolor{plotblue!18}\textbf{+5.2} [3.6, 6.8] & \cellcolor{plotpurple!11}\textbf{+1.9} [0.6, 3.2] \\
\strut InternVL3.5-14B & 57.0 & \cellcolor{plotamber!32}40.2 & 49.6 & 53.7 & \cellcolor{plotblue!7}52.8 & 48.2 & \cellcolor{plotblue!12}52.6 & \cellcolor{plotblue!28}\textbf{+12.6} [10.7, 14.6] & $-0.2$ [$-1.1$, 0.7] \\
\strut InternVL3.5-8B & 50.2 & \cellcolor{plotamber!31}34.6 & 48.1 & 46.0 & \cellcolor{plotblue!7}43.6 & 35.3 & \cellcolor{plotblue!12}44.2 & \cellcolor{plotblue!23}\textbf{+9.0} [7.1, 10.9] & +0.6 [$-0.4$, 1.5] \\
\specialrule{\heavyrulewidth}{0pt}{0pt}
\end{tabular}
\end{table*}

%% file: fig/figure5_excalidraw.tex
\begin{figure}[tbp]
  \centering
  \includegraphics[trim=0 1bp 0 1bp,clip,width=0.98\linewidth]{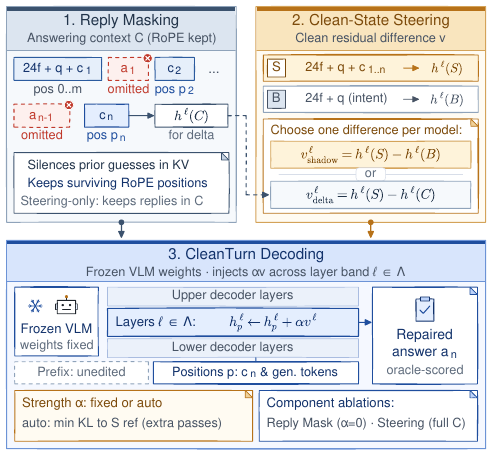}%
    \caption{\textbf{\cleanturn combines Reply Masking and Clean-State Steering.} Earlier reply tokens are excluded while remaining token positions are preserved. A per-query activation difference from a clean prompt is added at selected decoder layers during final-answer generation, without changing model weights.}
  \label{fig:method}
\end{figure}

%% file: sec/8_conclusion.tex
\section{Conclusion}
\label{sec:conclusion}

Even with all video frames continuously visible, revealing clues across turns cuts {\normalfont\full}-normalized retention from $97.0\%$ (\concat) to $75.2\%$ (\sharded), a $21.8$-point conversational drop across 18 models and 10,023 synthetic and real-world queries (6,822 videos).
Controlled history and greying interventions clarify why: retained assistant replies contribute materially to this loss alongside turn-level factors, wrong prior commitments actively bias subsequent answers, and accumulated history reduces behavioural sensitivity to annotated supporting frames until those replies are removed.
Practically, omitting or position-masking prior replies (\emph{Reply Masking}) provides a simple zero-extra-pass primary intervention, while instance-specific \emph{Clean-State Steering} (\cleanturn) serves as a model-dependent extension when turn-fragmentation loss remains (limitations in Supp.~J).

%% file: sec/9_llm_usage.tex
\section*{LLM Usage Statement}
\label{sec:llm_usage}

We used large language and multimodal model assistants (GPT, Claude and Gemini-family models accessed via coding-assistant interfaces), under the authors' direction and with every output reviewed by the authors, to polish and draft text from the authors' outlines, to search for related work and verify references, to discuss design alternatives and help implement and check the experiment, plotting, figure-layout and statistical-analysis code, and to help check derivations and numerical consistency across the main paper and supplementary material.
Within the evaluation pipeline itself, vision-language models were evaluated as experimental subjects (and supplied clean-context reference states and token distributions in \cref{sec:fix}), while GPT-4.1 (\texttt{gpt-4.1-2025-04-14}) was used only to extract structured answers from free-form replies not resolved deterministically prior to rule-based scoring (Supp.~B).
The research questions, method, experiments and scientific claims are the authors' own; every number in the paper traces to records produced by the evaluation harness and was verified by the authors, who take full responsibility for this work.

%% file: sec/X_suppl.tex
\clearpage
\setcounter{page}{1}
\maketitlesupplementary
\renewcommand{\thesection}{\Alph{section}}
\setcounter{section}{0}
\renewcommand{\thetable}{S\arabic{table}}
\setcounter{table}{0}
\renewcommand{\thefigure}{S\arabic{figure}}
\setcounter{figure}{0}
\renewcommand{\thepage}{S\arabic{page}}
\renewcommand{\theHsection}{supp.\Alph{section}}
\renewcommand{\theHtable}{supp.\arabic{table}}
\renewcommand{\theHfigure}{supp.\arabic{figure}}

{\footnotesize
\setcounter{tocdepth}{2}
\makeatletter
\renewcommand*\l@section[2]{%
  \ifnum \c@tocdepth >\z@
    \addpenalty\@secpenalty
    \addvspace{1.5pt \@plus\p@}%
    \setlength\@tempdima{1.4em}%
    \begingroup
      \parindent \z@ \rightskip \@pnumwidth
      \parfillskip -\@pnumwidth
      \leavevmode \bfseries
      \advance\leftskip\@tempdima
      \hskip -\leftskip
      #1\nobreak\hfil \nobreak\hb@xt@\@pnumwidth{\hss #2}\par
    \endgroup
  \fi}
\renewcommand*\l@subsection{\@dottedtocline{2}{1.4em}{2.35em}}
\noindent\textbf{Contents}\par\vspace{1pt}
\@starttoc{toc}
\makeatother
}

\begin{table*}[!b]
\centering\scriptsize\setlength{\tabcolsep}{3.0pt}
\caption{\textbf{24-frame evidence coverage and empirical single-turn answerability across all 10,023 benchmark queries.}
Evidence-frame counts use exact decoded 24-frame timestamps and each task's target-evidence mapping; empirical solvability reports queries answered accurately from those 24 frames by at least one (or at least two) of the 18 models.}
\label{tab:supp_frame_coverage}
\begin{tabular}{@{}lrrrrrrr@{}}
\toprule
Task & $n$ & Step med./max (s) & $|E|\ge 1$ & $|E|$ min/med./mean & Solved $\ge 1$ (\full) & Solved $\ge 2$ (\full) & Solved $\ge 1$ (\full/\concat) \\
\midrule
CLEVRER object & 2,859 & 0.21 / 0.21 & 2,859 (100.0\%) & 3 / 9 / 9.0 & 2,854 (99.8\%) & 2,844 (99.5\%) & 2,857 (99.9\%) \\
CLEVRER timing & 3,811 & 0.21 / 0.21 & 3,811 (100.0\%) & 2 / 8 / 8.3 & 3,808 (99.9\%) & 3,803 (99.8\%) & 3,810 (100.0\%) \\
Charades object & 1,738 & 1.28 / 1.67 & 1,738 (100.0\%) & 1 / 9 / 9.5 & 1,215 (69.9\%) & 1,087 (62.5\%) & 1,281 (73.7\%) \\
Charades grounding & 1,317 & 1.29 / 1.67 & 1,317 (100.0\%) & 1 / 5 / 4.7 & 1,122 (85.2\%) & 895 (68.0\%) & 1,204 (91.4\%) \\
Perception Test object & 298 & 1.01 / 1.46 & 298 (100.0\%) & 1 / 7 / 8.2 & 275 (92.3\%) & 259 (86.9\%) & 280 (94.0\%) \\
\midrule
Total / pooled & 10,023 & \pend & 10,023 (100.0\%) & \pend & 9,274 (92.5\%) & 8,888 (88.7\%) & 9,432 (94.1\%) \\
Five-task unweighted mean & \pend & \pend & 100.0\% & \pend & 89.4\% & 83.3\% & 91.8\% \\
\bottomrule
\end{tabular}
\end{table*}

\input{fig/supp_qual_walkthrough}

\section{Tasks, verification and candidate sets}
\label{sec:supp_datasets}

\subsection{Sampled-frame evidence coverage and single-turn answerability}
To check how well target evidence survives 24-frame subsampling across all 10,023 queries, we evaluate two dataset-wide diagnostics (\cref{tab:supp_frame_coverage}).
First, restricting real-world videos to $\le 40$\,s (median $30.8$\,s on Charades, $24.4$\,s on Perception Test) and actions to $\ge 1.0$\,s bounds the 24-frame stride at $\le 1.67$\,s on Charades (median $1.29$\,s), $\le 1.46$\,s on Perception Test (median $1.01$\,s) and $0.21$\,s on CLEVRER; using exact decoded frame timestamps, all 10,023 queries ($100.0\%$) retain at least one sampled target-evidence frame among the 24 input frames (task means $4.7$ to $9.5$ frames).
Second, across the 18 evaluated models, $94.1\%$ of queries ($9{,}432/10{,}023$) are answered accurately from the exact 24 frames by at least one single-turn route (\full or \concat; $92.5\%$, or $9{,}274/10{,}023$, under \full alone, and $88.7\%$, or $8{,}888/10{,}023$, by at least two models under \full), while greying all 24 frames in the dose study collapses \full accuracy (\eg, to $0.0\text{--}7.9\%$ on Charades and $2.2\text{--}2.8\%$ on CLEVRER object; \cref{tab:supp_dose}), confirming that single-turn answers rely on sampled visual content rather than text priors alone.

\subsection{Task pools, candidate sets and split structure}
The full benchmark pools comprise 2,859 CLEVRER object, 3,811 CLEVRER timing, 1,317 Charades grounding, 1,738 Charades object and 298 Perception Test object queries (10,023 total across 6,822 unique videos).
Their frozen diagnostic slices contain 858, 1,144, 396, 522 and 298 queries, respectively (3,218 total across 2,942 unique videos).
All numerical tables follow that five-task order unless explicitly labeled otherwise (main Table~1 and the dose figure list Charades object before grounding).

CLEVRER~\cite{clevrer2020} queries use annotated objects and collision/visibility events; Charades~\cite{charades2016} queries use annotated actions, objects and temporal spans; and Perception Test~\cite{perceptiontest2023} queries use annotation-generated object constraints verified by an independent constraint checker.
In every query, the intersection of all supplied clues selects a unique target in the scene's annotation-defined candidate set (Perception Test verifies full-set uniqueness without implementing the intermediate per-clue contradiction oracle used on CLEVRER and Charades).
Exploratory pilots on other public video-QA datasets clarify why standard open-ended benchmarks do not cleanly isolate progressive clue accumulation without single-turn floor confounds: NExT-GQA~\cite{nextgqa2024} native questions express single non-decomposable causal facts, AGQA~\cite{agqa2021} scene-graph queries exhibited severe visual class confusions near blind-guess accuracy, and STAR~\cite{star2021} reused shorter Charades clips with predominantly two-clue chains.
Small-scale domain checks on $76$ relation-verified outdoor/social clips from VidOR~\cite{shang2019vidor} and $50$ $640$\,px fisheye egocentric kitchen clips from HD-EPIC~\cite{hdepic2025} illustrate the same single-turn perceptual boundary: open 8B models sit near floor on fine-grained fisheye HD-EPIC ($4\text{--}14\%$ single-turn accuracy) and show wide two-clue intervals on VidOR ($n=76$), whereas higher-ceiling GPT-4.1 exhibits a resolved multi-turn drop on VidOR ($+21.1$\,pp $[+10.5,+31.6]$), a positive point drop on HD-EPIC ($+10.0$\,pp $[0.0,+22.2]$; two-domain mean $+15.5$\,pp $[+8.2,+23.3]$), and resolved \replystripped recovery on both ($+9.2$\,pp $[+2.6,+17.1]$ and $+12.0$\,pp $[+2.0,+22.5]$; mean $+10.6$\,pp $[+4.5,+17.2]$).

Object tasks resolve the extracted object or class against the scene's candidate set; CLEVRER timing requires absolute error $\le 0.5$\,s; and Charades grounding requires temporal $\text{IoU}\ge 0.5$.
Unresolvable descriptions, invalid timestamps and nonanswers score zero in primary accuracy (the stored ``answered'' metadata flag indicates oracle-resolvable commitment rather than mere nonempty text).
Repair tuning uses 150 case-disjoint queries from each of the four larger tasks (no Perception Test tuning).
Within each task, every query comes from a distinct video; across tasks of the same dataset, at most two queries share a video (200 CLEVRER and 76 Charades videos in the 3,218-query diagnostic slice), and 126 diagnostic queries share a video with a tuning query from the paired task of the same dataset (\cref{sec:supp_repair_results} verifies sensitivity after excluding all 126).

\section{Prompt construction and answer extraction}
\label{sec:supp_simuser}
\label{sec:supp_prompts}

\begin{table*}[!t]
\centering\scriptsize\setlength{\tabcolsep}{2.6pt}
\renewcommand{\arraystretch}{1.06}
\caption{\textbf{Prompt lengths, per-turn completion cutoffs and zero-truncation conversational losses across all 18 models ($10{,}023$ queries per model).} \emph{Med./max tok} reports median and maximum final-turn \sharded input tokens (24 frames plus full dialogue history). Cutoff columns count queries hitting \texttt{max\_tokens} in \full (\emph{F}), \concat (\emph{C}), any intermediate turn of \sharded (\emph{S-int}), the final turn (\emph{S-fin}), or any turn (\emph{S-any}). Zero-cutoff columns recompute five-task $\overline C-\overline S$ and \full-normalized $\Delta$ (4,000-draw dataset-stratified video-clustered 95\% CIs) strictly on queries with zero cutoffs in \full, \concat and all turns of \sharded.}
\label{tab:supp_truncation}
\begin{tabular}{@{}lrrrrrrrrcc@{}}
\toprule
Model & Med./max tok & F cut & C cut & S-int & S-fin & S-any & Clean $n$ & All $\Delta$ & Zero-cut $\overline C-\overline S$ (pp) & Zero-cut $\Delta$ (pp) \\
\midrule
InternVL3.5-2B & 6,758 / 12,541 & 10 & 26 & 46 & 8 & 53 & 9,934 & 21.4 & +6.2 [4.6, 7.8] & 21.5 [15.9, 27.1] \\
Qwen3-VL-4B & 4,082 / 7,572 & 1 & 2 & 3 & 1 & 3 & 10,017 & 12.7 & +6.2 [5.0, 7.4] & 12.7 [10.4, 15.1] \\
LLaVA-Video-7B & 5,477 / 6,521 & 0 & 0 & 0 & 0 & 0 & 10,023 & 37.0 & +11.6 [10.2, 13.0] & 37.0 [32.5, 41.4] \\
VideoLLaMA3-7B & 4,951 / 9,732 & 0 & 0 & 0 & 0 & 0 & 10,023 & 20.2 & +7.2 [5.6, 8.8] & 20.2 [15.6, 24.6] \\
Qwen3-VL-8B & 4,079 / 7,572 & 3 & 0 & 6 & 3 & 9 & 10,011 & 21.2 & +11.2 [9.9, 12.4] & 21.2 [18.9, 23.6] \\
InternVL3.5-8B & 6,676 / 12,643 & 0 & 336 & 5 & 1 & 5 & 9,682 & 42.0 & +21.3 [19.9, 22.8] & 43.4 [40.5, 46.5] \\
Qwen3.5-9B & 4,087 / 7,576 & 46 & 487 & 14 & 9 & 22 & 9,480 & 20.8 & +11.3 [9.8, 12.8] & 22.7 [19.7, 25.7] \\
InternVL3.5-14B & 6,648 / 7,140 & 0 & 0 & 0 & 0 & 0 & 10,023 & 30.0 & +17.3 [15.8, 18.7] & 30.0 [27.5, 32.5] \\
Kimi-VL-A3B & 5,667 / 10,312 & 0 & 25 & 0 & 0 & 0 & 9,998 & 28.4 & +9.1 [7.6, 10.6] & 28.7 [23.9, 33.3] \\
Qwen3-VL-30B-A3B & 4,113 / 10,060 & 440 & 281 & 384 & 255 & 445 & 9,071 & 18.2 & +8.7 [7.2, 10.2] & 17.2 [14.4, 20.1] \\
Gemma-4-31B & 6,774 / 7,188 & 0 & 0 & 0 & 0 & 0 & 10,023 & 15.4 & +9.5 [8.3, 10.8] & 15.4 [13.5, 17.4] \\
Qwen3-VL-32B & 4,094 / 10,048 & 149 & 149 & 111 & 31 & 125 & 9,667 & 16.1 & +10.1 [8.9, 11.4] & 16.1 [14.2, 18.1] \\
Qwen3.5-122B-A10B & 4,134 / 8,551 & 58 & 102 & 132 & 69 & 166 & 9,724 & 29.8 & +17.1 [15.5, 18.6] & 29.6 [26.9, 32.2] \\
Qwen3.5-397B-A17B & 4,135 / 8,377 & 21 & 83 & 15 & 1 & 15 & 9,905 & 26.4 & +16.1 [14.5, 17.5] & 26.7 [24.2, 29.1] \\
GPT-4.1 & 2,584 / 4,726 & 0 & 1 & 0 & 0 & 0 & 10,022 & 19.1 & +10.7 [9.5, 12.0] & 19.1 [17.0, 21.2] \\
GPT-5.4 (none) & 4,456 / 6,300 & 0 & 0 & 0 & 0 & 0 & 10,023 & 19.0 & +11.5 [10.3, 12.8] & 19.0 [16.9, 21.0] \\
o3 & 2,442 / 2,591 & 2 & 1 & 4 & 1 & 5 & 10,015 & 6.3 & +3.9 [2.8, 5.0] & 6.3 [4.5, 8.1] \\
Grok-4.3 & 8,089 / 55,915 & 0 & 0 & 0 & 0 & 0 & 10,023 & 8.0 & +5.2 [4.2, 6.2] & 8.0 [6.4, 9.5] \\
\midrule
\emph{Pooled / mean} & \pend & 730 (0.40\%) & 1,493 (0.83\%) & 720 (0.40\%) & 379 (0.21\%) & 848 (0.47\%) & \pend & 21.8 & +10.8 [10.3, 11.2] & 21.9 [21.1, 22.8] \\
\bottomrule
\end{tabular}
\end{table*}

\paragraph{Answer extraction and 100-reply manual verification.}
Every benchmark reply passes through a unified GPT-4.1 structured-extraction stage prior to rule-based task scoring (Main Sec.~3.2 and LLM Usage Statement): short canonical replies ($\le 60$ characters that already match a valid schema form) are parsed deterministically first ($59.05\%$, $59.11\%$ and $59.36\%$ of benchmark outputs under \full, \concat and \sharded, respectively), and all remaining non-empty free-form replies are mapped to the identical schema by GPT-4.1 (\texttt{gpt-4.1-2025-04-14}) before applying the same deterministic task resolver, with pipeline extraction-error rates of $0.04\%$, $0.10\%$ and $0.07\%$ across the three routes.
Neither step ever guesses an absent target; ambiguity, nonanswers, length cutoffs and unresolved outputs remain in the denominator and score zero.
On a seeded 100-reply stratified manual check specifically sampling free-form benchmark outputs processed by GPT-4.1 ($20$ per task: $10$ single-turn and $10$ \sharded across the 18 models, excluding trivial canonical strings already resolved deterministically), $99/100$ ($99.0\%$) yield identical binary correctness to direct inspection of the raw reply ($98/100$ exact candidate/status agreement; the sole score-affecting case is a two-clause temporal reply labeled ambiguous), and restricting evaluation strictly to deterministically parsed replies preserves positive conversational losses on $17$ of $18$ models.

\paragraph{Verbatim clue templates and system instructions.}
All main-benchmark studies deliver stored clue strings verbatim in a fixed case-specific order, without a user-simulator LLM or paraphrasing (\cref{sec:supp_interactive} evaluates response-dependent adaptive clue ordering).
The first user message contains the 24 timestamped frames, the underspecified question and the first clue; each subsequent user message supplies one scheduled clue.
\full joins the question and all ordered clues with spaces into a single user message; \concat presents the question followed by one bullet per clue in the identical order preceded by \emph{``(The lines below are clues about the answer, not answer options.)''}; and \sharded supplies one ordered clue per user turn with the question prepended only to the first turn.
Each task's system prompt is followed by \emph{``The video is shown as chronological frames, each labelled with its time in seconds.''} and the task-specific answer-format instruction, with every sampled frame preceded by its timestamp (\eg, \texttt{1.25s}).
\emph{Recap} appends a final-turn restatement inside the ongoing conversation: \emph{``Just to recapitulate, the entire task is: [full request] Now please complete the task correctly taking all the information into account.''}

\section{Sampling, pairing and statistical summaries}
\label{sec:supp_statistics}

\subsection{Token-budget, truncation and verbosity audit across 18 models}
\label{sec:supp_truncation}
A key methodological question is whether the $1{,}000$-token limit ($16{,}384$ on o3 and Grok-4.3) truncates multi-turn dialogue context in \sharded.
In all experiments, \texttt{max\_tokens} bounds only a single turn's generated completion; the harness sends all 24 timestamped frames and every prior user and assistant message intact at every turn without client-side history dropping.
Across all $180{,}414$ completed \sharded dialogues ($18$ models $\times 10{,}023$ queries), logged final-turn prompt lengths (\cref{tab:supp_truncation}) have medians of $2{,}442$ to $8{,}089$ tokens and a maximum of $12{,}643$ tokens across 17 models ($55{,}915$ on Grok-4.3 under a $\ge 128$k advertised window), sitting well below every model's advertised $\ge 32$k context limit (hosted APIs do not expose internal server-side attention mechanics beyond reported \texttt{prompt\_tokens}).

\Cref{tab:supp_truncation} audits how often generated replies hit the per-turn \texttt{max\_tokens} cap (\texttt{finish\_reason="length"}) in \full, \concat, intermediate turns of \sharded, the final turn of \sharded, and any turn of \sharded.
Pooled across all $180{,}414$ model-query pairs per route, \sharded hits the completion cap on $379$ final turns ($0.21\%$), $720$ dialogues on at least one intermediate turn ($0.40\%$), and $848$ dialogues on any turn ($0.47\%$), fewer than single-turn \concat ($1{,}493/180{,}414$, or $0.83\%$; \full: $730$, or $0.40\%$).
Eight of the 18 models record zero \sharded length cutoffs across all $10{,}023$ queries on every turn, and conditioning strictly on queries with zero cutoffs across \full, \concat and every turn of \sharded keeps $9{,}071$ to $10{,}023$ paired queries per model and leaves all 18 paired video-clustered 95\% CIs strictly above zero ($\Delta=21.9$ $[21.1, 22.8]$ vs.\ $21.8$ $[20.9, 22.6]$ overall).
Partitioning zero-cutoff queries into within-task terciles of prior intermediate reply length (\texttt{turns[:-1]}) likewise yields positive drops across all 18 models in the shortest-reply tercile (18-model mean $+11.6$\,pp vs.\ $+10.0$\,pp in the longest-reply tercile).

\subsection{Backends, rerun stability and video-clustered bootstrap}
Every main-benchmark evaluation uses 24 chronological frames resized so the longest side is at most 640 pixels (\texttt{detail="low"} on image-list APIs), temperature $1.0$ (omitted on o3), and model-default top-$p$/top-$k$ sampling (100-token cap during repair tuning; 1,000-token cap in full evaluation, or 16,384 on o3 and Grok-4.3; 24-token greedy reference in \texttt{auto} KL calibration).
Qwen3-VL-8B, InternVL3.5-8B, InternVL3.5-14B and Kimi-VL-A3B run through vLLM (\texttt{bf16}) in the main benchmark and serving-level delivery study (\cref{sec:supp_delivery}), and through Hugging Face white-box execution when attention masks or hidden-state hooks are required (\cref{sec:supp_ppr,sec:supp_repair_settings,sec:supp_repair_results}); LLaVA-Video-7B uses the LLaVA-NeXT white-box backend throughout, and every white-box intervention is compared strictly against its own matched white-box baseline within the same backend.

\begin{table}[!htbp]
\centering\scriptsize\setlength{\tabcolsep}{5pt}
\caption{\textbf{Subset and unsuffixed regeneration stability.}
Five-task mean accuracy (\%) under the exact unsuffixed main-benchmark \sharded prompt on the full 10,023-query pool, the 3,218-query diagnostic subset, and an independent final-turn regeneration.}
\label{tab:baseline_bridge}
\begin{tabular}{@{}lccc@{}}
\toprule
Model & Full pool & Subset only & Regenerated \\
\midrule
LLaVA-Video-7B  & 17.44 & 17.63 & 18.24 \\
Qwen3-VL-8B    & 40.88 & 41.56 & 40.99 \\
InternVL3.5-8B & 29.25 & 29.23 & 30.10 \\
InternVL3.5-14B & 38.52 & 38.78 & 38.34 \\
Kimi-VL-A3B    & 14.58 & 13.57 & 13.91 \\
GPT-4.1       & 46.26 & 46.90 & 46.36 \\
\bottomrule
\end{tabular}
\end{table}

\Cref{tab:baseline_bridge} shows that under the exact unsuffixed main-benchmark \sharded prompt, restricting from the 10,023-query pool to the 3,218-query diagnostic subset shifts five-task accuracy by $\le 1.0$ point, and independent final-turn regeneration shifts accuracy by $\le 0.9$ points across local and API models ($0.5$ points on GPT-4.1; $0.1$ points on o3 across independent ungreyed \sharded runs, $56.7\%$ vs.\ $56.8\%$).
Separately from rerun variance, recorded-history diagnostic and repair studies (Main Tables~3--4 and \cref{sec:supp_delivery,sec:supp_ppr,sec:supp_repair_results}) append a shared final-answer instruction (\emph{``Now give your best answer to my question, using everything I have told you so far.''}) to both their matched \sharded baseline and all non-\emph{Recap} post-hoc conditions so within-study contrasts hold the final user turn identical; that shared instruction accounts for cross-table baseline differences relative to unsuffixed Main Table~2 and \cref{tab:baseline_bridge} (\eg, $44.6\%$ serving / $44.5\%$ white-box matched \sharded vs.\ $41.0\%$ unsuffixed regeneration on Qwen3-VL-8B).
All paired 95\% confidence intervals use 4,000 dataset-stratified video-clustered bootstrap draws (fixed RNG seed: 0 across main, delivery, repair and interactive suites; 20261113 in the greying archive) that resample unique videos within each dataset, share weights across same-video queries and conditions, recompute task accuracies before equal-task averaging, and take the 2.5th and 97.5th percentiles.

\begin{table*}[!t]
\centering\scriptsize\setlength{\tabcolsep}{2.2pt}
\renewcommand{\arraystretch}{1.07}
\caption{\textbf{Interactive adaptive clarification and progressive streaming video ($150$ balanced queries across five tasks; $30$ per task).} \emph{Adaptive \sharded} selects each next clue to refute wrong intermediate guesses; \emph{Streaming} delivers $24/m$ new chronological frames per turn alongside each clue. Intervals are 4,000-draw dataset-stratified video-clustered 95\% CIs.}
\label{tab:supp_interactive}
\resizebox{\textwidth}{!}{%
\begin{tabular}{@{}lcccccccccc@{}}
\toprule
& & \multicolumn{4}{c}{Adaptive response-dependent clue selection} & \multicolumn{5}{c}{Progressive streaming video frames across turns} \\
\cmidrule(lr){3-6}\cmidrule(l){7-11}
Model & \full & Fixed \sharded & Adapt.\ \sharded & Adapt.\ \replystripped & Adapt.\ $\replystripped - \sharded$ & Stream \full & Stream \sharded & Stream \replystripped & Stream $\full - \sharded$ & Stream $\replystripped - \sharded$ \\
\midrule
Qwen3-VL-8B & 52.0 & 43.3 & 41.3 & 48.0 & +6.7 [+0.3, +13.1] & 50.0 & 27.3 & 46.0 & +22.7 [+14.9, +30.5] & +18.7 [+11.8, +25.8] \\
InternVL3.5-8B & 48.7 & 28.7 & 25.3 & 40.0 & +14.7 [+6.1, +23.2] & 51.3 & 22.0 & 32.7 & +29.3 [+20.9, +37.4] & +10.7 [+2.7, +18.9] \\
GPT-4.1 & 54.7 & 46.7 & 44.7 & 56.7 & +12.0 [+5.5, +18.7] & 49.3 & 35.3 & 45.3 & +14.0 [+6.5, +21.2] & +10.0 [+2.6, +17.3] \\
\bottomrule
\end{tabular}%
}
\end{table*}

\section{Benchmark reproducibility, confidence intervals and robustness}
\label{sec:supp_scale}

\begin{table*}[!b]
\centering\scriptsize\setlength{\tabcolsep}{3.8pt}
\renewcommand{\arraystretch}{1.06}
\caption{\textbf{Effect of temporal frame-sampling density ($12$, $24$ and $48$ frames) across $150$ balanced five-task queries.} Denser visual evidence raises single-turn ceilings without removing the multi-turn conversational drop.}
\label{tab:supp_frame_density}
\begin{tabular}{@{}llccccccc@{}}
\toprule
Model & Sampled frames & \full (\%) & \concat (\%) & \sharded (\%) & \replystripped (\%) & $\overline C-\overline S$ (pp) & \full-norm.\ $\Delta$ (pp) & $\replystripped - \sharded$ (pp) \\
\midrule
Qwen3-VL-8B & 12 frames & 45.3 & 45.3 & 39.3 & 43.3 & +6.0 [$-1.0$, +13.1] & 13.2 & +4.0 [$-3.1$, +11.3] \\
Qwen3-VL-8B & 24 frames & 52.0 & 51.3 & 43.3 & 48.7 & +8.0 [+1.5, +15.0] & 15.4 & +5.3 [$-1.5$, +12.3] \\
Qwen3-VL-8B & 48 frames & 50.7 & 51.3 & 46.0 & 49.3 & +5.3 [$-0.8$, +11.6] & 10.5 & +3.3 [$-2.4$, +8.9] \\
\midrule
InternVL3.5-8B & 12 frames & 44.7 & 48.7 & 28.7 & 42.0 & +20.0 [+11.9, +28.3] & 44.8 & +13.3 [+5.1, +21.5] \\
InternVL3.5-8B & 24 frames & 48.7 & 50.7 & 28.7 & 38.7 & +22.0 [+14.9, +29.1] & 45.2 & +10.0 [+2.5, +17.2] \\
InternVL3.5-8B & 48 frames & 50.0 & 50.7 & 30.0 & 39.3 & +20.7 [+12.2, +29.1] & 41.3 & +9.3 [+0.7, +18.3] \\
\midrule
GPT-4.1 & 12 frames & 54.7 & 59.3 & 48.0 & 57.3 & +11.3 [+3.5, +19.1] & 20.7 & +9.3 [+1.9, +16.7] \\
GPT-4.1 & 24 frames & 54.7 & 54.7 & 46.7 & 51.3 & +8.0 [+0.5, +15.4] & 14.6 & +4.7 [$-2.9$, +12.0] \\
GPT-4.1 & 48 frames & 56.0 & 62.7 & 48.7 & 57.3 & +14.0 [+7.7, +20.2] & 25.0 & +8.7 [+2.4, +15.1] \\
\bottomrule
\end{tabular}
\end{table*}

\subsection{Adaptive clarification and progressive streaming video}
\label{sec:supp_interactive}
Our main benchmark uses a fixed clue order with all 24 frames visible from Turn~1 to isolate delivery structure without confounding changes in evidence.
On a balanced 150-query five-task slice (first 30 rank-ordered diagnostic queries per task across Qwen3-VL-8B, InternVL3.5-8B and GPT-4.1; \cref{tab:supp_interactive}), we test two more realistic interactive regimes.

\paragraph{Response-dependent adaptive clarification.}
In \emph{Adaptive \sharded}, after each assistant turn $t$, the user extracts the model's current commitment and deterministically selects the next unrevealed true clue from the query's verified pool that \emph{refutes that wrong candidate} under the task's constraint oracle (falling back to maximum candidate elimination when no wrong candidate was named; $24.2\%$ to $26.7\%$ of CLEVRER and Charades dialogues alter clue order relative to the seeded schedule).
Even when subsequent clues are actively chosen to falsify wrong intermediate guesses, \emph{Adaptive \sharded} still degrades substantially below \full across all three models ($-10.0$ to $-23.3$\,pp; all 95\% CIs excluding zero), whereas omitting the model's prior replies on the identical adaptive clue sequence (\emph{Adaptive \replystripped}) recovers $+6.7$ to $+14.7$\,pp (all three 95\% CIs above zero).

\paragraph{Progressive streaming video across turns.}
In \emph{Streaming \sharded}, the 24 chronological timestamped frames are partitioned into $m$ consecutive temporal segments delivered progressively alongside each turn's clue, so all 24 frames are in context by the final turn.
When clues and video segments both stream across turns while intermediate replies remain in context, accuracy drops sharply relative to scripted-wait \emph{Streaming \full} ($-14.0$ to $-29.3$\,pp across all three models; all 95\% CIs excluding zero), because early turns form commitments from incomplete visual spans as well as incomplete clues; omitting those intermediate assistant replies at the final turn (\emph{Streaming \replystripped}) recovers $+10.0$ to $+18.7$\,pp across all three models.

\subsection{Temporal frame-sampling density: 12 vs.\ 24 vs.\ 48 frames}
\label{sec:supp_density}
\begin{table*}[!t]
\centering\scriptsize\setlength{\tabcolsep}{2.4pt}
\renewcommand{\arraystretch}{1.07}
\caption{\textbf{Fine-grained conversational degradation and history-removal recovery by reasoning category and clue-constraint primitive.} Task-type rows weight constituent tasks equally; primitive rows pool queries containing each primitive per model before averaging across 18 models. Right column reports the model range of \replystripped ($6$ models) and white-box Reply Masking ($5$ open models) gains on the $3{,}218$-query diagnostic set.}
\label{tab:supp_reasoning_taxonomy}
\resizebox{\textwidth}{!}{%
\begin{tabular}{@{}lrrcccccc@{}}
\toprule
Reasoning category / clue primitive & Queries $n$ & Models $>0$ & \full (\%) & \concat (\%) & \sharded (\%) & $\overline C-\overline S$ (pp) & $\Delta$ (pp) & \replystripped / Mask gain range (pp) \\
\midrule
\emph{Task reasoning type} & & & & & & & & \\
Entity / object identification (3 tasks) & 4,895 & 18/18 & 55.8 & 54.4 & 44.3 & +10.2 & 18.8 & $+4.2\text{ to }+14.5$ / $+0.1\text{ to }+15.6$ \\
Instantaneous event timing (CLEVRER) & 3,811 & 17/18 & 66.8 & 66.3 & 49.9 & +16.4 & 27.1 & $-1.7\text{ to }+23.2$ / $-2.2\text{ to }+19.2$ \\
Temporal interval localization (Charades) & 1,317 & 17/18 & 21.1 & 20.5 & 14.0 & +6.6 & 29.0 & $-1.8\text{ to }+8.1$ / $-0.5\text{ to }+4.8$ \\
\midrule
\emph{Clue constraint primitives present} & & & & & & & & \\
Relational temporal order (\texttt{before}/\texttt{after}) & 8,514 & 18/18 & 55.4 & 54.7 & 42.8 & +11.9 & 22.5 & \pend \\
Visibility / scene-entry window & 3,773 & 18/18 & 74.2 & 73.2 & 58.5 & +14.7 & 20.6 & \pend \\
Multi-event counting (\texttt{exactly $k$}) & 2,523 & 18/18 & 71.5 & 70.1 & 57.9 & +12.2 & 17.2 & \pend \\
Visual attribute / action verb only & 288 & 18/18 & 62.9 & 60.6 & 45.2 & +15.5 & 25.9 & \pend \\
\bottomrule
\end{tabular}%
}
\end{table*}
\Cref{tab:supp_frame_density} tests whether providing denser visual sampling helps models override incorrect intermediate commitments across the 150-query five-task slice ($12$, $24$ and $48$ uniformly sampled timestamped frames).
Increasing frame density from $12$ to $48$ frames raises single-turn \full accuracy ($45.3\%\rightarrow 50.7\%$ on Qwen3-VL-8B, $44.7\%\rightarrow 50.0\%$ on InternVL3.5-8B and $54.7\%\rightarrow 56.0\%$ on GPT-4.1), yet \emph{does not eliminate conversational degradation}: at $48$ frames, InternVL3.5-8B still loses $20.7$\,pp ($\Delta=41.3\%$) and GPT-4.1 loses $14.0$\,pp ($\Delta=25.0\%$), with \replystripped recovering $+9.3$\,pp and $+8.7$\,pp, respectively.
Denser visual evidence improves single-turn perception but does not by itself dislodge self-generated history commitments.

\subsection{Fine-grained breakdown by reasoning type and clue primitives}
\label{sec:supp_reasoning}
\begin{table*}[!b]
\centering\scriptsize\setlength{\tabcolsep}{4pt}
\renewcommand{\arraystretch}{1.07}
\caption{\textbf{Real-world versus synthetic video verification across all four study stages.} Real-world tasks combine Charades object, Charades grounding and Perception Test object (3,353 benchmark queries across 2,464 videos; 1,216 diagnostic queries); synthetic tasks combine CLEVRER object and timing (6,670 queries across 4,358 videos; 2,002 diagnostic queries). Every qualitative finding holds on real-world videos alone.}
\label{tab:supp_realworld}
\begin{tabular}{@{}lllcc@{}}
\toprule
Stage & Metric & Models & Real-world tasks (3 tasks) & Synthetic CLEVRER (2 tasks) \\
\midrule
1. Delivery drop & Mean $C/F$ / $S/F$ retention (\%) & 18 models & 96.5 [95.3, 97.8] / 74.4 [72.9, 76.0] & 97.8 [97.3, 98.3] / 76.4 [75.8, 76.9] \\
1. Delivery drop & Mean \full-normalized $\Delta$ (pp) & 18 models & 22.1 [20.4, 23.7] (18/18 $>0$) & 21.4 [20.8, 22.0] (18/18 $>0$) \\
2. History edits & \replystripped $-$ \sharded range (pp) & 6 models & $+3.0$ to $+9.2$ (6/6 $>0$) & $+0.7$ to $+16.6$ (6/6 $>0$) \\
2. History edits & Reply Mask $-$ \sharded range (pp) & 5 open models & $+0.3$ to $+10.1$ (5/5 $>0$) & $+0.2$ to $+16.3$ (5/5 $>0$) \\
3. Frame greying & Mean $R$ in \full / \replystripped / \sharded (pp) & 7 models & 14.3 / 14.1 / 6.0 (7/7 recover) & 16.0 / 13.0 / 5.2 (7/7 recover) \\
4. Repair & Reply Mask / \cleanturn gain over \sharded (pp) & 4 open models & $+2.0$--$+10.1$ / $+2.1$--$+10.3$ & $+0.2$--$+16.3$ / $+7.5$--$+16.0$ \\
\bottomrule
\end{tabular}
\end{table*}

\begin{table*}[!b]
\centering\scriptsize\setlength{\tabcolsep}{4.5pt}
\renewcommand{\arraystretch}{1.07}
\caption{\textbf{Clue-order sensitivity across three independent seeded permutations} (\texttt{rep} $=0,1,2$). Evaluated on 250 balanced queries (first 50 rank-ordered queries per task; 249 unique videos, 2,250 evaluations per model). Intervals are paired dataset-stratified video-bootstrap 95\% CIs (4,000 draws). Conversational degradation remains positive across every tested order.}
\label{tab:supp_clue_order}
\begin{tabular}{@{}llccccc@{}}
\toprule
Model & Clue order & \full (\%) & \concat (\%) & \sharded (\%) & $\overline C-\overline S$ (pp) & $\Delta$ (pp) \\
\midrule
Qwen3-VL-8B & Order 0 (main benchmark order) & 54.0 & 52.0 & 44.4 & +7.6 [2.4, 12.8] & 14.1 [4.6, 23.4] \\
Qwen3-VL-8B & Order 1 (second random shuffle) & 54.0 & 54.4 & 41.2 & +13.2 [7.5, 19.0] & 24.4 [14.1, 34.6] \\
Qwen3-VL-8B & Order 2 (third random shuffle) & 52.4 & 51.6 & 46.0 & +5.6 [0.3, 10.7] & 10.7 [0.5, 20.3] \\
Qwen3-VL-8B & \emph{Three-order mean} & \emph{53.5} & \emph{52.7} & \emph{43.9} & \emph{+8.8 [4.9, 12.7]} & \emph{16.5 [9.5, 23.3]} \\
\midrule
InternVL3.5-8B & Order 0 (main benchmark order) & 49.6 & 50.0 & 31.2 & +18.8 [13.1, 24.6] & 37.9 [26.7, 49.1] \\
InternVL3.5-8B & Order 1 (second random shuffle) & 50.4 & 50.8 & 28.0 & +22.8 [16.5, 29.3] & 45.2 [33.0, 58.4] \\
InternVL3.5-8B & Order 2 (third random shuffle) & 50.8 & 50.8 & 31.2 & +19.6 [13.2, 25.9] & 38.6 [26.0, 51.7] \\
InternVL3.5-8B & \emph{Three-order mean} & \emph{50.3} & \emph{50.5} & \emph{30.1} & \emph{+20.4 [16.3, 24.3]} & \emph{40.6 [33.2, 48.2]} \\
\bottomrule
\end{tabular}
\end{table*}
\Cref{tab:supp_reasoning_taxonomy} stratifies conversational degradation across all 18 models ($10{,}023$ queries) and reply-removal recovery across the six diagnostic models ($3{,}218$ queries) along two descriptive axes: (i)~task output reasoning type (equal-task mean across the three object-identification tasks vs.\ instantaneous event timing and temporal interval localization), and (ii)~query-pooled subsets containing each clue-constraint primitive (\emph{Relational temporal ordering} such as \texttt{before}/\texttt{after}, \emph{Multi-event counting}, \emph{Visibility/scene-entry windows} and \emph{Attribute/action-verb matching only}, where $\Delta$ averages model-level \full-normalized drops).
Descriptively, instantaneous event timing ($\overline C-\overline S=+16.4$\,pp, $\Delta=27.1\%$) and temporal interval localization ($\Delta=29.0\%$) exhibit larger relative drops than object identification ($\Delta=18.8\%$), while removing prior replies recovers up to $+23.2$\,pp (\replystripped) and $+19.2$\,pp (Reply Masking) on event timing (InternVL3.5-8B) and $+8.1$\,pp on interval localization (GPT-4.1), and all four clue-primitive subsets degrade across all $18/18$ models.

\subsection{Real-world split, clue permutations, main CIs and model registry}
\begin{table*}[!htbp]
\centering\scriptsize\setlength{\tabcolsep}{3.5pt}
\renewcommand{\arraystretch}{1.06}
\caption{\textbf{Main-benchmark conversational losses with paired video-clustered 95\% CIs.} Entries report $\concat-\sharded$ accuracy drops (pp) per task and for the unweighted five-task mean ($\overline C-\overline S$), plus \full{}-normalized $\Delta=100(\overline C-\overline S)/\overline F$ (pp), across 10,023 queries (6,822 unique videos; 4,000 dataset-stratified video-bootstrap resamples).}
\label{tab:supp_main_ci}
\begin{tabular}{@{}lccccccc@{}}
\toprule
Model & CLEVRER Obj & CLEVRER Tim & Charades Obj & Charades Grd & Perception Obj & $\overline C-\overline S$ (pp) & $\Delta$ (pp) \\
\midrule
InternVL3.5-2B & +7.5 [5.3, 9.6] & +4.2 [2.3, 6.0] & +5.8 [3.6, 7.9] & +1.9 [0.2, 3.6] & +11.4 [4.7, 18.1] & +6.1 [4.5, 7.8] & 21.4 [15.9, 26.9] \\
Qwen3-VL-4B & +10.2 [8.6, 11.8] & +10.2 [8.6, 11.7] & $-0.3$ [$-2.1$, 1.3] & +7.4 [5.1, 9.8] & +3.4 [$-0.7$, 8.1] & +6.2 [5.0, 7.4] & 12.7 [10.4, 15.2] \\
LLaVA-Video-7B & +19.0 [16.8, 21.1] & +13.3 [11.6, 14.9] & +3.3 [1.1, 5.6] & $-0.2$ [$-1.0$, 0.5] & +22.8 [16.4, 29.2] & +11.6 [10.2, 13.0] & 37.0 [32.5, 41.4] \\
VideoLLaMA3-7B & +0.5 [$-2.0$, 2.9] & +13.2 [11.2, 15.1] & +2.9 [0.6, 5.1] & +4.9 [2.5, 7.3] & +14.8 [8.4, 21.1] & +7.2 [5.6, 8.8] & 20.2 [15.6, 24.6] \\
Qwen3-VL-8B & +11.4 [9.7, 13.1] & +20.7 [19.1, 22.4] & +2.0 [0.4, 3.5] & +7.7 [5.2, 10.4] & +14.1 [9.1, 19.1] & +11.2 [9.9, 12.4] & 21.2 [18.9, 23.6] \\
InternVL3.5-8B & +28.1 [26.1, 30.1] & +48.3 [46.4, 50.1] & +6.9 [4.7, 9.2] & +5.0 [2.6, 7.4] & +14.8 [9.1, 20.5] & +20.6 [19.2, 22.1] & 42.0 [39.2, 45.0] \\
Qwen3.5-9B & +8.2 [6.0, 10.4] & +19.2 [17.1, 21.3] & +2.5 [0.3, 4.8] & +7.4 [4.5, 10.4] & +14.1 [7.7, 20.1] & +10.3 [8.8, 11.8] & 20.8 [17.7, 23.8] \\
InternVL3.5-14B & +22.5 [20.4, 24.6] & +29.5 [27.8, 31.1] & +2.9 [0.8, 5.1] & +9.0 [6.4, 11.4] & +22.5 [16.8, 28.5] & +17.3 [15.8, 18.7] & 30.0 [27.5, 32.5] \\
Kimi-VL-A3B & +3.0 [0.7, 5.2] & +15.0 [13.0, 16.8] & +11.7 [9.7, 13.7] & +2.5 [1.1, 4.0] & +13.1 [6.7, 19.1] & +9.1 [7.5, 10.5] & 28.4 [23.6, 33.0] \\
Qwen3-VL-30B-A3B & +11.3 [9.4, 13.1] & +10.4 [8.6, 12.2] & +6.8 [4.8, 8.9] & +6.4 [3.7, 9.1] & +9.4 [3.7, 15.1] & +8.9 [7.4, 10.3] & 18.2 [15.3, 21.1] \\
Gemma-4-31B & +13.4 [11.8, 15.1] & +9.2 [8.1, 10.2] & +4.3 [2.6, 6.0] & +6.0 [3.6, 8.3] & +14.8 [9.7, 19.8] & +9.5 [8.3, 10.8] & 15.4 [13.5, 17.4] \\
Qwen3-VL-32B & +9.3 [7.9, 10.7] & +9.3 [7.9, 10.6] & +4.5 [2.7, 6.3] & +9.5 [7.0, 12.0] & +17.4 [12.4, 22.5] & +10.0 [8.8, 11.3] & 16.1 [14.2, 18.1] \\
Qwen3.5-122B-A10B & +11.9 [9.9, 13.9] & +32.9 [31.1, 34.8] & +10.3 [8.1, 12.5] & +6.5 [3.6, 9.5] & +23.5 [17.1, 29.5] & +17.0 [15.4, 18.6] & 29.8 [27.1, 32.4] \\
Qwen3.5-397B-A17B & +20.4 [18.5, 22.2] & +26.9 [25.1, 28.6] & +7.3 [5.1, 9.5] & +7.0 [3.8, 10.2] & +17.8 [11.7, 23.8] & +15.9 [14.3, 17.3] & 26.4 [23.9, 28.9] \\
GPT-4.1 & +15.1 [13.1, 17.1] & +12.1 [10.7, 13.6] & +5.2 [3.4, 7.0] & +8.5 [6.1, 10.9] & +12.8 [8.0, 17.4] & +10.7 [9.5, 12.0] & 19.1 [17.0, 21.2] \\
GPT-5.4 (none) & +19.9 [18.1, 21.7] & +8.9 [7.9, 9.9] & +10.9 [8.6, 13.3] & +9.0 [6.6, 11.3] & +9.1 [4.0, 14.4] & +11.5 [10.3, 12.8] & 19.0 [16.9, 21.0] \\
o3 & +1.5 [0.04, 3.1] & $-0.1$ [$-1.4$, 1.1] & +3.0 [1.2, 4.9] & +12.9 [10.2, 15.5] & +2.0 [$-1.7$, 6.0] & +3.9 [2.8, 5.0] & 6.3 [4.5, 8.1] \\
Grok-4.3 & +4.4 [3.3, 5.6] & +11.4 [10.1, 12.7] & +0.7 [$-0.9$, 2.4] & +6.6 [4.1, 9.2] & +2.7 [$-1.0$, 6.4] & +5.2 [4.2, 6.2] & 8.0 [6.4, 9.5] \\
\midrule
\emph{Mean (18 models)} & +12.1 [11.6, 12.7] & +16.4 [15.9, 16.8] & +5.0 [4.4, 5.7] & +6.6 [5.8, 7.3] & +13.3 [11.5, 15.2] & +10.7 [10.2, 11.1] & 21.8 [20.9, 22.6] \\
\bottomrule
\end{tabular}
\end{table*}

\begin{table*}[!htbp]
\centering\scriptsize\setlength{\tabcolsep}{3.5pt}
\renewcommand{\arraystretch}{1.06}
\caption{\textbf{Model registry, pinned revisions and decoding settings for the 18-model benchmark.} All runs use 24 timestamped frames ($\le 640$\,px; API \texttt{detail="low"}), temperature $1.0$ (omitted by the o3 API) and model-default top-$p$/top-$k$ sampling, evaluated Nov 3--9, 2026 (follow-up diagnostics Nov 8--23, 2026).}
\label{tab:supp_repro_models}
\begin{tabular}{@{}llllrr@{}}
\toprule
Model & Exact identifier / API snapshot & Pinned HF commit & Backend & Max tokens & Notes \\
\midrule
InternVL3.5-2B & \texttt{OpenGVLab/InternVL3\_5-2B-HF} & \texttt{3f301ffcf3} & vLLM (bf16) & 1,000 & 1 tile/frame \\
Qwen3-VL-4B & \texttt{Qwen/Qwen3-VL-4B-Instruct} & \texttt{ebb281ec70} & vLLM (bf16) & 1,000 & all-local rerun \\
LLaVA-Video-7B & \texttt{lmms-lab/LLaVA-Video-7B-Qwen2} & \texttt{013210b3af} & HF (fp16/bf16) & 1,000 & LLaVA-NeXT adapter \\
VideoLLaMA3-7B & \texttt{DAMO-NLP-SG/VideoLLaMA3-7B} & \texttt{d5b763e368} & HF (fp16/bf16) & 1,000 & native adapter \\
Qwen3-VL-8B & \texttt{Qwen/Qwen3-VL-8B-Instruct} & \texttt{0c351dd01e} & vLLM (bf16) & 1,000 & core diagnostic model \\
InternVL3.5-8B & \texttt{OpenGVLab/InternVL3\_5-8B-HF} & \texttt{741a7d0302} & vLLM (bf16) & 1,000 & 1 tile/frame \\
Qwen3.5-9B & \texttt{Qwen/Qwen3.5-9B} & \texttt{c202236235} & vLLM (bf16) & 1,000 & thinking off \\
InternVL3.5-14B & \texttt{OpenGVLab/InternVL3\_5-14B-HF} & \texttt{226b96d591} & vLLM (bf16) & 1,000 & 1 tile/frame \\
Kimi-VL-A3B & \texttt{moonshotai/Kimi-VL-A3B-Instruct} & \texttt{398eede090} & vLLM (bf16) & 1,000 & 16B total MoE \\
Qwen3-VL-30B-A3B & \texttt{Qwen/Qwen3-VL-30B-A3B-Instruct} & \texttt{9c4b90e1e4} & vLLM (bf16) & 1,000 & 30B total MoE \\
Gemma-4-31B & \texttt{google/gemma-4-31B-it} & API-hosted & API & 1,000 & Nov 5, 2026 \\
Qwen3-VL-32B & \texttt{Qwen/Qwen3-VL-32B-Instruct} & \texttt{0cfaf48183} & vLLM (bf16) & 1,000 & dense 32B \\
Qwen3.5-122B-A10B & \texttt{Qwen/Qwen3.5-122B-A10B} & API-hosted & API & 1,000 & thinking off, Nov 5 \\
Qwen3.5-397B-A17B & \texttt{Qwen/Qwen3.5-397B-A17B-GPTQ-Int4} & API-hosted & API (Int4) & 1,000 & thinking off, Nov 6 \\
GPT-4.1 & \texttt{gpt-4.1\_2025-04-14} & API snapshot & API & 1,000 & Nov 5, 2026 \\
GPT-5.4 (none) & \texttt{gpt-5.4\_2026-03-05} & API snapshot & API & 1,000 & reasoning=none, Nov 9 \\
o3 & \texttt{o3\_2025-04-16} & API snapshot & API & 16,384 & temp omitted, Nov 5 \\
Grok-4.3 & \texttt{grok-4.3\_1} & API snapshot & API & 16,384 & reasoning on, Nov 6 \\
\bottomrule
\end{tabular}
\end{table*}
To verify that our benchmark conclusions hold on real-world indoor videos and across clue orders on the full 250-query permutation subset, \cref{tab:supp_realworld} separates the three real-world tasks ($3{,}353$ queries across $2{,}464$ Charades and Perception Test videos) from synthetic CLEVRER across all four stages ($\Delta=22.1$ vs.\ $21.4$\,pp, $18/18$ models $>0$ on both), and \cref{tab:supp_clue_order} confirms positive drops across three independent seeded clue permutations (\texttt{rep} $=0,1,2$).
\Cref{tab:supp_main_ci} provides paired video-clustered 95\% CIs for all 90 model-task cells in Main Table~2 ($82/90$ cells and $18/18$ five-task means strictly above zero), and \cref{tab:supp_repro_models} lists pinned Hugging Face commits, API snapshots and Nov 2026 evaluation dates (two exploratory models were excluded prior to freezing the 18-model table: floor-level Qwen3-VL-2B and hedged multi-timestamp Kimi-K2.5).

\section{Delivery and reply-content interventions}
\label{sec:supp_delivery}

\begin{table*}[!htbp]
\centering\scriptsize\setlength{\tabcolsep}{4.2pt}
\renewcommand{\arraystretch}{1.07}
\caption{\textbf{Intermediate-turn error accumulation and streak lock-in across all 18 models.} Left: turn-by-turn progression averaged across 18 models on 7-turn CLEVRER dialogues. Right: next-turn revision rate by consecutive wrong-commitment streak length, and final \sharded accuracy (conditioned on single-turn \full success) by total prior wrong turns across all five tasks.}
\label{tab:supp_turn_accum}
\begin{tabular}{@{}lcccc@{\hspace{14pt}}lcc@{}}
\toprule
\multicolumn{5}{c}{7-turn progression (18-model mean, \%)} & \multicolumn{3}{c}{Progressive lock-in across streak length (18-model mean)} \\
\cmidrule(r){1-5}\cmidrule(l){6-8}
Turn $t$ & Committed & Violates clues $1..t$ & Repeat prev.\ wrong & Target acc. & Wrong-streak / count $s$ & Revise at $t+1$ (\%) & Final acc.\ $|\;\full=1$ (\%) \\
\midrule
Turn 1 & 87.2 & 14.2 & \pend & 40.0 & $s = 0$ prior wrong turns & \pend & 75.3 \\
Turn 2 & 82.3 & 26.1 & 21.7 & 40.6 & $s = 1$ wrong turn & 61.5 & 58.8 \\
Turn 3 & 82.3 & 29.6 & 20.1 & 44.8 & $s = 2$ streak / count & 44.4 & 60.6 \\
Turn 4 & 82.5 & 31.1 & 19.0 & 47.7 & $s = 3$ streak / count & 32.9 & 54.4 \\
Turn 5 & 82.6 & 31.3 & 17.9 & 50.0 & $s \ge 4$ streak / count & 23.5 & 34.6 \\
Turn 6 & 82.7 & 30.7 & 17.2 & 51.7 & & & \\
Turn 7 (Final) & 82.7 & 29.5 & 16.4 & 53.2 & & & \\
\bottomrule
\end{tabular}
\end{table*}

\subsection{Intermediate-turn error accumulation and progressive lock-in}
\label{sec:supp_accumulation}
Scoring every intermediate turn across all 18 models on the $10{,}023$-query benchmark reveals how errors accumulate across turns (\cref{tab:supp_turn_accum}).
First, models commit prematurely: on 7-turn CLEVRER dialogues ($3\text{ to }10$ scene candidates, median $5$), $87.2\%$ of Turn~1 replies already name a specific candidate while multiple candidates remain consistent with the first clue.
As additional clues narrow the feasible set across Turns~2 to 7, the share of replies that \emph{contradict an already-revealed clue} rises from $14.2\%$ at Turn~1 to $31.3\%$ by Turn~5 and remains at $29.5\%$ at Turn~7, with $16.4\%$ to $21.7\%$ of all dialogues repeating the exact wrong prediction from the immediately preceding turn.
Second, observationally across turns (reflecting both progressive self-anchoring and query-difficulty selection, which we disentangle via controlled history edits below), longer wrong-commitment streaks coincide with lower next-turn revision rates: across all 18 models, the empirical probability of abandoning a wrong candidate drops monotonically from $61.5\%$ after stating it once, to $44.4\%$ after a two-turn streak, $32.9\%$ after a three-turn streak, and $23.5\%$ after four or more consecutive turns.
Conditioned on queries the same model solves under single-turn \full, final \sharded accuracy falls from $75.3\%$ when zero intermediate turns committed to a wrong answer down to $34.6\%$ when four or more intermediate turns did so.

\subsection{Resistance to contradictory clues and explicit user corrections}
\label{sec:supp_contradictions}
\begin{table*}[!htbp]
\centering\scriptsize\setlength{\tabcolsep}{3.3pt}
\renewcommand{\arraystretch}{1.06}
\caption{\textbf{Model resistance to newly contradictory clues ($9{,}725$ CLEVRER/Charades queries) and explicit user corrections ($150$ serving / $100$ white-box five-task queries).} Left: exact transitions where Clue~$t$ newly rules out Turn-$(t-1)$'s previously consistent answer (representative models across families; all 18 follow the same open-vs.-frontier gradient). Right: explicit oracle-assisted user negation of wrong prior answers versus zero-oracle reply removal/masking.}
\label{tab:supp_contradiction_corr}
\begin{tabular}{@{}lrrrr@{\hspace{10pt}}lccccc@{}}
\toprule
\multicolumn{5}{c}{Natural contradiction transitions ($t-1 \rightarrow t$)} & \multicolumn{6}{c}{Explicit user correction vs.\ reply removal / masking} \\
\cmidrule(r){1-5}\cmidrule(l){6-11}
Model & Events & Repeat refuted & Valid rev. & Final ok & Model & \sharded & Final corr. & Online corr. & \replystripped / Mask & Corr.\ $-$ \sharded \\
\midrule
Qwen3-VL-30B-A3B & 3,420 & 63.2\% & 14.8\% & 17.6\% & Qwen3-VL-8B (serv.) & 43.3 & 46.0 & 46.7 & 48.7 (\replystripped) & +2.7 [$-1.3$, +6.8] \\
VideoLLaMA3-7B & 2,524 & 60.3\% & 13.7\% & 14.4\% & InternVL3.5-8B (serv.) & 28.7 & 40.0 & 33.3 & 38.7 (\replystripped) & +11.3 [+6.3, +16.9] \\
LLaVA-Video-7B & 1,894 & 56.9\% & 7.7\% & 6.4\% & GPT-4.1 (API) & 46.7 & 49.3 & 52.7 & 51.3 (\replystripped) & +2.7 [$-1.8$, +7.4] \\
Qwen3-VL-8B & 3,435 & 52.5\% & 22.4\% & 27.0\% & LLaVA-Video-7B (WB) & 15.0 & 16.0 & \pend & 37.0 (Mask) & +1.0 [0.0, +3.3] \\
InternVL3.5-8B & 2,465 & 31.0\% & 29.3\% & 28.3\% & & & & & & \\
InternVL3.5-14B & 3,088 & 30.4\% & 38.2\% & 37.2\% & & & & & & \\
Gemma-4-31B & 3,750 & 30.5\% & 44.2\% & 52.3\% & & & & & & \\
GPT-4.1 & 3,615 & 29.4\% & 38.9\% & 44.1\% & & & & & & \\
o3 & 3,681 & 22.4\% & 52.1\% & 61.9\% & & & & & & \\
Grok-4.3 & 3,750 & 20.7\% & 61.7\% & 68.7\% & & & & & & \\
GPT-5.4 (none) & 3,724 & 14.7\% & 53.8\% & 60.5\% & & & & & & \\
\bottomrule
\end{tabular}
\end{table*}
To quantify how resistant different models are to revising their own previous answers, \cref{tab:supp_contradiction_corr} (left) uses the exact per-clue constraint oracles on CLEVRER and Charades ($9{,}725$ queries per model) to isolate every turn transition $(t-1\rightarrow t)$ where a model's Turn-$(t-1)$ commitment was consistent with clues $1\dots t-1$ and is \emph{newly contradicted} by Clue~$t$.
Open-weight models stubbornly repeat the exact same refuted answer on $29.0\%$ to $65.0\%$ of newly contradicted turns across the 14 open models (\eg, $65.0\%$ on Qwen3-VL-4B, $63.2\%$ on Qwen3-VL-30B-A3B, $60.3\%$ on VideoLLaMA3-7B, $56.9\%$ on LLaVA-Video-7B and $52.5\%$ on Qwen3-VL-8B, with valid revision spanning $7.7\text{--}44.2\%$), whereas the four frontier API models repeat the refuted answer on $14.7\%$ to $29.4\%$ of transitions and revise to a clue-consistent candidate on $38.9\%$ to $61.7\%$ ($61.7\%$ on Grok-4.3, $53.8\%$ on GPT-5.4 and $52.1\%$ on o3), tracking the lower conversational degradation observed on frontier models.

\Cref{tab:supp_contradiction_corr} (right) further tests \emph{explicit user corrections} on the 150-query five-task slice (and 100-query white-box slice on LLaVA-Video-7B): whenever an intermediate turn commits to an incorrect answer $\hat y$, the user explicitly negates it either before the final clue (\emph{Final correction}: \emph{``Wait, your earlier answer ($\hat y$) is incorrect. Please revise your answer using all clues...''}) or after every wrong intermediate turn (\emph{Online correction}).
Explicit oracle-assisted user negation yields significant recovery on InternVL3.5-8B ($+11.3$\,pp $[6.3, 16.9]$), modest positive point shifts on Qwen3-VL-8B ($+2.7$ to $+3.3$\,pp) and GPT-4.1 ($+2.7$ to $+6.0$\,pp), and almost no effect on LLaVA-Video-7B ($+1.0$\,pp $[0.0, 3.3]$ vs.\ $+22.0$\,pp $[11.5, 32.8]$ from white-box Reply Masking, paired difference $+21.0$\,pp $[10.1, 32.2]$).
Similarly, a prompt-only instruction to \emph{``ignore your earlier answers''} without ground-truth error feedback fails to improve accuracy on 300 paired 7-turn CLEVRER dialogues on Qwen3-VL-8B ($58.0\%\rightarrow 54.7\%$).
Oracle-free reply removal or masking therefore reaches accuracy comparable to oracle-assisted user negation on Qwen3-VL-8B, InternVL3.5-8B and GPT-4.1 (paired $\replystripped-\text{Final corr.}$ within $[-1.3, +2.7]$\,pp, all 95\% CIs spanning zero) and substantially exceeds it on LLaVA-Video-7B, without requiring ground-truth feedback on which intermediate guesses were wrong.

\subsection{Explicit wrong answer vs.\ wrong visual observation vs.\ unsupported rationale}
\label{sec:supp_obs_rationale}
\begin{table*}[!htbp]
\centering\scriptsize\setlength{\tabcolsep}{2.4pt}
\renewcommand{\arraystretch}{1.07}
\caption{\textbf{Decomposing planted assistant history into unsupported rationale, deferred false visual observation and explicit wrong commitment ($150$ balanced five-task queries).} Full $3{,}218$-query natural \sharded and \replystripped baselines appear in \cref{tab:supp_delivery_macro}; contrasts report 4,000-draw dataset-stratified video-clustered 95\% CIs.}
\label{tab:supp_plant_decomp}
\resizebox{\textwidth}{!}{%
\begin{tabular}{@{}lcccccccc@{}}
\toprule
Model & \noted & Unsupp.\ rationale & Deferred wrong obs. & Explicit wrong ans. & Rationale $-$ \noted & Wrong obs.\ $-$ \noted & Wrong ans.\ $-$ \noted & Wrong ans.\ $-$ Wrong obs. \\
\midrule
Qwen3-VL-8B & 44.0 & 41.3 & 50.7 & 19.3 & $-2.7$ [$-7.3$, +1.7] & +6.7 [+0.4, +13.1] & $-24.7$ [$-31.6$, $-17.3$] & $-31.3$ [$-39.1$, $-23.7$] \\
InternVL3.5-8B & 42.0 & 38.7 & 26.7 & 19.3 & $-3.3$ [$-11.7$, +4.6] & $-15.3$ [$-23.8$, $-6.7$] & $-22.7$ [$-30.7$, $-14.2$] & $-7.3$ [$-15.1$, +0.4] \\
GPT-4.1 & 50.0 & 52.7 & 55.3 & 44.0 & +2.7 [$-2.9$, +8.5] & +5.3 [$-0.1$, +10.8] & $-6.0$ [$-12.1$, +0.5] & $-11.3$ [$-18.0$, $-4.8$] \\
\bottomrule
\end{tabular}%
}
\end{table*}
\input{tables/supp_delivery_macro}
To examine how different types of planted intermediate text affect final answers, \cref{tab:supp_plant_decomp} reports descriptive contrasts across four controlled intermediate-reply templates on the 150-query five-task slice under paired final-turn seeds (see \cref{tab:supp_delivery_macro} for full $3{,}218$-query natural \sharded and \replystripped baselines): (i)~short neutral \noted; (ii)~verbose \emph{Unsupported rationale} (plausible analytical prose with neither a visual claim nor a candidate commitment); (iii)~\emph{Deferred false visual observation} (a deterministic false scene description contradicting the target's interactions or active half of the video while explicitly stating \emph{``I will wait before naming an answer''}); and (iv)~\emph{Explicit wrong answer} (\texttt{It's [decoy].}, naming a wrong candidate without visual description; because templates vary length, deferral wording and content jointly, contrasts compare structured reply regimes rather than single-token swaps).
Across all three models, generic unsupported rationale produces no significant harm relative to \noted ($-3.3$ to $+2.7$\,pp, all 95\% CIs spanning zero).
Pairing a false visual observation with explicit answer deferral still hurts InternVL3.5-8B substantially ($-15.3$\,pp $[-23.8, -6.7]$ vs.\ \noted) while explicit deferral slightly lifts Qwen3-VL-8B and GPT-4.1 above bare \noted ($+6.7$ and $+5.3$\,pp); crucially, replacing deferral with an \emph{explicit wrong answer commitment} collapses accuracy across every model relative to both \noted ($-24.7$\,pp $[-31.6, -17.3]$ on Qwen3-VL-8B, $-22.7$\,pp $[-30.7, -14.2]$ on InternVL3.5-8B, $-6.0$\,pp $[-12.1, +0.5]$ on GPT-4.1) and deferred false visual observation ($-31.3$\,pp $[-39.1, -23.7]$, $-7.3$\,pp $[-15.1, +0.4]$ and $-11.3$\,pp $[-18.0, -4.8]$).

\subsection{Full 12-condition delivery study and paired per-model task tables}
\input{tables/supp_delivery_tasks_grid}
The serving-level delivery study covers six models across 12 conditions on all 3,218 diagnostic queries (\cref{tab:supp_delivery_macro}; paired two-column per-model task breakdowns in \cref{tab:delivery_tasks_qwen3-vl-8b,tab:delivery_tasks_internvl3.5-8b,tab:delivery_tasks_internvl3.5-14b,tab:delivery_tasks_llava-video-7b,tab:delivery_tasks_kimi-vl-a3b,tab:delivery_tasks_gpt-4.1}).
To ensure every final-turn history intervention is evaluated under an identical final-turn instruction, both the matched regenerated \sharded baseline (\eg, $44.6\%$ on Qwen3-VL-8B vs.\ $41.0\%$ raw unsuffixed regeneration in \cref{tab:baseline_bridge}) and all post-hoc final-turn conditions in \cref{tab:supp_delivery_macro,tab:supp_ppr,tab:supp_repair_controls} append \emph{``Now give your best answer to my question, using everything I have told you so far.''} to the final user turn (except \emph{Recap}, which uses its own recapitulation suffix in \cref{sec:supp_prompts}, and regenerated multi-turn split conditions).
Regenerated \sharded, \replystripped, acknowledgments (\noted), answer commitments (\texttt{My answer: [extracted answer].}), oracle filtering (replacing clue-violating replies on CLEVRER and Charades), planted wrong/correct answers, single-turn text transcript and \emph{Recap} operate on fixed recorded intermediate histories, whereas two-clues-per-turn, two-turn split ($\lceil m/2\rceil$) and six-turn filler (\texttt{Let me get my thoughts in order first.}) regenerate intermediate turns.
All nonanswers, format echoes and cutoffs remain in the 3,218-query denominator (including Kimi-VL-A3B copying \noted on $2{,}126/3{,}218$ post-hoc acknowledgment edits).

\section{Position-preserving Reply Masking and gap summary}
\label{sec:supp_ppr}

\begin{table*}[!htbp]
\centering\scriptsize\setlength{\tabcolsep}{2.5pt}
\renewcommand{\arraystretch}{1.07}
\caption{\textbf{Applying Reply Masking / \replystripped at every turn throughout the dialogue versus only before the final answer.} Intermediate columns average across all non-final turns $t<m$; paired final-turn contrasts use 4,000 video-clustered bootstrap draws.}
\label{tab:supp_online_mask}
\resizebox{\textwidth}{!}{%
\begin{tabular}{@{}llcccccc@{}}
\toprule
Regime ($n$) & Model & Interm.\ acc.\ (\sharded $\rightarrow$ Online) & Interm.\ viol.\ (\sharded $\rightarrow$ Online) & \sharded final & Final-only mask/\replystripped & All-turn online & Online $-$ \sharded (pp) \\
\midrule
White-box Mask ($100$) & Qwen3-VL-8B & $31.8\% \rightarrow 38.7\%$ & $40.0\% \rightarrow 34.5\%$ & 43.0 & 49.0 & 49.0 & +6.0 [$-3.3$, +14.9] \\
White-box Mask ($100$) & InternVL3.5-8B & $23.7\% \rightarrow 34.5\%$ & $33.2\% \rightarrow 30.3\%$ & 30.0 & 45.0 & 42.0 & +12.0 [+2.7, +21.4] \\
White-box Mask ($100$) & LLaVA-Video-7B & $9.2\% \rightarrow 17.1\%$ & $29.7\% \rightarrow 30.0\%$ & 19.0 & 30.0 & 29.0 & +10.0 [+0.5, +19.7] \\
\midrule
Serving \replystripped ($150$) & Qwen3-VL-8B & $34.9\% \rightarrow 41.1\%$ & $37.7\% \rightarrow 31.2\%$ & 43.3 & 48.7 & 51.3 & +8.0 [+1.5, +14.6] \\
Serving \replystripped ($150$) & InternVL3.5-8B & $28.9\% \rightarrow 36.3\%$ & $26.1\% \rightarrow 30.2\%$ & 28.7 & 38.7 & 40.0 & +11.3 [+3.5, +19.5] \\
Serving \replystripped ($150$) & GPT-4.1 & $35.6\% \rightarrow 42.6\%$ & $27.9\% \rightarrow 26.1\%$ & 46.7 & 51.3 & 54.7 & +8.0 [+0.4, +15.3] \\
\bottomrule
\end{tabular}%
}
\end{table*}

\subsection{Online all-turn Reply Masking versus final-turn masking}
\label{sec:supp_online_mask}
\begin{figure*}[!b]
\centering
\begin{minipage}[t]{0.47\textwidth}
\vspace{0pt}
\centering
\resizebox{\linewidth}{!}{%
\begin{tikzpicture}[
  font=\scriptsize,
  blk/.style={draw=black!55, rounded corners=1.5pt, minimum height=15pt, inner xsep=3pt, align=center},
  drop/.style={draw=okverm!80!black, dashed, fill=okverm!10, rounded corners=1.5pt, minimum height=15pt, inner xsep=3pt, align=center}
]
  \node[blk, fill=plotpurple!14] (sys) at (0,0) {Sys + 24 Frames\\$[0, a_1)$};
  \node[blk, fill=plotblue!14, right=2.5pt of sys] (u1) {User $1$\\$[a_1, b_1)$};
  \node[drop, right=2.5pt of u1] (r1) {Asst $1$\\$[a_2, b_2)$};
  \node[blk, fill=plotblue!14, right=2.5pt of r1] (u2) {User $2$\\$[a_3, b_3)$};
  \node[drop, right=2.5pt of u2] (r2) {Asst $2$\\$[a_4, b_4)$};
  \node[blk, fill=plotteal!16, right=2.5pt of r2] (um) {User $m$+Gen\\$[a_{2m-1}, L)$};

  \draw[-{Stealth[length=4pt]}, thick, okverm!80!black] (r1.south) -- ++(0,-0.36) node[below, font=\tiny, align=center] {Omitted from\\KV \& attn};
  \draw[-{Stealth[length=4pt]}, thick, okverm!80!black] (r2.south) -- ++(0,-0.36) node[below, font=\tiny, align=center] {Omitted from\\KV \& attn};

  \node[blk, fill=plotpurple!14] (ksys) at (0.15,-1.55) {Sys + Frames\\pos $[0, a_1)$};
  \node[blk, fill=plotblue!14, right=3pt of ksys] (ku1) {User $1$\\pos $[a_1, b_1)$};
  \node[blk, fill=plotblue!14, right=3pt of ku1] (ku2) {User $2$\\pos $[a_3, b_3)$};
  \node[blk, fill=plotteal!16, right=3pt of ku2] (kum) {User $m$+Gen\\pos $[a_{2m-1}, L)$};
  \node[blk, fill=okgreen!15, right=3pt of kum] (gen) {Decode\\pos $L, L\!+\!1,\dots$};

  \draw[-{Stealth[length=4pt]}, black!65] (sys.south) -- (ksys.north);
  \draw[-{Stealth[length=4pt]}, black!65] (u1.south) -- (ku1.north);
  \draw[-{Stealth[length=4pt]}, black!65] (u2.south) -- (ku2.north);
  \draw[-{Stealth[length=4pt]}, black!65] (um.south) -- (kum.north);

  \draw[black!70]
    ([yshift=-1.5pt]ksys.south west) -- ++(0,-3.5pt) -- ([yshift=-5pt]ku2.south east) -- ++(0,3.5pt);
  \node[font=\tiny] at ($(ksys.south west)!0.5!(ku2.south east)+(0,-10pt)$) {Prefix KV cache (frames + prior user clues; original RoPE / 3D M-RoPE IDs)};
  \draw[black!70]
    ([yshift=-1.5pt]kum.south west) -- ++(0,-3.5pt) -- ([yshift=-5pt]gen.south east) -- ++(0,3.5pt);
  \node[font=\tiny] at ($(kum.south west)!0.5!(gen.south east)+(0,-10pt)$) {Final-turn prefill \& decoding};
\end{tikzpicture}%
}
\end{minipage}\hfill
\begin{minipage}[t]{0.50\textwidth}
\vspace{0pt}
\fbox{\parbox{\dimexpr\linewidth-2\fboxsep-2\fboxrule\relax}{%
\scriptsize\ttfamily
\textbf{Procedure:} Position-Preserving Reply Masking (White-Box)\\[2pt]
1:\ ids,\ mm,\ spans\ $\leftarrow$\ TemplateEncode(messages,\ frames)\\
2:\ pos\ $\leftarrow$\ GetRoPEIndex(ids,\ mm)\ \textrm{\textit{// [3,1,L] Qwen3-VL; [1,L] 1D}}\\
3:\ drop\ $\leftarrow\ \bigcup_{j \in \text{AsstTurns}}\ \{a_j, \dots, b_j - 1\}$\ \textrm{\textit{// header..turn-end}}\\
4:\ keep\ $\leftarrow\ [i \in \{0,\dots,L-1\} : i \notin \text{drop}]$\\
5:\ SetRotaryTableMinLen($L + \text{max\_tokens}$)\ \textrm{\textit{// valid for pos $\ge|$keep$|$}}\\
6:\ split\ $\leftarrow$\ StartOfFinalUserTurn(keep)\\
7:\ cache\ $\leftarrow$\ Forward(ids[keep[:split]],\ pos[...,\ keep[:split]])\\
8:\ logits,\ cache\ $\leftarrow$\ Forward(ids[keep[split:]],\ pos[...,\ keep[split:]],\ cache)\\
9:\ $p_{\text{next}} \leftarrow \max(\text{pos}[...,\ \text{keep}[-1]]) + 1$\\
10:\ \textbf{for}\ $t = 1 \dots \text{max\_tokens}$:\ $y_t \leftarrow$ Sample(logits)\\
11:\ \ \ logits,\ cache\ $\leftarrow$\ Forward($[y_t]$,\ $p_{\text{next}}$,\ cache);\ $p_{\text{next}}\!\mathrel{+}=1$
}}
\end{minipage}
\caption{\textbf{Schematic and pseudocode for position-preserving Reply Masking.} Full dialogue tokenization assigns every visual and text token its unmasked 1D RoPE (InternVL3.5, LLaVA-Video-7B, Kimi-VL-A3B) or 3D M-RoPE (Qwen3-VL-8B) coordinate before assistant message spans $[a_j, b_j)$ are removed. Surviving tokens pass their sliced original \texttt{position\_ids} into every decoder layer, excluding prior assistant keys and values from the cache while preserving exact relative rotary distances.}
\label{fig:supp_mask_schematic}
\end{figure*}
Main Secs.~5 and 7 evaluate Reply Masking and \replystripped at the final turn on recorded conversations so intermediate user turns share identical histories.
In deployed multi-turn assistants, however, Reply Masking (or serving-level \replystripped) can be applied \emph{online at every turn $t=2\dots m$} so early errors never enter subsequent turns.
\Cref{tab:supp_online_mask} evaluates all-turn online masking across three white-box models ($100$ balanced five-task queries) and all-turn online \replystripped across three serving/API models ($150$ balanced queries).
Across all six model-backend pairs, applying the intervention throughout the dialogue lifts pooled \emph{intermediate-turn accuracy} on the four multi-turn oracle-verified tasks ($+6.2$ to $+10.8$\,pp across all turns $t<m$; \eg, lifting LLaVA-Video-7B intermediate accuracy from $9.2\%$ to $17.1\%$ and InternVL3.5-8B from $23.7\%$ to $34.5\%$) while improving five-task final-turn accuracy by $+6.0$ to $+12.0$\,pp over \sharded (five of six 95\% CIs strictly above zero, and matching final-turn-only Reply Masking within sampling noise because surviving user tokens attend to no prior assistant KV entries in either case).

\subsection{RoPE preservation schematic and KV-cache token footprint}
\label{sec:supp_mask_rope}
\begin{table*}[!htbp]
\centering\scriptsize\setlength{\tabcolsep}{3.4pt}
\renewcommand{\arraystretch}{1.07}
\caption{\textbf{Active KV-cache token reduction under Reply Masking versus fixed-structure semantic content swings ($3{,}218$ diagnostic queries).} Left: exact logged unmasked prompt medians, mean masked assistant tokens and mean percentage of prompt tokens removed by white-box Reply Masking. Right: five-task accuracy swings across serving-level edits that preserve identical turn counts and short assistant footprints (\emph{Noted}, \emph{Answer commitments}, \emph{Wrong plant} and \emph{Correct plant}; paired video-clustered 95\% CIs).}
\label{tab:supp_kv_footprint}
\begin{tabular}{@{}lrrrrccc@{}}
\toprule
Model & Med.\ tok & Masked tok & Share (\%) & Mask $-$ WB base & \emph{Noted} $-$ \sharded & \emph{Ans.\ only} $-$ \sharded & \emph{Correct} $-$ \emph{Wrong} plant \\
\midrule
Qwen3-VL-8B & 4,098 & 50.9 & 1.26\% & +5.1 [3.6, 6.8] & +2.5 [0.9, 4.2] & $-1.4$ [$-2.2$, $-0.6$] & +77.0 [75.4, 78.6] \\
InternVL3.5-8B & 6,685 & 82.8 & 1.22\% & +9.0 [7.1, 10.8] & +7.4 [5.4, 9.3] & +0.9 [$-0.9$, 2.7] & +61.2 [59.3, 63.0] \\
InternVL3.5-14B & 6,666 & 57.8 & 0.86\% & +12.6 [10.7, 14.6] & +8.8 [6.8, 10.8] & $-0.1$ [$-1.8$, 1.7] & +47.5 [45.7, 49.4] \\
LLaVA-Video-7B & 5,494 & 58.0 & 1.05\% & +2.2 [0.5, 3.9] & $-0.7$ [$-2.5$, 1.0] & +0.5 [$-0.7$, 1.7] & +84.7 [83.2, 86.2] \\
Kimi-VL-A3B & 5,686 & 39.4 & 0.71\% & +4.3 [2.5, 6.2] & $-12.8$ [$-14.6$, $-11.1$] & +1.6 [0.1, 3.0] & +68.0 [66.2, 69.8] \\
GPT-4.1 & \pend & \pend & \pend & \pend & +6.9 [5.3, 8.5] & $-0.9$ [$-2.2$, 0.3] & +13.1 [11.2, 15.1] \\
\bottomrule
\end{tabular}
\end{table*}
\Cref{fig:supp_mask_schematic} presents the exact token-span and rotary-index mechanics of position-preserving \emph{Reply Masking}.
Full dialogue tokenization first assigns every visual and text token its unmasked rotary coordinate: 3D multimodal M-RoPE indices $(t,h,w)$ of shape $[3,1,L]$ via \texttt{get\_rope\_index} on Qwen3-VL-8B, and 1D rotary indices $[0,\dots,L-1]$ on InternVL3.5, LLaVA-Video-7B and Kimi-VL-A3B.
Assistant message spans $[a_j,b_j)$ (role header through turn-end marker) are omitted before layer-1 execution so assistant $(K,V)$ states never enter \texttt{DynamicCache}, while rotary lookup tables are sized to $\ge L+\text{max\_tokens}$ so $\cos/\sin$ values at surviving original positions remain exact and decoding begins at $p_{\text{next}}=\max(\text{pos}[\dots,\text{keep}[-1]])+1$.
Across all $3{,}218$ diagnostic queries (\cref{tab:supp_kv_footprint}), the 24 video frames dominate prompt length ($4{,}098\text{--}6{,}685$ median tokens), so Reply Masking removes only $39.4\text{--}82.8$ tokens on average ($0.71\%\text{--}1.26\%$ of the prompt), whereas fixed-structure candidate edits (\emph{Correct} vs.\ \emph{Wrong} plant) swing accuracy by $13.1\text{--}84.7$\,pp at identical turn counts.

\subsection{3,218-query white-box Reply Masking and side-by-side gap decomposition}
\input{tables/supp_ppr}
\begin{table*}[!htbp]
\centering\scriptsize\setlength{\tabcolsep}{4.5pt}
\renewcommand{\arraystretch}{1.07}
\caption{\textbf{Side-by-side summary across the 3,218-query diagnostic subset (five-task mean accuracy, \%).} \full and \concat come from the main benchmark restricted to the 3,218 subset; matched instruction-suffixed \sharded, \replystripped and Recap come from the serving-backend delivery study (\cref{tab:supp_delivery_macro}); white-box Reply Masking (with its matched white-box \sharded baseline in parentheses) comes from the five-model \texttt{delivery\_k1} archive (\cref{tab:supp_ppr}). Because interventions alter overlapping prompt and sequence properties, gains are non-additive.}
\label{tab:supp_gap_decomp}
\begin{tabular}{@{}lcccccc@{}}
\toprule
Model & \full & \concat & \sharded & \replystripped & Reply Mask (WB base) & Recap \\
\midrule
Qwen3-VL-8B & 52.9 & 52.1 & 44.6 & 50.8 (+6.2) & 49.6 (44.5, +5.1) & 46.2 (+1.6) \\
InternVL3.5-8B & 48.1 & 49.4 & 35.1 & 45.1 (+10.0) & 43.6 (34.6, +9.0) & 46.8 (+11.7) \\
InternVL3.5-14B & 57.0 & 55.9 & 40.4 & 52.5 (+12.1) & 52.8 (40.2, +12.6) & 48.9 (+8.5) \\
LLaVA-Video-7B & 31.4 & 29.4 & 22.4 & 24.8 (+2.4) & 24.6 (22.3, +2.2) & 25.8 (+3.4) \\
Kimi-VL-A3B & 31.9 & 24.0 & 20.8 & 27.1 (+6.3) & 25.7 (21.4, +4.3) & 26.9 (+6.2) \\
GPT-4.1 & 56.5 & 57.0 & 48.2 & 53.8 (+5.6) & \pend & 50.4 (+2.2) \\
\bottomrule
\end{tabular}
\end{table*}
\Cref{tab:supp_ppr} reports the five-model 3,218-query white-box Reply Masking ablation archive (\texttt{delivery\_k1}, corresponding to Main Sec.~5.1 and including \emph{First only}, which masks only the first assistant turn and coincides with all-turn masking on the 768 two-clue queries).
Note on archive provenance: \cref{tab:supp_kv_footprint,tab:supp_ppr,tab:supp_gap_decomp} report this five-model \texttt{delivery\_k1} archive (\eg, $49.6\%$, $+5.1$\,pp on Qwen3-VL-8B; $22.3\%$ baseline on LLaVA-Video-7B), whereas Main Table~4 and \cref{tab:supp_ct_transfer,tab:supp_repair_controls,tab:repair_tasks_llava-video-7b,tab:repair_tasks_qwen3-vl-8b,tab:repair_tasks_internvl3.5-14b,tab:repair_tasks_internvl3.5-8b} report the independent four-model white-box repair archive (\texttt{fix}/\texttt{fix\_k1}, giving $49.7\%$, $+5.2$\,pp on Qwen3-VL-8B and $22.4\%$ baseline on LLaVA-Video-7B, matching within $\le 0.1$\,pp across all four shared models under temperature-$1.0$ sampling).
\Cref{tab:supp_gap_decomp} places unsuffixed benchmark \full and \concat alongside matched instruction-suffixed \sharded, \replystripped, white-box Reply Masking and \emph{Recap} on the 3,218-query subset, showing how retained assistant replies and non-reply turn factors jointly contribute to the single-turn gap.

\section{Raw greying accuracy and dose controls}
\label{sec:supp_greying}

\subsection{Robustness across blur, blackout masking, frame removal and $|E|$ strata}
\label{sec:supp_vis_perturb}
\begin{table*}[!t]
\centering\scriptsize\setlength{\tabcolsep}{3.6pt}
\renewcommand{\arraystretch}{1.06}
\caption{\textbf{Visual sensitivity $R$ (pp) across four distinct frame-perturbation operators ($150$ balanced five-task queries; matched budget $k=\min(|E|,24-|E|)$).} Suppression of $R$ in \sharded and recovery under \replystripped hold across neutral greying, Gaussian blur ($r=24$\,px), blackout ($\text{RGB}(0,0,0)$) and complete frame removal ($24-k$ frames retained).}
\label{tab:supp_alt_perturb}
\begin{tabular}{@{}llccccc@{}}
\toprule
Model & Perturbation operator & \full $R$ & \sharded $R$ & \replystripped $R$ & $R_{\full} - R_{\sharded}$ (pp) & $R_{\replystripped} - R_{\sharded}$ (pp) \\
\midrule
Qwen3-VL-8B & Neutral grey $\text{RGB}(128,128,128)$ & 20.0 & 3.3 & 21.3 & +16.7 [+8.4, +25.6] & +18.0 [+9.3, +26.4] \\
Qwen3-VL-8B & Gaussian blur ($r=24$\,px) & 19.3 & 1.3 & 18.0 & +18.0 [+9.7, +26.6] & +16.7 [+8.0, +25.5] \\
Qwen3-VL-8B & Blackout mask $\text{RGB}(0,0,0)$ & 24.7 & 0.0 & 18.7 & +24.7 [+15.9, +34.2] & +18.7 [+10.2, +27.2] \\
Qwen3-VL-8B & Complete frame removal ($24-k$ kept) & 11.3 & 5.3 & 15.3 & +6.0 [$-2.0$, +13.9] & +10.0 [+2.1, +18.0] \\
\midrule
InternVL3.5-8B & Neutral grey $\text{RGB}(128,128,128)$ & 18.0 & 2.7 & 10.7 & +15.3 [+7.8, +22.7] & +8.0 [+1.5, +14.8] \\
InternVL3.5-8B & Gaussian blur ($r=24$\,px) & 12.0 & 6.7 & 10.0 & +5.3 [$-2.9$, +13.3] & +3.3 [$-5.2$, +11.5] \\
InternVL3.5-8B & Blackout mask $\text{RGB}(0,0,0)$ & 11.3 & 6.7 & 9.3 & +4.7 [$-3.5$, +12.7] & +2.7 [$-6.5$, +11.4] \\
InternVL3.5-8B & Complete frame removal ($24-k$ kept) & 12.7 & 0.0 & 16.0 & +12.7 [+3.3, +22.1] & +16.0 [+6.9, +25.3] \\
\midrule
GPT-4.1 & Neutral grey $\text{RGB}(128,128,128)$ & 24.0 & 11.3 & 16.7 & +12.7 [+3.9, +21.5] & +5.3 [$-3.7$, +14.6] \\
GPT-4.1 & Gaussian blur ($r=24$\,px) & 15.3 & 6.7 & 9.3 & +8.7 [+0.4, +16.8] & +2.7 [$-5.0$, +10.4] \\
GPT-4.1 & Blackout mask $\text{RGB}(0,0,0)$ & 17.3 & 8.7 & 14.7 & +8.7 [+0.01, +17.6] & +6.0 [$-3.1$, +14.7] \\
GPT-4.1 & Complete frame removal ($24-k$ kept) & 16.7 & 7.3 & 18.0 & +9.3 [$-0.6$, +19.2] & +10.7 [+1.1, +20.6] \\
\bottomrule
\end{tabular}
\end{table*}
\begin{table}[!htbp]
\centering\scriptsize\setlength{\tabcolsep}{3.5pt}
\renewcommand{\arraystretch}{1.07}
\caption{\textbf{Visual sensitivity $R$ stratified by annotated evidence-frame count $|E|$ across all $3{,}214$ eligible diagnostic queries.} Entries report seven-model mean pooled $R$ (pp) and model counts satisfying $\full>\sharded$ and $\replystripped>\sharded$.}
\label{tab:supp_greying_ebins}
\begin{tabular}{@{}lrrrrc@{}}
\toprule
$|E|$ frames & Eligible $n$ & \full & \sharded & \replystripped & Models ($>\sharded$) \\
\midrule
$1\text{--}4$ & 341 & 12.0 & 3.8 & 11.7 & 7/7, 7/7 \\
$5\text{--}8$ & 1,432 & 16.2 & 5.3 & 13.9 & 7/7, 7/7 \\
$9\text{--}23$ & 1,441 & 16.4 & 6.5 & 13.8 & 7/7, 7/7 \\
\bottomrule
\end{tabular}
\end{table}
To verify that the visual-sensitivity attenuation in \sharded and its recovery under \replystripped are not specific to neutral RGB $(128,128,128)$ greying, \cref{tab:supp_alt_perturb} repeats the matched-budget $k=\min(|E|,24-|E|)$ experiment on the 150-query five-task slice across three alternative interventions: (i)~strong \emph{Gaussian blur} ($r=24$\,px, destroying spatial object detail while preserving color and luminance statistics), (ii)~uniform \emph{Blackout masking} ($\text{RGB}(0,0,0)$), and (iii)~\emph{Complete frame removal} (omitting the $k$ selected frames and their timestamps altogether so only $24-k$ frames are passed, introducing zero synthetic pixels).

\begin{table*}[!b]
\centering\scriptsize\setlength{\tabcolsep}{4.0pt}
\renewcommand{\arraystretch}{1.06}
\caption{\textbf{Five-task visual sensitivity $R$ (pp) and paired video-clustered 95\% bootstrap CIs across all seven models ($3{,}214$ eligible queries, $2{,}938$ videos, 4,000 draws).} Directly verifies the Main Sec.~6.1 confidence-interval summary: all seven $R_{\sharded}-R_{\full}$ intervals lie strictly below zero and all seven $R_{\replystripped}-R_{\sharded}$ intervals lie strictly above zero.}
\label{tab:supp_greying_ci}
\begin{tabular}{@{}lcccccc@{}}
\toprule
Model & $R_{\full}$ & $R_{\concat}$ & $R_{\replystripped}$ & $R_{\sharded}$ & $R_{\sharded} - R_{\full}$ (pp) & $R_{\replystripped} - R_{\sharded}$ (pp) \\
\midrule
Qwen3-VL-8B & 22.3 & 22.3 & 21.9 & 4.0 & $-18.3$ [$-20.3$, $-16.3$] & +17.9 [+15.9, +19.8] \\
InternVL3.5-8B & 13.5 & 16.8 & 12.7 & 4.7 & $-8.8$ [$-10.7$, $-7.0$] & +8.0 [+6.0, +9.9] \\
InternVL3.5-14B & 16.7 & 18.1 & 16.4 & 6.1 & $-10.6$ [$-12.6$, $-8.6$] & +10.3 [+8.2, +12.3] \\
LLaVA-Video-7B & 7.6 & 6.5 & 5.5 & 0.2 & $-7.4$ [$-8.8$, $-6.0$] & +5.3 [+4.0, +6.6] \\
Kimi-VL-A3B & 8.2 & 6.2 & 3.9 & 0.9 & $-7.3$ [$-9.0$, $-5.7$] & +3.0 [+1.5, +4.5] \\
GPT-4.1 & 17.5 & 18.8 & 16.2 & 8.1 & $-9.5$ [$-11.4$, $-7.5$] & +8.1 [+6.2, +10.0] \\
o3 & 19.2 & 19.7 & 19.1 & 15.7 & $-3.5$ [$-5.6$, $-1.4$] & +3.3 [+1.2, +5.4] \\
\midrule
\emph{Seven-model mean} & \emph{15.0} & \emph{15.5} & \emph{13.7} & \emph{5.7} & \emph{$-9.3$} & \emph{+8.0} \\
\bottomrule
\end{tabular}
\end{table*}

In point estimate across all 12/12 model-perturbation cells (\cref{tab:supp_alt_perturb}), sensitivity $R$ is lower in \sharded than in \full and rebounds under \replystripped; notably, under complete frame removal (zero synthetic pixel replacement), $\replystripped-\sharded$ recovers $R$ with 95\% CIs strictly above zero on all three models ($+10.0$\,pp $[+2.1,+18.0]$ on Qwen3-VL-8B, $+16.0$\,pp $[+6.9,+25.3]$ on InternVL3.5-8B and $+10.7$\,pp $[+1.1,+20.6]$ on GPT-4.1; smaller $n=150$ blur and blackout recovery CIs are wider on InternVL3.5-8B and GPT-4.1).
\Cref{tab:supp_greying_ebins} further confirms that across all $3{,}214$ eligible diagnostic queries, partitioning by annotated evidence-set size $|E|\in[1,4]$, $[5,8]$ and $[9,23]$ preserves seven-model mean pooled ordering $\full > \replystripped > \sharded$, with $7/7$ models satisfying both $\full>\sharded$ and $\replystripped>\sharded$ in every stratum.

\subsection{Paired 7-model confidence intervals, dose controls and raw greying tables}
\input{tables/supp_dose}
\input{tables/supp_greying}
Matched frame greying on the full diagnostic slice evaluates seven models across $3{,}214$ eligible queries ($858 / 1{,}144 / 396 / 519 / 297$ across $2{,}938$ unique videos, excluding the four $|E|=24, k=0$ queries), with a $\pm 0.3$\,s temporal buffer around annotated clue-supporting events.
\Cref{tab:supp_greying_ci} provides the exact five-task visual sensitivities $R$ and dataset-stratified video-clustered 95\% bootstrap confidence intervals (4,000 draws) reported in Main Sec.~6.1: every model has $R_{\sharded}-R_{\full}<0$ with its 95\% CI strictly below zero ($-3.5$ $[-5.6,-1.4]$ on o3 to $-18.3$ $[-20.3,-16.3]$ on Qwen3-VL-8B) and $R_{\replystripped}-R_{\sharded}>0$ with its 95\% CI strictly above zero ($+3.0$ $[+1.5,+4.5]$ on Kimi-VL-A3B and $+3.3$ $[+1.2,+5.4]$ on o3 up to $+17.9$ $[+15.9,+19.8]$ on Qwen3-VL-8B).
\Cref{tab:supp_dose} reports graded dose curves ($0\%$ to $100\%$ evidence share at fixed budget $k$) plus \emph{Ev.\ only}, \emph{Non-ev.\ only} and \emph{All grey} controls, and \cref{tab:greying_full,tab:greying_concat_clues,tab:greying_sharded,tab:greying_strip} report complete per-task raw accuracies.

\section{\cleanturn implementation, settings and coverage}
\label{sec:supp_repair_settings}

\subsection{Practical latency, compute and KV-cache memory overhead}
\label{sec:supp_overhead}
\begin{table*}[!htbp]
\centering\scriptsize\setlength{\tabcolsep}{2.4pt}
\renewcommand{\arraystretch}{1.07}
\caption{\textbf{Measured KV-cache memory (MB) and synchronized final-turn wall-clock latency (ms) on an NVIDIA A100-80GB GPU ($50$ balanced five-task queries per model).} Columns 6--9 report median / mean ms; Column 10 reports cold vs.\ visual-prefix-cached \full median ms; Column 11 reports fixed-$\alpha$ vs.\ auto-$\alpha$ \cleanturn median ms.}
\label{tab:supp_overhead}
\resizebox{\textwidth}{!}{%
\begin{tabular}{@{}lrrrrcccccc@{}}
\toprule
& \multicolumn{4}{c}{Mean KV-cache footprint (MB)} & \multicolumn{4}{c}{Session latency: median / mean (ms)} & \multicolumn{2}{c}{Clean-pass median (ms)} \\
\cmidrule(lr){2-5}\cmidrule(lr){6-9}\cmidrule(l){10-11}
Model & 24-frame vis. & \sharded & Mask & Parallel tail & \sharded cached & Mask (cached pre.) & Mask (post-hoc) & Recap cached & \full (cold / vis-c.) & \cleanturn (fix / auto) \\
\midrule
LLaVA-Video-7B & 280.3 & 297.6 & 294.6 & +14.4 (4.8\%) & 243 / 267 & 229 / 340 & 286 / 393 & 214 / 245 & 751 / 200 & 809 / 1,834 \\
Qwen3-VL-8B & 569.6 & 594.3 & 581.7 & +12.1 (2.0\%) & 188 / 594 & 197 / 254 & 244 / 299 & 194 / 672 & 866 / 208 & 881 / 1,906 \\
InternVL3.5-8B & 910.7 & 932.9 & 922.9 & +12.2 (1.3\%) & 174 / 395 & 212 / 346 & 262 / 395 & 232 / 642 & 1,423 / 508 & 1,962 / 2,726 \\
InternVL3.5-14B & 1,011.9 & 1,035.0 & 1,025.4 & +13.6 (1.3\%) & 354 / 547 & 376 / 480 & 439 / 544 & 324 / 635 & 1,871 / 478 & 1,999 / 3,456 \\
\bottomrule
\end{tabular}%
}
\end{table*}
\Cref{tab:supp_overhead} quantifies synchronized wall-clock latency and exact GPU KV-cache footprints across all four open-weight repair models on an idle NVIDIA A100-80GB GPU ($50$ balanced five-task queries per model).
Three practical deployment findings emerge:
\begin{enumerate}[leftmargin=1.4em,itemsep=1pt,topsep=2pt]
  \item \textbf{Zero-overhead cached-prefix Reply Masking:} When the masked user-only prefix is maintained across turns (via turn-end KV rollback), final-turn decoding requires \textbf{zero extra memory} (final-turn KV cache is $3.0\text{--}12.6$\,MB smaller than unmasked \sharded) and \textbf{$197\text{--}376$\,ms median final-turn latency} (measured from the cached masked prefix), matching cached unmasked \sharded ($174\text{--}354$\,ms) with zero auxiliary passes.
  \item \textbf{Negligible cost of a parallel user-only KV tail:} If an application must simultaneously retain an unmasked KV cache alongside a reply-masked stream, both streams share the read-only 24-frame visual prefix ($280.3\text{--}1{,}011.9$\,MB, or $94\text{--}98\%$ of total KV memory); maintaining a parallel user-only text KV tail adds only \textbf{$12.1\text{--}14.4$\,MB} ($1.3\%\text{--}4.8\%$ of KV memory), or \textbf{$47\text{--}63$\,ms} if recomputing the short user-clue tail post hoc atop the cached visual prefix.
  \item \textbf{Clean-restart vs.\ in-place \cleanturn trade-off:} Cold \full restart re-encodes all 24 frames ($751\text{--}1{,}871$\,ms median; $200\text{--}508$\,ms if reusing a cached visual prefix), whereas \cleanturn trades one clean $S$ prefill ($809\text{--}1{,}999$\,ms median under fixed $\alpha$) plus 24-token reference KL scoring under \texttt{auto}-$\alpha$ ($1{,}834\text{--}3{,}456$\,ms median) for in-place decoder steering when user-turn fragmentation leaves Reply Masking below \full.
\end{enumerate}

\subsection{\cleanturn construction and frozen model configurations}
For each query, the clean reference prompt $S$ concatenates the video, intent and all revealed clues into one user turn, while the base prompt $B$ contains only the video and intent.
Residual-stream differences are extracted at the final prompt token, and Clean-State Steering adds $\alpha\delta_\ell$ across an inclusive decoder layer band $\Lambda$ to all final-user-turn tokens and subsequently generated tokens.
When combined with Reply Masking in \cleanturn, both the current-context vector ($\delta^{\text{delta}}_\ell = h_\ell(S)-h_\ell(H_{\text{mask}})$) and per-query KL calibration operate on the reply-masked context rather than reusing unmasked activations.
Under \texttt{auto} calibration, a 24-token greedy reference continuation (terminated with the model's turn-end token) is generated from $S$, and $\alpha\in\{0,0.1,0.15,0.25,0.35,0.5,0.75,1\}$ is chosen per query to minimize next-token KL divergence to the clean $S$ distribution along that reference without target labels.
All four repair models have complete 3,218-query evaluations across all seven unmasked conditions and all three reply-masked conditions, with model-level settings (\cref{tab:supp_settings}) frozen on the 600-case tuning split (150 cases from each of the four larger tasks; 400 scored cases on InternVL3.5-8B, where nonzero candidate selection chose 45.8\% vs.\ 47.2\% for zero steering).

\begin{table}[!htbp]
  \caption{\textbf{Frozen \cleanturn and Steering-only settings.} Inclusive implementation layer indices; strength is fixed or selected per case by auto. Steering-only and \cleanturn configurations are selected separately before evaluation.}
  \label{tab:supp_settings}
  \centering\scriptsize\setlength{\tabcolsep}{2.4pt}
  \resizebox{\linewidth}{!}{%
  \begin{tabular}{@{}lcccccc@{}}
    \toprule
    & \multicolumn{3}{c}{Steering only} & \multicolumn{3}{c}{\cleanturn (Reply Mask + Steering)} \\
    \cmidrule(lr){2-4}\cmidrule(l){5-7}
    Model & Difference & Layers & Strength & Difference & Layers & Strength \\
    \midrule
    LLaVA-Video-7B & delta & 18--24 & 0.25 & delta & 18--24 & 0.25 \\
    Qwen3-VL-8B & delta & 21--29 & auto & delta & 23--31 & auto \\
    InternVL3.5-14B & shadow & 26--35 & 0.15 & delta & 18--27 & auto \\
    InternVL3.5-8B & shadow & 13--21 & 0.25 & shadow & 20--28 & auto \\
    \bottomrule
  \end{tabular}%
  }
\end{table}

\section{\cleanturn details, harms and sensitivity analyses}
\label{sec:supp_repair_results}

\subsection{Zero-retuning transfer across lengths, clue orders and unseen tasks}
\label{sec:supp_ct_transfer}
\begin{table*}[!htbp]
\centering\scriptsize\setlength{\tabcolsep}{3.1pt}
\renewcommand{\arraystretch}{1.07}
\caption{\textbf{Transfer of frozen \cleanturn configurations without retuning across unseen Perception Test ($n=298$), conversation lengths ($3{,}215$ queries across $2$, $3$ and $7$ turns), and clue permutations.} Right block reports full $3{,}218$-query \texttt{Order 0} alongside two unseen shuffled orders (\texttt{Order 1} and \texttt{Order 2}, $n=100$ five-task queries each).}
\label{tab:supp_ct_transfer}
\begin{tabular}{@{}lcccc@{\hspace{8pt}}ccc@{\hspace{8pt}}ccc@{}}
\toprule
& \multicolumn{4}{c}{Zero-shot unseen Perception Test ($n=298$)} & \multicolumn{3}{c}{$\cleanturn - \text{Mask}$ by length (pp)} & \multicolumn{3}{c}{Clue orders: Mask / \cleanturn (\%)} \\
\cmidrule(r){2-5}\cmidrule(lr){6-8}\cmidrule(l){9-11}
Model & \sharded & Mask & \cleanturn & $\cleanturn - \text{Mask}$ (pp) & 2 turns & 3 turns & 7 turns & Order 0 ($3{,}218$) & Order 1 ($100$) & Order 2 ($100$) \\
\midrule
LLaVA-Video-7B & 37.6 & 45.6 & 57.0 & +11.4 [+6.0, +17.1] & +9.8 & +4.1 & +9.7 & 24.6 / 32.7 & 30.0 / 31.0 & 31.0 / 34.0 \\
Qwen3-VL-8B & 53.0 & 59.1 & 60.7 & +1.7 [$-1.7$, +5.4] & +0.0 & +4.1 & +3.8 & 49.7 / 51.6 & 56.0 / 59.0 & 50.0 / 54.0 \\
InternVL3.5-8B & 34.9 & 48.0 & 49.0 & +1.0 [$-0.3$, +2.3] & +0.1 & +0.0 & +0.6 & 43.6 / 44.2 & 41.0 / 48.0 & 36.0 / 43.0 \\
InternVL3.5-14B & 35.2 & 57.4 & 56.0 & $-1.3$ [$-4.4$, +1.7] & $-0.1$ & $-6.1$ & +0.4 & 52.8 / 52.6 & \pend & \pend \\
\bottomrule
\end{tabular}
\end{table*}
A key question is whether frozen \cleanturn configurations transfer across conversation lengths, shuffled clue schedules and unseen tasks without per-setting retuning (\cref{tab:supp_ct_transfer}):
\begin{itemize}[leftmargin=1.4em,itemsep=1pt,topsep=2pt]
  \item \textbf{Zero-shot unseen-task transfer (\emph{Perception Test}, $298$ queries):} Zero Perception Test queries were used during hyperparameter tuning (\cref{sec:supp_datasets}). On this unseen real-world dataset in the frozen repair archive, \cleanturn improves LLaVA-Video-7B by $+11.4$\,pp beyond Reply Masking alone ($45.6\%\rightarrow 57.0\%$, 95\% CI $[+6.0, +17.1]$, matching \full $56.7\%$) and preserves Reply Masking gains on Qwen3-VL-8B ($59.1\%\rightarrow 60.7\%$), InternVL3.5-8B ($48.0\%\rightarrow 49.0\%$) and InternVL3.5-14B ($57.4\%\rightarrow 56.0\%$).
  \item \textbf{Conversation-length transfer ($2$, $3$ and $7$ turns across $3{,}215$ queries; excluding $3$ four-turn Perception Test queries):} On LLaVA-Video-7B, frozen steering adds $+9.8$\,pp over Reply Masking on short 2-turn dialogues ($n=768$) and $+9.7$\,pp on long 7-turn dialogues ($n=2{,}398$); on Qwen3-VL-8B, where Reply Masking already closes the 2-turn gap ($44.9\%$), steering adds $+4.1$\,pp on 3-turn ($n=49$) and $+3.8$\,pp on 7-turn dialogues where multi-turn user fragmentation is largest.
  \item \textbf{Shuffled clue-order transfer (\texttt{Order 1} and \texttt{Order 2}, $100$ balanced five-task queries per order, zero retuning):} Applying the exact \texttt{Order 0} frozen settings to two unseen clue permutations yields positive $\cleanturn-\sharded$ gains in all $6/6$ held-out model-order cells ($+5.0$ to $+18.0$\,pp; $5/6$ held-out CIs excluding zero) while matching or exceeding Reply Masking on both held-out orders across all three tested models ($+1.0$ to $+7.0$\,pp; paired $100$-query \texttt{Order 0} scores appear alongside VCD and RVD in \cref{tab:supp_pub_mitigations}).
\end{itemize}

\subsection{Regime check with published decoding mitigations and recent LVLM work}
\label{sec:supp_pub_mitig}
\begin{table*}[!htbp]
\centering\scriptsize\setlength{\tabcolsep}{2.8pt}
\renewcommand{\arraystretch}{1.07}
\caption{\textbf{Diagnostic regime check of unmodified single-turn decoding formulations (VCD~\cite{leng2024vcd} and RVD~\cite{mmhalsnowball2024}) on a common 100-query ($20$/task) multi-clue video slice.} Evaluated without claiming general superiority outside progressive multi-clue disclosure (Main Sec.~2); full $3{,}218$-query repair comparisons appear in \cref{tab:supp_repair_controls}. Intervals use 4,000 video-clustered bootstrap draws.}
\label{tab:supp_pub_mitigations}
\resizebox{\textwidth}{!}{%
\begin{tabular}{@{}lccccccccc@{}}
\toprule
Model & WB \sharded & VCD~\cite{leng2024vcd} & RVD~\cite{mmhalsnowball2024} & Reply Mask & \cleanturn & VCD $-$ \sharded & RVD $-$ \sharded & Mask $-$ RVD (pp) & \cleanturn $-$ RVD (pp) \\
\midrule
LLaVA-Video-7B & 15.0 & 13.0 & 19.0 & 37.0 & 35.0 & $-2.0$ [$-6.0$, +1.5] & +4.0 [$-2.8$, +11.1] & +18.0 [+9.3, +27.1] & +16.0 [+6.7, +25.4] \\
Qwen3-VL-8B & 47.0 & 42.0 & 38.0 & 52.0 & 53.0 & $-5.0$ [$-9.2$, $-1.0$] & $-9.0$ [$-16.5$, $-1.7$] & +14.0 [+6.1, +22.0] & +15.0 [+6.1, +24.3] \\
InternVL3.5-8B & 34.0 & 38.0 & 31.0 & 43.0 & 47.0 & +4.0 [$-2.7$, +10.9] & $-3.0$ [$-10.9$, +6.0] & +12.0 [+2.8, +21.7] & +16.0 [+6.4, +25.7] \\
\bottomrule
\end{tabular}%
}
\end{table*}

\begin{table*}[!htbp]
\centering\scriptsize\setlength{\tabcolsep}{3.8pt}
\renewcommand{\arraystretch}{1.07}
\caption{\textbf{Positioning relative to recent LVLM evidence-reliance diagnostics and training-free white-box interventions.}}
\label{tab:supp_related_positioning}
\begin{tabular}{@{}llp{0.23\textwidth}p{0.18\textwidth}p{0.26\textwidth}@{}}
\toprule
Work & Regime & Target phenomenon or mechanism & Bank / extra passes & Relation to our study \\
\midrule
WhatIfVis~\cite{li2026whatifvis} & Static image, 1-turn & Vision vs.\ language-prior utilization & Probe / SFT or learned vector & Establishes post-perceptual utilization bottleneck on images \\
CircuitProbe~\cite{zhang2025circuitprobe} & Video QA, 1-turn & Mid-to-late temporal head tracing & Head localization; 0 extra pass & Explains why upper-layer bands carry consolidated video state \\
REVEAL~\cite{sethuraman2026reveal} & Video QA, 1-turn & Temporal shortcuts \& user sycophancy & Diagnostic stress benchmark & Tests adversarial user assertions; we test truthful multi-turn clues \\
CARVE~\cite{alhamidi2026carve} & Video agents, 1-turn & SHAM vs.\ DESTROY replay sensitivity & $2k$ black-box agent replays & Counterfactual evidence probe for routing; we match $|E|$ vs.\ $\bar E$ across turns \\
\midrule
PTI~\cite{zhang2026pti} & Static image, 1-turn & Prefill KV steering against hallucination & Contrastive bank; 0 extra pass & Modality-decoupled static direction vs.\ zero-bank Reply Masking \\
DMAS~\cite{yin2026dmas} & Static image, 1-turn & Clustered truthfulness + noisy-image heads & Labeled cluster DB + noisy pass & Cluster-retrieved hallucination vector vs.\ label-free per-query $S$ state \\
IKOD~\cite{yang2025ikod} & Static image, 1-turn & KV text merging against attention decay & 0 bank; auxiliary KV branch & Compresses low-attn single-turn text; Reply Mask omits prior assistant spans \\
\midrule
\textbf{Reply Masking (ours)} & \textbf{Video QA, multi-turn} & \textbf{Self-generated assistant anchoring} & \textbf{0 bank; 0 extra passes} & \textbf{Omits assistant KV spans (rollback or parallel tail) at exact 1D/3D RoPE IDs} \\
\textbf{\cleanturn (ours)} & \textbf{Video QA, multi-turn} & \textbf{Residual user-turn fragmentation} & \textbf{Tuning band; $S$ prefill (+ KL)} & \textbf{Model-dependent per-query clean-state residual atop Reply Masking} \\
\bottomrule
\end{tabular}
\end{table*}
\input{tables/supp_repair_controls}
\input{tables/supp_repair_tasks_grid}
Consistent with Main Sec.~2, we do not claim superiority over training or decoding mitigations developed for visual hallucination or self-contained queries.
To clarify how those published formulations behave when applied without redesign to dispersed multi-clue video dialogues, \cref{tab:supp_pub_mitigations} reports a diagnostic regime check on a common 100-query ($20$/task) white-box slice alongside Reply Masking and \cleanturn: \textbf{Residual Visual Decoding (RVD)}~\cite{mmhalsnowball2024} (weighting conversation logits against a residual video-plus-final-query stream via Jensen--Shannon divergence, $\beta=2.0$) and \textbf{Visual Contrastive Decoding (VCD)}~\cite{leng2024vcd} (contrasting intact-video against diffusion-noised video logits under adaptive plausibility constraints, $\alpha=1.0,\beta=0.1$).
In this multi-clue disclosure regime, unmodified RVD and VCD remain near or below \sharded ($13.0\text{--}42.0\%$, vs.\ $37.0\text{--}53.0\%$ for Reply Masking and \cleanturn; paired differences over unmodified RVD are $+12.0$ to $+18.0$\,pp across all three models).
Mechanistically, VCD contrasts visual noise while leaving erroneous prior assistant replies intact in both branches, whereas unmodified RVD's residual branch conditions only on the video and final user turn, omitting clues $1\dots m-1$ from earlier user turns (unlike single-turn \emph{MMHalSnowball} queries where the final question is self-contained; we did not evaluate a modified JS-divergence-weighted RVD variant with consolidated earlier clues, and fixed $0.5$-weight \emph{Clean-context mix} in Main Table~4 serves only as a separate clue-consolidated logit-mixing control rather than an algorithmic equivalent of RVD).
Training-time policy distillation approaches such as MAIGO~\cite{maigo2026} instead address multi-turn drift via trajectory supervision and weight updates, making zero-training Reply Masking complementary at inference time.

\Cref{tab:supp_related_positioning} further situates our diagnostics and mitigations relative to concurrent LVLM studies: \textbf{WhatIfVis}~\cite{li2026whatifvis} (showing via a trained reconstruction probe on static images that coarse visual attributes remain encoded in final-layer image tokens even when predictions follow language priors), \textbf{CircuitProbe}~\cite{zhang2025circuitprobe} (tracing single-turn video temporal consolidation to mid-to-late decoder layers), \textbf{REVEAL}~\cite{sethuraman2026reveal} (stress-testing single-turn VidLMs under user sycophancy and temporal shortcuts), \textbf{CARVE}~\cite{alhamidi2026carve} (contrasting \textsc{sham} vs.\ \textsc{destroy} agent replays as a routing signal), \textbf{PTI}~\cite{zhang2026pti} (single-turn prefill KV steering), \textbf{DMAS}~\cite{yin2026dmas} (clustered truthfulness and noisy-image head steering) and \textbf{IKOD}~\cite{yang2025ikod} (merging low-image-attention generated text KV entries during single-turn captioning).
Whereas PTI and DMAS estimate static or cluster-level hallucination directions on single-turn images (within our clean-prompt setup on 300 7-turn CLEVRER dialogues on Qwen3-VL-8B, a fold-averaged CAA direction \texttt{avg\_vector} trails instance-specific steering by $8.3$\,pp $[4.0, 12.7]$), and IKOD merges single-turn generated tokens via two-branch decoding, multi-turn chat explicitly demarcates prior assistant turns by role headers, allowing \emph{Reply Masking} to omit them at zero extra passes while \cleanturn addresses residual user-turn fragmentation.

\subsection{3,218-query paired contrasts, harms, multiplicity and per-task tables}
Across all 3,218 evaluation queries in the four-model repair archive (\texttt{fix}/\texttt{fix\_k1}, matching Main Table~4; see \cref{sec:supp_ppr} for provenance relative to the five-model \texttt{delivery\_k1} archive), \cref{tab:supp_repair_controls} reports component contrasts and negative controls, and \cref{tab:repair_tasks_llava-video-7b,tab:repair_tasks_qwen3-vl-8b,tab:repair_tasks_internvl3.5-14b,tab:repair_tasks_internvl3.5-8b} report complete paired two-column per-task repair accuracies.
Unmasked \emph{Steering only} flips $467/196$ (wrong-to-right / right-to-wrong) cases on LLaVA-Video-7B, $398/175$ on Qwen3-VL-8B, $390/97$ on InternVL3.5-14B and $205/196$ on InternVL3.5-8B; full \cleanturn on Qwen3-VL-8B flips $473/163$ ($+310$ net cases, $+9.63$\,pp pooled, $+7.08$\,pp five-task macro), while transferring the conversation-derived vector onto clean \full degrades Qwen3-VL-8B ($-2.0$\,pp) and InternVL3.5-8B ($-10.3$\,pp, $460/3{,}218$ cutoffs).
Familywise 95\% Bonferroni bootstrap intervals for Steering-over-Reply-Mask across the four models are $+8.18$ $[5.97, 10.55]$ (LLaVA-Video-7B), $+1.88$ $[0.25, 3.53]$ (Qwen3-VL-8B, 20,000 draws), $-0.21$ $[-1.38, 0.99]$ (InternVL3.5-14B) and $+0.57$ $[-0.63, 1.78]$ (InternVL3.5-8B); for $\cleanturn-\full$, $+1.40$ $[-0.27, 3.03]$, $-0.56$ $[-1.83, 0.80]$, $-4.37$ $[-6.16, -2.62]$ and $-6.04$ $[-8.12, -4.14]$.
Excluding the 126 diagnostic queries sharing a video with another task's tuning split ($3{,}092$ strictly video-disjoint queries) leaves Steering-over-Reply-Mask unchanged: $+8.35$ $[6.62, 10.05]$, $+1.84$ $[0.56, 3.16]$, $-0.21$ $[-1.14, 0.68]$ and $+0.49$ $[-0.46, 1.44]$.

\section{Limitations, result freeze and reproducibility}
\label{sec:supp_audit}

\paragraph{Study limitations.}
Our main 18-model benchmark isolates progressive disclosure of a fixed multi-clue target under uniform 24-frame sampling, complemented by targeted 100-to-250-query interactive studies on response-dependent adaptive clarification, explicit user corrections, progressive frame streaming and $12/24/48$-frame density plus exploratory VidOR and HD-EPIC domain checks (\cref{sec:supp_datasets,sec:supp_scale,sec:supp_delivery}); open-ended multi-topic chat, user-initiated goal changes and continuous multi-minute video streams remain important future extensions, and 8B open-weight models remain near floor on fine-grained fisheye egocentric clips without higher-resolution perception.
Matched-budget visual perturbations (neutral greying, Gaussian blur, blackout and frame removal; \cref{sec:supp_greying}) measure behavioural sensitivity $R$ of final answers to annotated supporting event frames conditional on the dialogue context rather than isolating an internal visual-grounding circuit.
Finally, while position-preserving \emph{Reply Masking} adds zero forward passes via KV rollback (or $12\text{--}14$\,MB for a parallel user-only tail; \cref{sec:supp_overhead}), omitting assistant replies removes prior assistant text if a later user turn explicitly refers back to an earlier explanation, and \cleanturn requires white-box residual hooks, tuning-split band selection and extra clean-prompt passes (\cref{sec:supp_repair_settings,sec:supp_repair_results}).

%% file: fig/supp_qual_walkthrough.tex
\begin{figure*}[!t]
\centering\scriptsize
\setlength{\fboxsep}{3.5pt}
\renewcommand{\arraystretch}{1.04}

\fbox{\begin{minipage}{0.978\textwidth}
\textbf{(a) Real-world object identification} (\emph{Perception Test}, \texttt{video\_11123-1}, $34.9$\,s; LLaVA-Video-7B; Ground truth: \textbf{bottle}; Single-turn \full: \emph{``The yellow bottle''} \textcolor{okgreen!85!black}{\textbf{[Correct]}})\par\vspace{2.5pt}
\begin{minipage}[t]{0.241\linewidth}\centering
  \includegraphics[width=\linewidth]{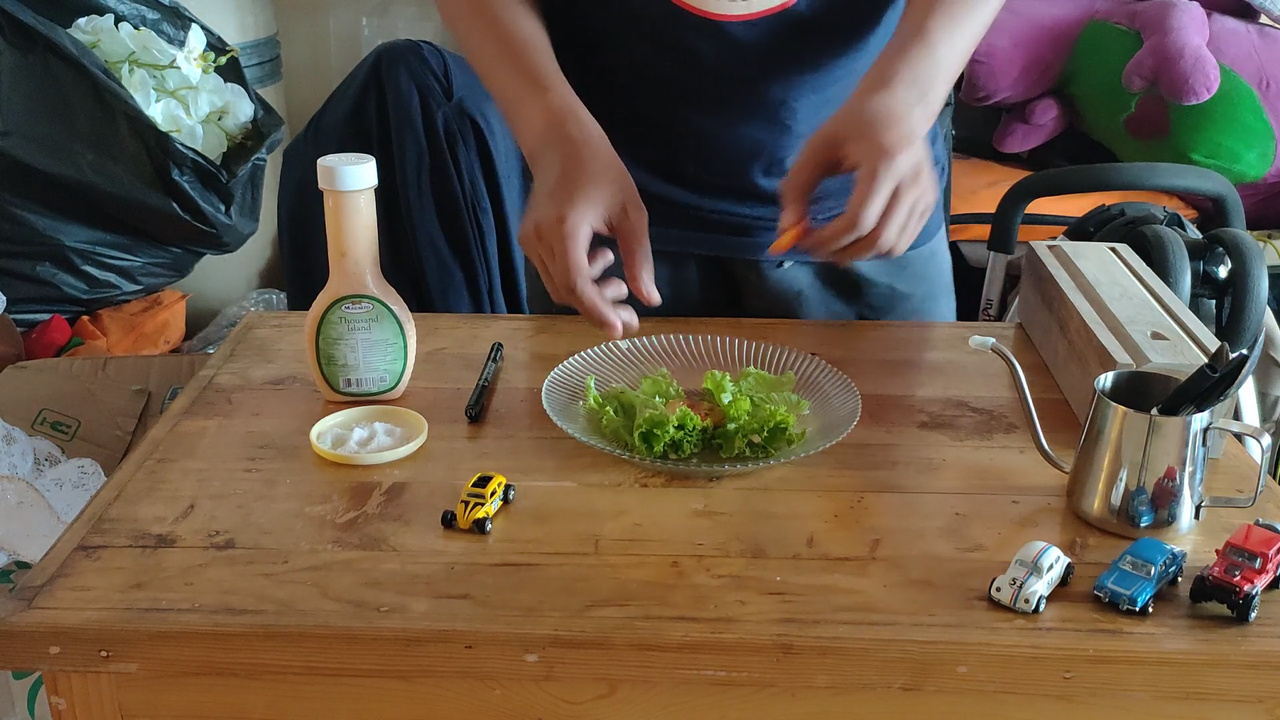}\\[-1pt]
  {\tiny \texttt{12.36s}: 1st half (bottle untouched)}
\end{minipage}\hfill
\begin{minipage}[t]{0.241\linewidth}\centering
  \includegraphics[width=\linewidth]{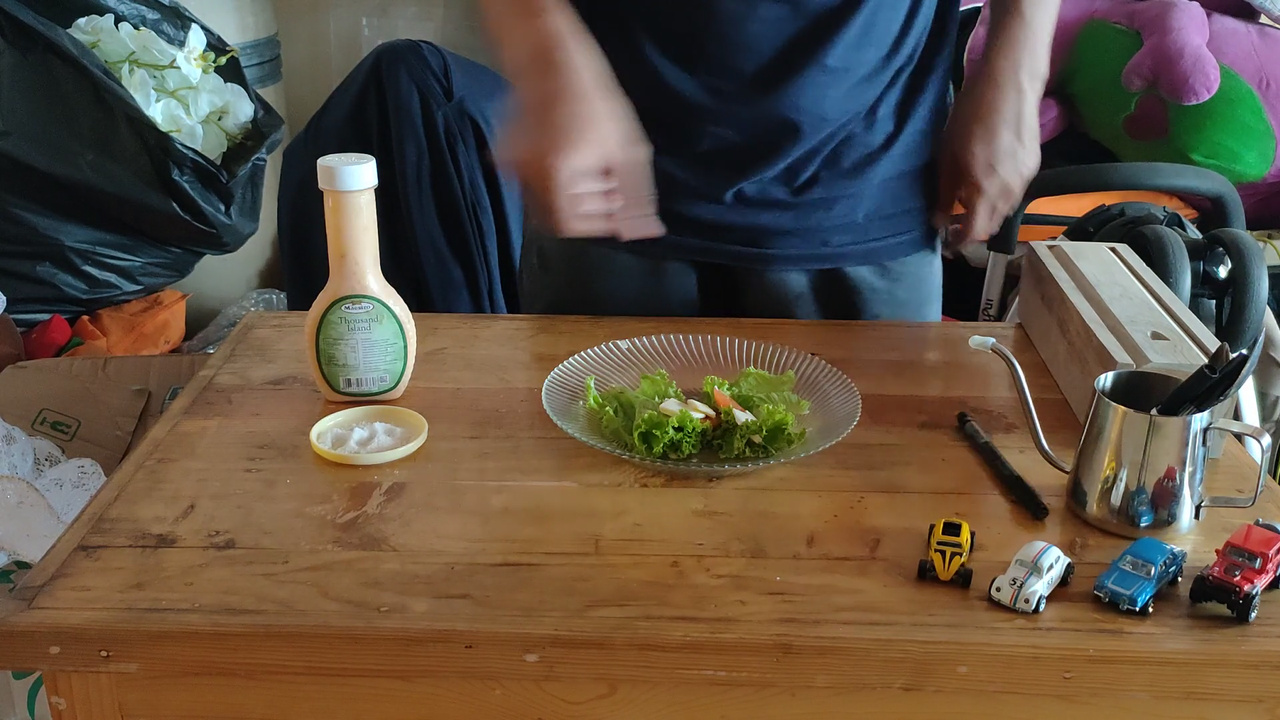}\\[-1pt]
  {\tiny \texttt{21.09s}: 2nd half (reaches for bottle)}
\end{minipage}\hfill
\begin{minipage}[t]{0.241\linewidth}\centering
  \includegraphics[width=\linewidth]{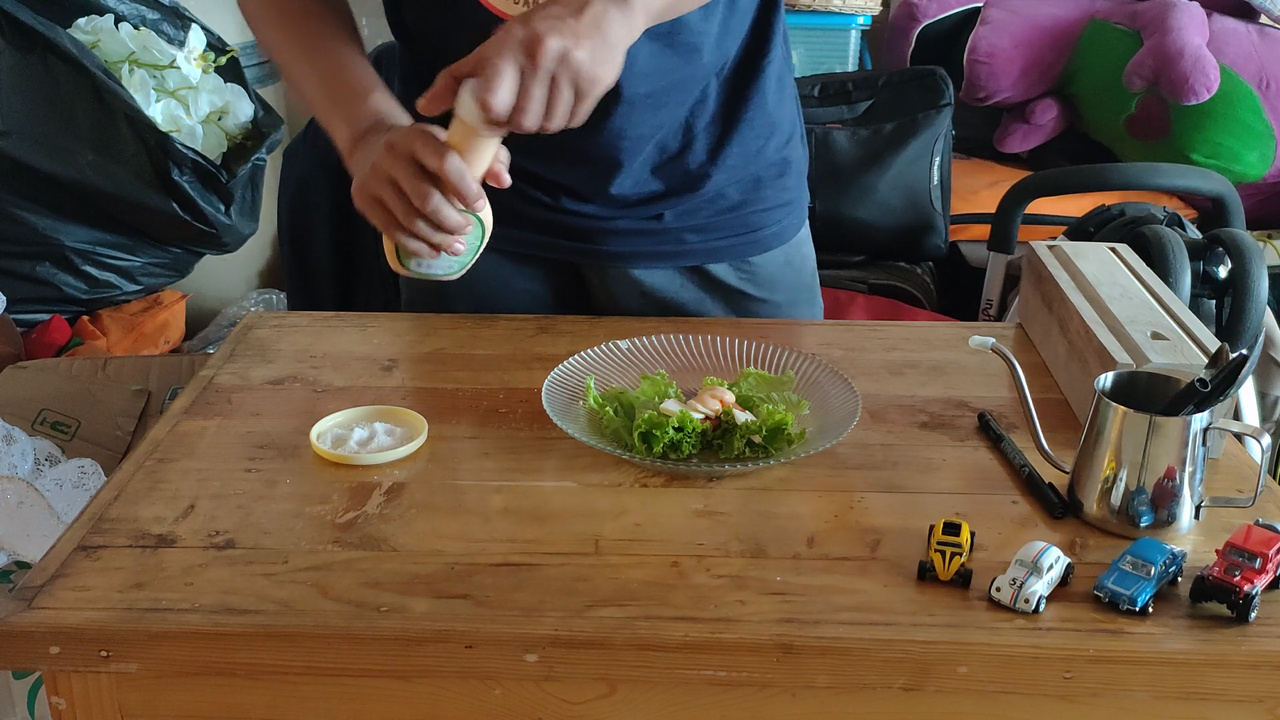}\\[-1pt]
  {\tiny \texttt{26.93s}: caps yellow bottle \textbf{[Clue 1]}}
\end{minipage}\hfill
\begin{minipage}[t]{0.241\linewidth}\centering
  \includegraphics[width=\linewidth]{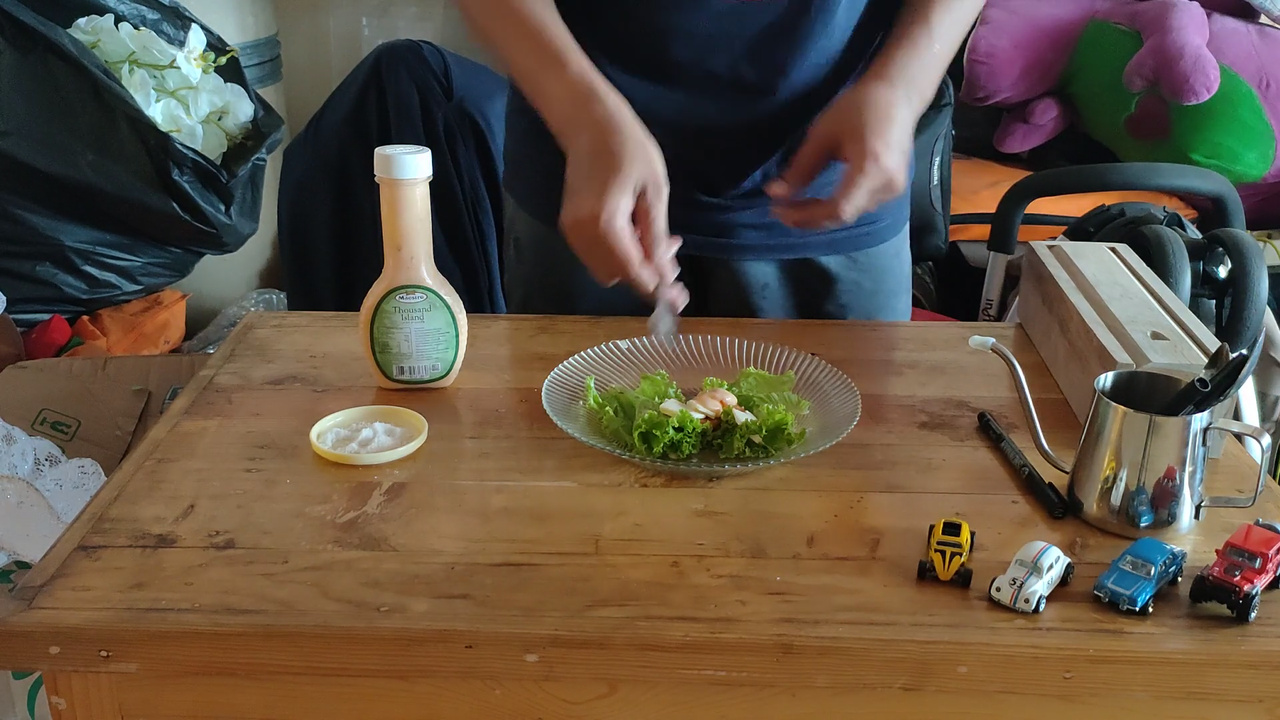}\\[-1pt]
  {\tiny \texttt{31.26s}: mixes salad bowl \textbf{[Clue 3]}}
\end{minipage}\par\vspace{2.5pt}
\textbf{Turn 1 User:} \emph{Which object does the person handle in this video? They cover it with something.} $\rightarrow$ \textbf{\sharded Asst 1:} \texttt{``The person handles the bowl in this video, covering it with something.''} \textcolor{okverm!90!black}{\textbf{[Early wrong guess]}}\\
\textbf{Turn 2 User:} \emph{They only handle it during the second half of the video.} $\rightarrow$ \textbf{\sharded Asst 2:} \texttt{``Yes, the person handles the bowl only during the second half of the video.''} \textcolor{okverm!90!black}{\textbf{[Rationalizes prior guess]}}\\
\textbf{Turn 3 User:} \emph{They are already done handling it before they start mixing something with something.} $\rightarrow$ \textbf{\sharded Final:} \texttt{``No. They are not done handling it before they start mixing something...''} \textcolor{okverm!90!black}{\textbf{[Disputes clue; Wrong]}}\\
\textbf{Repaired Final:} \textbf{Reply Masking:} \texttt{``The bottle of salad dressing.''} \textcolor{okgreen!85!black}{\textbf{[Correct]}} \quad|\quad \textbf{\cleanturn:} \texttt{``bottle''} \textcolor{okgreen!85!black}{\textbf{[Correct]}}
\end{minipage}}\par\vspace{3pt}

\fbox{\begin{minipage}{0.978\textwidth}
\textbf{(b) Real-world temporal action localization} (\emph{Charades grounding}, \texttt{5EEXF}, $33.5$\,s; Qwen3-VL-8B; Ground truth: \textbf{16.9--23.4\,s} [equiv.\ span \textbf{18.1--23.6\,s}, \emph{putting a blanket}]; \full: \texttt{``20.27 to 24.43 seconds''} \textcolor{okgreen!85!black}{\textbf{[max IoU $=0.53$, Correct]}})\par\vspace{2.5pt}
\begin{minipage}[t]{0.241\linewidth}\centering
  \includegraphics[width=\linewidth]{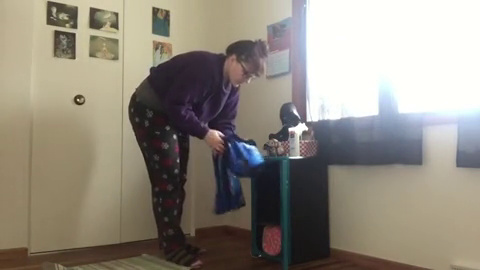}\\[-1pt]
  {\tiny \texttt{6.30s}: cabinet (locked wrong span)}
\end{minipage}\hfill
\begin{minipage}[t]{0.241\linewidth}\centering
  \includegraphics[width=\linewidth]{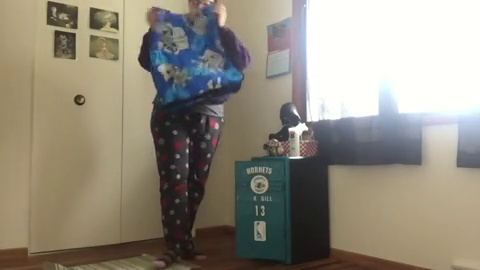}\\[-1pt]
  {\tiny \texttt{13.27s}: cabinet closed \textbf{[Clues 5--7]}}
\end{minipage}\hfill
\begin{minipage}[t]{0.241\linewidth}\centering
  \includegraphics[width=\linewidth]{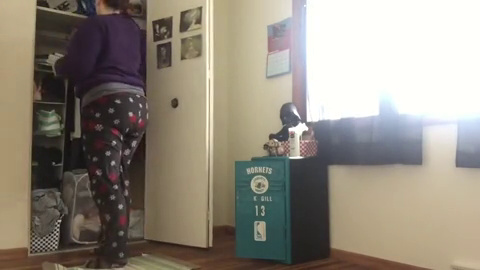}\\[-1pt]
  {\tiny \texttt{20.27s}: putting blanket \textbf{[Target]}}
\end{minipage}\hfill
\begin{minipage}[t]{0.241\linewidth}\centering
  \includegraphics[width=\linewidth]{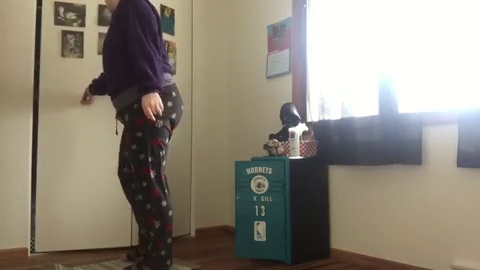}\\[-1pt]
  {\tiny \texttt{25.83s}: seated on bed \textbf{[Clue 2]}}
\end{minipage}\par\vspace{2.5pt}
\textbf{Turn 1 User:} \emph{When in this video does the person do something? Give the start and end time in seconds. The person is putting something.} $\rightarrow$ \textbf{\sharded Asst 1:} \texttt{``from 1.4 to 10.0 seconds''} \textcolor{okverm!90!black}{\textbf{[Premature wrong span; violates T1]}}\\
\textbf{Turns 2--6 User:} \emph{(T2) Ends before standing-to-sitting. (T3) Happens while throwing a blanket. (T4) Involves a blanket. (T5) Starts after taking a blanket. (T6) Starts after closing a cabinet.} $\rightarrow$ \textbf{\sharded Asst 2--6:} \texttt{``from 1.4 to 10.0 seconds''} repeated verbatim on every turn \textcolor{okverm!90!black}{\textbf{[Violates all revealed clues]}}\\
\textbf{Turn 7 User:} \emph{It starts after the person is done opening a cabinet.} $\rightarrow$ \textbf{\sharded Final:} \texttt{``from 1.4 to 10.0 seconds''} \textcolor{okverm!90!black}{\textbf{[IoU $=0.00$, Wrong]}}\\
\textbf{Repaired Final:} \textbf{Reply Masking:} \texttt{``from 16.07 to 23.03 seconds''} \textcolor{okgreen!85!black}{\textbf{[IoU $=0.84$, Correct]}} \quad|\quad \textbf{\cleanturn:} \texttt{``18.87 to 23.03 seconds''} \textcolor{okgreen!85!black}{\textbf{[IoU $=0.76$, Correct]}}
\end{minipage}}\par\vspace{3pt}

\fbox{\begin{minipage}{0.978\textwidth}
\textbf{(c) Multi-event collision reasoning} (\emph{CLEVRER object}, \texttt{14641}, $5.0$\,s; Qwen3-VL-8B; Ground truth: \textbf{purple metal cylinder}; \full: \texttt{``the purple metal cylinder''} \textcolor{okgreen!85!black}{\textbf{[Correct]}})\par\vspace{2.5pt}
\begin{minipage}[t]{0.241\linewidth}\centering
  \includegraphics[width=\linewidth]{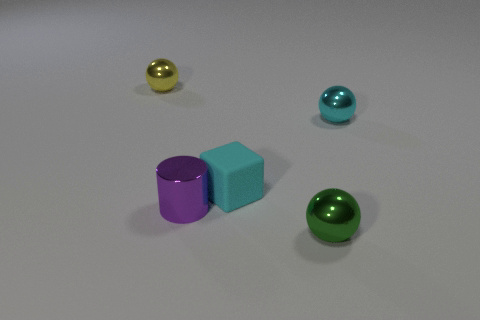}\\[-1pt]
  {\tiny \texttt{1.40s}: purple cyl.\ $\leftrightarrow$ cyan cube \textbf{[T1,7]}}
\end{minipage}\hfill
\begin{minipage}[t]{0.241\linewidth}\centering
  \includegraphics[width=\linewidth]{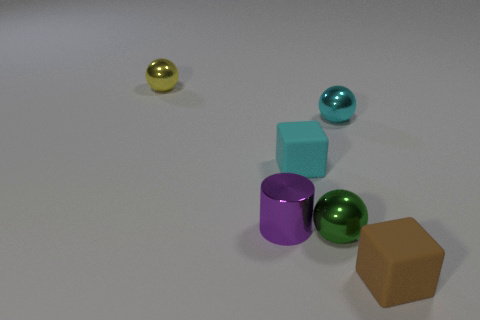}\\[-1pt]
  {\tiny \texttt{2.48s}: purple cyl.\ $\leftrightarrow$ green sph.\ \textbf{[T2,3,6]}}
\end{minipage}\hfill
\begin{minipage}[t]{0.241\linewidth}\centering
  \includegraphics[width=\linewidth]{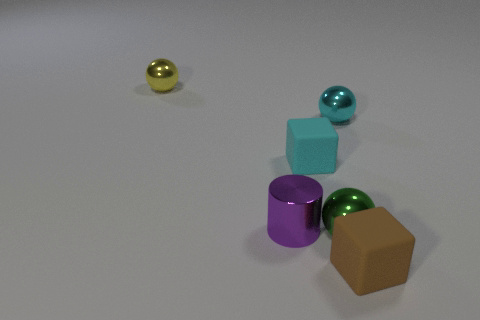}\\[-1pt]
  {\tiny \texttt{2.68s}: green $\leftrightarrow$ brown collision \textbf{[T5]}}
\end{minipage}\hfill
\begin{minipage}[t]{0.241\linewidth}\centering
  \includegraphics[width=\linewidth]{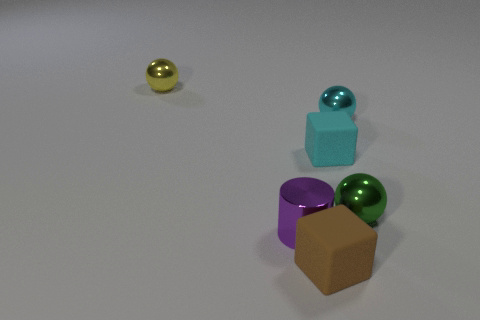}\\[-1pt]
  {\tiny \texttt{3.12s}: purple cyl.\ $\leftrightarrow$ brown cube \textbf{[T4]}}
\end{minipage}\par\vspace{2.5pt}
\textbf{Turn 1 User:} \emph{Which object in this video am I thinking of? It collides with something made of rubber.} $\rightarrow$ \textbf{\sharded Asst 1:} \texttt{``the cyan cube''} \textcolor{okverm!90!black}{\textbf{[Violates T1]}}\\
\textbf{Turns 2--6 User:} \emph{(T2) Collides with a sphere. (T3) Collides with metal. (T4) Collides with a cube. (T5) Enters before green--brown collision. (T6) Collides with green.} $\rightarrow$ \textbf{\sharded Asst 2--6:} \texttt{``the cyan cube''} repeated verbatim across all five turns \textcolor{okverm!90!black}{\textbf{[Violates revealed collision clues]}}\\
\textbf{Turn 7 User:} \emph{It collides with something cyan.} $\rightarrow$ \textbf{\sharded Final:} \texttt{``the cyan cube''} \textcolor{okverm!90!black}{\textbf{[Violates T1--T3, T6--T7; Wrong]}}\\
\textbf{Repaired Final:} \textbf{Reply Masking:} \texttt{``The purple metal cylinder.''} \textcolor{okgreen!85!black}{\textbf{[Correct]}} \quad|\quad \textbf{\cleanturn:} \texttt{``the purple metal cylinder''} \textcolor{okgreen!85!black}{\textbf{[Correct]}}
\end{minipage}}

\caption{\textbf{Illustrative conversation walkthroughs from frozen benchmark and repair logs} (selected from held-out diagnostic queries where single-turn \full and both repairs succeed while \sharded exhibits multi-turn lock-in). Each panel shows four exact sampled video frames at full width alongside the verbatim trajectory. Although all 24 timestamped frames stay visible at every turn, an early wrong guess in \sharded persists despite refuting clues (disputing Clue~3 in \textbf{a}, repeating a refuted early timestamp span across all seven turns in \textbf{b}, and repeating a candidate refuted by five collision clues across seven turns in \textbf{c}), whereas \emph{Reply Masking} and \cleanturn recover the target.}
\label{fig:supp_qual_walkthrough}
\end{figure*}

%% file: tables/supp_delivery_macro.tex
\begin{table*}[!htbp]
\caption{\textbf{All delivery variants.} Equal-task mean accuracy (\%), with nonanswers and failures retained. Qwen/InternVL/Kimi use vLLM; LLaVA uses white-box execution; GPT-4.1 uses API serving. These are separate from the white-box Reply Masking controls below.}
\label{tab:supp_delivery_macro}
\centering\scriptsize\setlength{\tabcolsep}{3pt}
\renewcommand{\arraystretch}{1.12}
\begin{tabular}{@{}lcccccc@{}}
\toprule
Condition & Qwen3-VL-8B & InternVL3.5-8B & InternVL3.5-14B & LLaVA-Video-7B & Kimi-VL-A3B & GPT-4.1 \\
\midrule
\sharded & 44.6 & 35.1 & 40.4 & 22.4 & 20.8 & 48.2 \\
\replystripped & 50.8 & 45.1 & 52.5 & 24.8 & 27.1 & 53.8 \\
Acknowledgments & 47.1 & 42.5 & 49.2 & 21.7 & 7.9 & 55.1 \\
Answer commitments & 43.2 & 35.9 & 40.3 & 22.8 & 22.3 & 47.3 \\
Oracle filter & 49.6 & 39.2 & 43.2 & 25.6 & 21.6 & 50.7 \\
Wrong answer planted & 15.7 & 17.6 & 23.2 & 5.9 & 8.9 & 46.0 \\
Correct answer planted & 92.7 & 78.8 & 70.7 & 90.5 & 76.9 & 59.1 \\
Text transcript & 43.3 & 34.4 & 47.1 & 21.8 & 24.1 & 48.2 \\
Recap & 46.2 & 46.8 & 48.9 & 25.8 & 26.9 & 50.4 \\
Two clues per turn & 46.4 & 37.0 & 47.9 & 27.4 & 25.4 & 49.1 \\
Two turns & 47.2 & 40.3 & 46.6 & 24.8 & 20.0 & 47.1 \\
Filler then full request & 53.1 & 48.5 & 54.2 & 29.8 & 26.6 & 56.0 \\
\bottomrule
\end{tabular}
\end{table*}

%% file: tables/supp_delivery_tasks_grid.tex
\begin{table*}[!htbp]
\centering
\begin{minipage}[t]{0.488\textwidth}
\vspace{0pt}
\caption{\textbf{Qwen3-VL-8B: delivery accuracy (\%).} Each task uses its full frozen evaluation slice.}
\label{tab:delivery_tasks_qwen3-vl-8b}
\centering\scriptsize\setlength{\tabcolsep}{2.8pt}
\renewcommand{\arraystretch}{1.12}
\resizebox{\linewidth}{!}{\begin{tabular}{@{}lcccccc@{}}
\toprule
Condition & C. obj. & C. time & Ch. ground & Ch. obj. & P. obj. & Mean \\
\midrule
\sharded & 58.4 & 54.5 & 22.2 & 35.6 & 52.3 & 44.6 \\
\replystripped & 69.8 & 65.1 & 20.5 & 37.4 & 61.4 & 50.8 \\
Acknowledgments & 66.8 & 55.3 & 21.0 & 36.4 & 56.0 & 47.1 \\
Answer commitments & 57.8 & 50.7 & 21.2 & 34.7 & 51.7 & 43.2 \\
Oracle filter & 69.2 & 62.8 & 26.3 & 37.2 & 52.3 & 49.6 \\
Wrong answer planted & 4.0 & 12.1 & 13.4 & 14.8 & 34.2 & 15.7 \\
Correct answer planted & 99.7 & 93.1 & 98.0 & 83.5 & 89.3 & 92.7 \\
Text transcript & 58.5 & 55.4 & 21.7 & 33.7 & 47.3 & 43.3 \\
Recap & 57.2 & 59.4 & 23.2 & 35.6 & 55.7 & 46.2 \\
Two clues per turn & 64.2 & 58.0 & 17.2 & 35.4 & 57.0 & 46.4 \\
Two turns & 69.3 & 68.3 & 17.7 & 33.1 & 47.3 & 47.2 \\
Filler then full request & 76.8 & 75.9 & 19.7 & 33.5 & 59.4 & 53.1 \\
\bottomrule
\end{tabular}}
\end{minipage}\hfill
\begin{minipage}[t]{0.488\textwidth}
\vspace{0pt}
\caption{\textbf{InternVL3.5-8B: delivery accuracy (\%).} Each task uses its full frozen evaluation slice.}
\label{tab:delivery_tasks_internvl3.5-8b}
\centering\scriptsize\setlength{\tabcolsep}{2.8pt}
\renewcommand{\arraystretch}{1.12}
\resizebox{\linewidth}{!}{\begin{tabular}{@{}lcccccc@{}}
\toprule
Condition & C. obj. & C. time & Ch. ground & Ch. obj. & P. obj. & Mean \\
\midrule
\sharded & 64.5 & 40.3 & 10.6 & 23.8 & 36.2 & 35.1 \\
\replystripped & 78.1 & 63.5 & 15.9 & 23.6 & 44.3 & 45.1 \\
Acknowledgments & 75.3 & 54.3 & 12.9 & 28.4 & 41.6 & 42.5 \\
Answer commitments & 65.7 & 42.6 & 12.4 & 25.1 & 33.9 & 35.9 \\
Oracle filter & 72.0 & 49.0 & 12.9 & 25.7 & 36.6 & 39.2 \\
Wrong answer planted & 10.5 & 6.6 & 10.9 & 21.1 & 38.9 & 17.6 \\
Correct answer planted & 98.6 & 97.8 & 93.7 & 47.7 & 56.0 & 78.8 \\
Text transcript & 61.5 & 46.9 & 9.6 & 21.1 & 32.9 & 34.4 \\
Recap & 78.2 & 64.9 & 13.1 & 28.5 & 49.0 & 46.8 \\
Two clues per turn & 59.7 & 41.7 & 10.9 & 23.2 & 49.7 & 37.0 \\
Two turns & 74.0 & 67.7 & 9.3 & 19.7 & 30.5 & 40.3 \\
Filler then full request & 81.1 & 67.9 & 14.4 & 27.2 & 52.0 & 48.5 \\
\bottomrule
\end{tabular}}
\end{minipage}
\end{table*}
\begin{table*}[!htbp]
\centering
\begin{minipage}[t]{0.488\textwidth}
\vspace{0pt}
\caption{\textbf{InternVL3.5-14B: delivery accuracy (\%).} Each task uses its full frozen evaluation slice.}
\label{tab:delivery_tasks_internvl3.5-14b}
\centering\scriptsize\setlength{\tabcolsep}{2.8pt}
\renewcommand{\arraystretch}{1.12}
\resizebox{\linewidth}{!}{\begin{tabular}{@{}lcccccc@{}}
\toprule
Condition & C. obj. & C. time & Ch. ground & Ch. obj. & P. obj. & Mean \\
\midrule
\sharded & 63.5 & 62.9 & 12.6 & 25.7 & 37.2 & 40.4 \\
\replystripped & 80.0 & 79.5 & 13.1 & 31.0 & 59.1 & 52.5 \\
Acknowledgments & 72.8 & 73.3 & 14.4 & 28.0 & 57.7 & 49.2 \\
Answer commitments & 67.0 & 62.4 & 14.4 & 26.6 & 31.2 & 40.3 \\
Oracle filter & 70.0 & 70.9 & 12.6 & 25.3 & 37.2 & 43.2 \\
Wrong answer planted & 17.4 & 23.6 & 12.1 & 25.1 & 37.9 & 23.2 \\
Correct answer planted & 93.5 & 96.1 & 89.6 & 38.3 & 36.2 & 70.7 \\
Text transcript & 70.3 & 77.7 & 16.4 & 26.6 & 44.3 & 47.1 \\
Recap & 71.9 & 84.4 & 13.9 & 25.9 & 48.7 & 48.9 \\
Two clues per turn & 66.6 & 68.8 & 16.2 & 29.9 & 58.1 & 47.9 \\
Two turns & 72.6 & 80.8 & 14.9 & 27.0 & 37.6 & 46.6 \\
Filler then full request & 85.1 & 88.5 & 17.4 & 23.2 & 56.7 & 54.2 \\
\bottomrule
\end{tabular}}
\end{minipage}\hfill
\begin{minipage}[t]{0.488\textwidth}
\vspace{0pt}
\caption{\textbf{LLaVA-Video-7B: delivery accuracy (\%).} Each task uses its full frozen evaluation slice.}
\label{tab:delivery_tasks_llava-video-7b}
\centering\scriptsize\setlength{\tabcolsep}{2.8pt}
\renewcommand{\arraystretch}{1.12}
\resizebox{\linewidth}{!}{\begin{tabular}{@{}lcccccc@{}}
\toprule
Condition & C. obj. & C. time & Ch. ground & Ch. obj. & P. obj. & Mean \\
\midrule
\sharded & 33.8 & 17.0 & 0.8 & 22.4 & 37.9 & 22.4 \\
\replystripped & 36.9 & 15.3 & 1.8 & 24.7 & 45.3 & 24.8 \\
Acknowledgments & 33.1 & 13.0 & 2.5 & 19.7 & 39.9 & 21.7 \\
Answer commitments & 33.3 & 19.1 & 3.0 & 22.2 & 36.6 & 22.8 \\
Oracle filter & 44.6 & 20.2 & 0.8 & 24.5 & 37.9 & 25.6 \\
Wrong answer planted & 0.5 & 3.0 & 10.4 & 4.2 & 11.4 & 5.9 \\
Correct answer planted & 99.8 & 100.0 & 98.7 & 72.0 & 82.2 & 90.5 \\
Text transcript & 34.0 & 16.2 & 0.5 & 23.8 & 34.6 & 21.8 \\
Recap & 38.9 & 19.3 & 0.5 & 27.4 & 43.0 & 25.8 \\
Two clues per turn & 38.0 & 17.6 & 1.5 & 29.7 & 50.0 & 27.4 \\
Two turns & 41.1 & 22.6 & 1.0 & 24.5 & 34.9 & 24.8 \\
Filler then full request & 43.2 & 24.3 & 1.8 & 24.3 & 55.4 & 29.8 \\
\bottomrule
\end{tabular}}
\end{minipage}
\end{table*}
\begin{table*}[!htbp]
\centering
\begin{minipage}[t]{0.488\textwidth}
\vspace{0pt}
\caption{\textbf{Kimi-VL-A3B: delivery accuracy (\%).} Each task uses its full frozen evaluation slice.}
\label{tab:delivery_tasks_kimi-vl-a3b}
\centering\scriptsize\setlength{\tabcolsep}{2.8pt}
\renewcommand{\arraystretch}{1.12}
\resizebox{\linewidth}{!}{\begin{tabular}{@{}lcccccc@{}}
\toprule
Condition & C. obj. & C. time & Ch. ground & Ch. obj. & P. obj. & Mean \\
\midrule
\sharded & 25.3 & 22.3 & 2.3 & 21.1 & 32.9 & 20.8 \\
\replystripped & 27.4 & 35.8 & 7.8 & 20.1 & 44.3 & 27.1 \\
Acknowledgments & 12.4 & 4.0 & 2.0 & 8.2 & 13.1 & 7.9 \\
Answer commitments & 27.5 & 25.6 & 4.8 & 20.1 & 33.6 & 22.3 \\
Oracle filter & 29.0 & 23.0 & 2.5 & 20.7 & 32.6 & 21.6 \\
Wrong answer planted & 6.1 & 4.7 & 10.6 & 5.4 & 17.8 & 8.9 \\
Correct answer planted & 89.6 & 93.6 & 96.0 & 52.5 & 52.7 & 76.9 \\
Text transcript & 29.7 & 27.1 & 6.1 & 21.3 & 36.2 & 24.1 \\
Recap & 31.4 & 35.8 & 6.8 & 22.2 & 38.6 & 26.9 \\
Two clues per turn & 23.3 & 27.7 & 4.8 & 23.0 & 48.3 & 25.4 \\
Two turns & 30.0 & 37.9 & 4.5 & 8.6 & 19.1 & 20.0 \\
Filler then full request & 42.0 & 31.4 & 1.8 & 14.6 & 43.3 & 26.6 \\
\bottomrule
\end{tabular}}
\end{minipage}\hfill
\begin{minipage}[t]{0.488\textwidth}
\vspace{0pt}
\caption{\textbf{GPT-4.1: delivery accuracy (\%).} Each task uses its full frozen evaluation slice.}
\label{tab:delivery_tasks_gpt-4.1}
\centering\scriptsize\setlength{\tabcolsep}{2.8pt}
\renewcommand{\arraystretch}{1.12}
\resizebox{\linewidth}{!}{\begin{tabular}{@{}lcccccc@{}}
\toprule
Condition & C. obj. & C. time & Ch. ground & Ch. obj. & P. obj. & Mean \\
\midrule
\sharded & 60.4 & 75.5 & 11.6 & 36.4 & 57.0 & 48.2 \\
\replystripped & 64.5 & 77.5 & 19.7 & 41.4 & 65.8 & 53.8 \\
Acknowledgments & 65.4 & 81.4 & 22.0 & 41.6 & 65.1 & 55.1 \\
Answer commitments & 55.5 & 75.3 & 14.1 & 37.7 & 53.7 & 47.3 \\
Oracle filter & 62.5 & 80.6 & 16.2 & 37.9 & 56.4 & 50.7 \\
Wrong answer planted & 38.6 & 73.3 & 21.2 & 38.3 & 58.7 & 46.0 \\
Correct answer planted & 73.2 & 85.4 & 31.3 & 46.0 & 59.7 & 59.1 \\
Text transcript & 59.0 & 74.4 & 13.9 & 37.5 & 56.0 & 48.2 \\
Recap & 58.5 & 75.3 & 11.6 & 42.1 & 64.4 & 50.4 \\
Two clues per turn & 52.8 & 72.2 & 13.1 & 39.8 & 67.4 & 49.1 \\
Two turns & 50.6 & 75.7 & 14.9 & 38.3 & 56.0 & 47.1 \\
Filler then full request & 69.9 & 81.6 & 19.7 & 40.6 & 68.1 & 56.0 \\
\bottomrule
\end{tabular}}
\end{minipage}
\end{table*}

%% file: tables/supp_ppr.tex
\begin{table*}[!htbp]
\caption{\textbf{Position-preserving Reply Masking.} Accuracy and paired gains against the same archive's white-box controls. Intervals are pointwise video-bootstrap 95\% CIs, not an equivalence test.}
\label{tab:supp_ppr}
\centering\scriptsize\setlength{\tabcolsep}{3pt}
\renewcommand{\arraystretch}{1.12}
\begin{tabular}{@{}lcccccc@{}}
\toprule
Model & \sharded & \replystripped & Reply Mask & First only & Mask $-$ \sharded & Mask $-$ \replystripped \\
\midrule
Qwen3-VL-8B & 44.5 & 50.6 & 49.6 & 45.9 & +5.1 [3.6, 6.8] & $-1.0$ [$-1.8$, $-0.2$] \\
InternVL3.5-8B & 34.6 & 45.5 & 43.6 & 37.7 & +9.0 [7.1, 10.8] & $-1.9$ [$-3.0$, $-0.7$] \\
InternVL3.5-14B & 40.2 & 52.8 & 52.8 & 47.4 & +12.6 [10.7, 14.6] & +0.0 [$-1.0$, 1.1] \\
LLaVA-Video-7B & 22.3 & 24.8 & 24.6 & 23.6 & +2.2 [0.5, 3.9] & $-0.2$ [$-1.1$, 0.6] \\
Kimi-VL-A3B & 21.4 & 25.3 & 25.7 & 20.5 & +4.3 [2.5, 6.2] & +0.4 [$-0.6$, 1.4] \\
\bottomrule
\end{tabular}
\end{table*}

%% file: tables/supp_dose.tex
\begin{table*}[!htbp]
\caption{\textbf{Dose and retained-frame controls.} Five-task mean accuracy (\%), the same 3,214 eligible cases. Dose changes the evidence share of a fixed greyed-frame count. Evidence/non-evidence only retains $k$ selected frames; all-grey retains image slots, timestamps and any original replies.}
\label{tab:supp_dose}
\centering\scriptsize\setlength{\tabcolsep}{2.5pt}
\renewcommand{\arraystretch}{1.12}
\begin{tabular}{@{}lccccccccc@{}}
\toprule
Model & Context & 0 & 25 & 50 & 75 & 100 & Ev. only & Non-ev. only & All grey \\
\midrule
Qwen3-VL-8B & \full & 56.6 & 51.7 & 45.7 & 40.3 & 34.4 & 72.6 & 29.0 & 5.6 \\
Qwen3-VL-8B & \sharded & 41.9 & 41.1 & 39.9 & 38.5 & 37.3 & 43.5 & 36.1 & 37.8 \\
InternVL3.5-8B & \full & 48.9 & 46.2 & 43.5 & 40.2 & 34.6 & 58.1 & 28.6 & 10.4 \\
InternVL3.5-8B & \sharded & 29.8 & 29.5 & 28.5 & 27.3 & 25.4 & 33.8 & 23.9 & 23.7 \\
InternVL3.5-14B & \full & 56.6 & 54.1 & 49.2 & 46.0 & 39.4 & 67.2 & 32.0 & 13.7 \\
InternVL3.5-14B & \sharded & 38.9 & 37.7 & 36.8 & 35.4 & 33.1 & 40.9 & 29.9 & 29.8 \\
LLaVA-Video-7B & \full & 31.1 & 29.9 & 28.5 & 27.4 & 23.4 & 30.2 & 20.8 & 10.0 \\
LLaVA-Video-7B & \sharded & 18.1 & 17.9 & 18.1 & 18.1 & 17.6 & 17.5 & 17.0 & 16.0 \\
Kimi-VL-A3B & \full & 32.7 & 30.5 & 29.0 & 27.3 & 23.8 & 36.3 & 21.0 & 10.7 \\
Kimi-VL-A3B & \sharded & 14.4 & 13.7 & 13.7 & 13.8 & 13.1 & 14.9 & 13.1 & 12.0 \\
\bottomrule
\end{tabular}
\end{table*}

%% file: tables/supp_greying.tex
\begin{table*}[!htbp]
\caption{\textbf{\full: final-answer accuracy under greying.} Cells are intact / evidence grey / non-evidence grey (\%). Eligible n: 858 / 1,144 / 396 / 519 / 297. \sharded retains original intact-video replies; \replystripped removes them.}
\label{tab:greying_full}
\centering\scriptsize\setlength{\tabcolsep}{3pt}
\renewcommand{\arraystretch}{1.12}
\begin{tabular}{@{}lccccc@{}}
\toprule
Model & CLEVRER obj. & CLEVRER time & Charades ground & Charades obj. & Perception \\
\midrule
Qwen3-VL-8B & 73.4 / 58.2 / 76.9 & 70.9 / 43.6 / 73.6 & 21.2 / 1.5 / 32.8 & 37.4 / 21.8 / 38.7 & 59.6 / 47.8 / 62.3 \\
InternVL3.5-8B & 84.7 / 74.4 / 83.7 & 73.2 / 44.8 / 72.4 & 11.9 / 2.5 / 13.9 & 23.1 / 16.0 / 25.4 & 48.5 / 37.0 / 46.8 \\
InternVL3.5-14B & 85.8 / 73.9 / 83.4 & 88.9 / 56.8 / 85.8 & 21.2 / 1.8 / 26.3 & 26.6 / 18.5 / 26.6 & 60.6 / 47.8 / 60.3 \\
LLaVA-Video-7B & 46.2 / 32.1 / 43.8 & 21.9 / 19.3 / 19.5 & 1.0 / 1.3 / 1.8 & 30.8 / 19.1 / 31.0 & 56.9 / 46.1 / 59.6 \\
Kimi-VL-A3B & 37.5 / 33.6 / 38.3 & 39.9 / 29.9 / 40.9 & 5.6 / 2.5 / 6.3 & 25.2 / 14.6 / 24.5 & 48.1 / 37.0 / 48.8 \\
GPT-4.1 & 69.3 / 60.7 / 68.6 & 81.4 / 52.1 / 79.0 & 21.2 / 2.8 / 23.0 & 40.3 / 24.3 / 42.0 & 68.4 / 52.9 / 67.7 \\
o3 & 79.7 / 65.9 / 78.2 & 83.7 / 58.0 / 82.7 & 25.5 / 3.0 / 28.0 & 38.3 / 24.1 / 41.8 & 77.4 / 59.3 / 75.4 \\
\bottomrule
\end{tabular}
\end{table*}
\begin{table*}[!htbp]
\caption{\textbf{\concat: final-answer accuracy under greying.} Cells are intact / evidence grey / non-evidence grey (\%). Eligible n: 858 / 1,144 / 396 / 519 / 297. \sharded retains original intact-video replies; \replystripped removes them.}
\label{tab:greying_concat_clues}
\centering\scriptsize\setlength{\tabcolsep}{3pt}
\renewcommand{\arraystretch}{1.12}
\begin{tabular}{@{}lccccc@{}}
\toprule
Model & CLEVRER obj. & CLEVRER time & Charades ground & Charades obj. & Perception \\
\midrule
Qwen3-VL-8B & 71.1 / 53.6 / 71.4 & 68.6 / 42.7 / 71.6 & 24.0 / 0.3 / 32.6 & 37.0 / 21.2 / 37.0 & 61.6 / 46.8 / 63.6 \\
InternVL3.5-8B & 85.4 / 72.0 / 85.2 & 75.6 / 45.3 / 76.0 & 13.4 / 3.5 / 18.4 & 27.7 / 17.3 / 30.1 & 50.5 / 38.0 / 50.5 \\
InternVL3.5-14B & 82.8 / 68.1 / 81.7 & 88.5 / 57.0 / 85.6 & 21.7 / 0.8 / 26.5 & 24.5 / 16.4 / 26.2 & 57.2 / 45.1 / 57.9 \\
LLaVA-Video-7B & 44.9 / 31.6 / 43.1 & 20.1 / 19.1 / 19.3 & 0.3 / 1.0 / 0.5 & 26.8 / 16.4 / 26.8 & 52.2 / 45.1 / 55.9 \\
Kimi-VL-A3B & 27.9 / 25.3 / 31.1 & 35.1 / 30.9 / 37.2 & 3.5 / 2.3 / 5.1 & 15.8 / 9.8 / 18.3 & 36.7 / 32.3 / 39.7 \\
GPT-4.1 & 70.0 / 61.7 / 71.1 & 81.6 / 52.8 / 79.6 & 19.4 / 3.5 / 27.8 & 41.8 / 26.4 / 43.2 & 69.7 / 51.2 / 68.0 \\
o3 & 79.3 / 65.7 / 79.7 & 83.5 / 58.0 / 80.3 & 28.5 / 5.6 / 30.3 & 38.0 / 22.5 / 41.2 & 76.4 / 59.3 / 77.8 \\
\bottomrule
\end{tabular}
\end{table*}
\begin{table*}[!htbp]
\caption{\textbf{\sharded: final-answer accuracy under greying.} Cells are intact / evidence grey / non-evidence grey (\%). Eligible n: 858 / 1,144 / 396 / 519 / 297. \sharded retains original intact-video replies; \replystripped removes them.}
\label{tab:greying_sharded}
\centering\scriptsize\setlength{\tabcolsep}{3pt}
\renewcommand{\arraystretch}{1.12}
\begin{tabular}{@{}lccccc@{}}
\toprule
Model & CLEVRER obj. & CLEVRER time & Charades ground & Charades obj. & Perception \\
\midrule
Qwen3-VL-8B & 57.3 / 54.4 / 57.7 & 48.3 / 45.8 / 48.9 & 18.4 / 13.4 / 18.7 & 34.7 / 30.3 / 37.0 & 45.5 / 42.1 / 43.8 \\
InternVL3.5-8B & 56.4 / 50.5 / 56.6 & 29.4 / 24.8 / 28.1 & 9.3 / 7.1 / 9.6 & 19.8 / 15.2 / 20.8 & 34.7 / 30.0 / 36.0 \\
InternVL3.5-14B & 60.6 / 58.2 / 61.7 & 59.1 / 47.3 / 57.5 & 11.4 / 9.3 / 12.6 & 25.0 / 19.5 / 28.1 & 36.0 / 30.3 / 35.4 \\
LLaVA-Video-7B & 27.7 / 26.8 / 27.5 & 10.2 / 10.2 / 10.6 & 0.3 / 0.8 / 0.3 & 22.9 / 22.4 / 23.1 & 30.0 / 29.0 / 28.6 \\
Kimi-VL-A3B & 23.8 / 23.9 / 25.2 & 18.6 / 17.1 / 18.4 & 0.5 / 0.8 / 0.5 & 5.6 / 6.0 / 6.4 & 21.2 / 17.8 / 19.9 \\
GPT-4.1 & 53.6 / 50.1 / 54.2 & 71.9 / 66.0 / 72.2 & 13.4 / 11.9 / 15.2 & 39.5 / 25.4 / 40.1 & 53.5 / 45.1 / 57.2 \\
o3 & 76.9 / 69.0 / 77.2 & 84.6 / 60.5 / 82.3 & 15.9 / 5.3 / 20.5 & 35.8 / 22.5 / 38.5 & 70.4 / 57.2 / 74.7 \\
\bottomrule
\end{tabular}
\end{table*}
\begin{table*}[!htbp]
\caption{\textbf{\replystripped: final-answer accuracy under greying.} Cells are intact / evidence grey / non-evidence grey (\%). Eligible n: 858 / 1,144 / 396 / 519 / 297. \sharded retains original intact-video replies; \replystripped removes them.}
\label{tab:greying_strip}
\centering\scriptsize\setlength{\tabcolsep}{3pt}
\renewcommand{\arraystretch}{1.12}
\begin{tabular}{@{}lccccc@{}}
\toprule
Model & CLEVRER obj. & CLEVRER time & Charades ground & Charades obj. & Perception \\
\midrule
Qwen3-VL-8B & 69.7 / 54.2 / 71.1 & 68.2 / 42.7 / 71.9 & 23.7 / 1.3 / 33.6 & 34.7 / 20.2 / 37.2 & 56.2 / 44.1 / 58.2 \\
InternVL3.5-8B & 75.9 / 66.1 / 74.4 & 43.7 / 31.5 / 42.8 & 15.9 / 4.3 / 22.7 & 25.6 / 16.8 / 28.9 & 43.8 / 33.0 / 46.1 \\
InternVL3.5-14B & 82.2 / 70.6 / 80.0 & 70.7 / 46.9 / 67.2 & 18.2 / 1.8 / 24.5 & 29.7 / 17.9 / 30.1 & 58.9 / 42.8 / 60.3 \\
LLaVA-Video-7B & 40.0 / 26.9 / 36.9 & 18.8 / 18.0 / 17.4 & 3.0 / 2.0 / 2.3 & 28.9 / 17.7 / 28.3 & 55.6 / 48.1 / 55.6 \\
Kimi-VL-A3B & 23.0 / 20.3 / 23.4 & 16.8 / 13.8 / 16.7 & 3.5 / 1.3 / 2.5 & 14.5 / 9.6 / 18.3 & 32.3 / 29.0 / 32.7 \\
GPT-4.1 & 66.0 / 57.9 / 68.1 & 78.3 / 52.6 / 74.4 & 20.2 / 2.3 / 21.5 & 43.0 / 26.8 / 41.8 & 67.3 / 52.5 / 67.3 \\
o3 & 79.0 / 64.9 / 78.7 & 83.7 / 56.8 / 81.4 & 26.0 / 3.8 / 27.8 & 37.8 / 24.7 / 40.5 & 79.8 / 57.6 / 74.7 \\
\bottomrule
\end{tabular}
\end{table*}

%% file: tables/supp_repair_controls.tex
\begin{table*}[!htbp]
\caption{\textbf{\cleanturn component contrasts and negative controls.} Paired accuracy differences, points [pointwise video-bootstrap 95\% CI]. Steering-on-\full transfers the conversation-derived steering difference onto the clean \full prompt rather than recalibrating on \full.}
\label{tab:supp_repair_controls}
\centering\scriptsize\setlength{\tabcolsep}{2.4pt}
\renewcommand{\arraystretch}{1.12}
\begin{tabular}{@{}lccccc@{}}
\toprule
Model & \shortstack{\cleanturn $-$\\\sharded} & Steering over Mask & \cleanturn $-$ \full & Matched $-$ other (Mask) & Steering on \full{} $-$ \full \\
\midrule
LLaVA-Video-7B & +10.3 [8.5, 12.2] & +8.2 [6.5, 10.0] & +1.4 [0.1, 2.7] & +6.0 [4.5, 7.5] & +1.8 [0.2, 3.3] \\
Qwen3-VL-8B & +7.1 [5.5, 8.6] & +1.9 [0.6, 3.2] & $-0.6$ [$-1.6$, 0.4] & +1.8 [0.7, 3.0] & $-2.0$ [$-3.5$, $-0.5$] \\
InternVL3.5-14B & +12.4 [10.5, 14.4] & $-0.2$ [$-1.1$, 0.7] & $-4.4$ [$-5.8$, $-3.0$] & +0.9 [$-0.1$, 1.8] & $-0.8$ [$-2.4$, 0.7] \\
InternVL3.5-8B & +9.6 [7.6, 11.5] & +0.6 [$-0.4$, 1.5] & $-6.0$ [$-7.6$, $-4.5$] & +0.5 [$-0.3$, 1.3] & $-10.3$ [$-12.1$, $-8.5$] \\
\bottomrule
\end{tabular}
\end{table*}

%% file: tables/supp_repair_tasks_grid.tex
\begin{table*}[!htbp]
\centering
\begin{minipage}[t]{0.488\textwidth}
\vspace{0pt}
\caption{\textbf{LLaVA-Video-7B: repair accuracy (\%).} All 3,218 evaluation cases.}
\label{tab:repair_tasks_llava-video-7b}
\centering\scriptsize\setlength{\tabcolsep}{2.0pt}
\renewcommand{\arraystretch}{1.12}
\resizebox{\linewidth}{!}{\begin{tabular}{@{}lcccccc@{}}
\toprule
Condition & C. obj. & C. time & Ch. ground & Ch. obj. & P. obj. & Mean \\
\midrule
\full & 46.0 & 21.9 & 1.0 & 31.0 & 56.7 & 31.3 \\
\sharded & 33.8 & 17.2 & 0.8 & 22.6 & 37.6 & 22.4 \\
Recap & 38.9 & 19.2 & 0.5 & 27.2 & 43.0 & 25.8 \\
Clean-context mix & 39.0 & 18.7 & 1.3 & 28.2 & 47.3 & 26.9 \\
Reply Mask only & 33.7 & 14.9 & 3.3 & 25.3 & 45.6 & 24.6 \\
Steering only & 45.1 & 24.0 & 2.8 & 30.5 & 53.7 & 31.2 \\
\cleanturn & 42.5 & 29.3 & 1.3 & 33.5 & 57.0 & 32.7 \\
Other-case steering & 30.5 & 19.5 & 1.5 & 22.4 & 32.2 & 21.2 \\
Mask + other-case steering & 34.0 & 24.8 & 1.0 & 27.4 & 46.3 & 26.7 \\
Steering on \full & 46.5 & 29.0 & 3.0 & 31.0 & 56.0 & 33.1 \\
\bottomrule
\end{tabular}}
\end{minipage}\hfill
\begin{minipage}[t]{0.488\textwidth}
\vspace{0pt}
\caption{\textbf{Qwen3-VL-8B: repair accuracy (\%).} All 3,218 evaluation cases.}
\label{tab:repair_tasks_qwen3-vl-8b}
\centering\scriptsize\setlength{\tabcolsep}{2.0pt}
\renewcommand{\arraystretch}{1.12}
\resizebox{\linewidth}{!}{\begin{tabular}{@{}lcccccc@{}}
\toprule
Condition & C. obj. & C. time & Ch. ground & Ch. obj. & P. obj. & Mean \\
\midrule
\full & 74.4 & 71.1 & 20.2 & 36.0 & 59.1 & 52.1 \\
\sharded & 58.5 & 54.4 & 21.0 & 35.6 & 53.0 & 44.5 \\
Recap & 57.3 & 59.5 & 23.5 & 34.9 & 55.7 & 46.2 \\
Clean-context mix & 63.5 & 59.1 & 22.7 & 36.0 & 54.7 & 47.2 \\
Reply Mask only & 68.9 & 63.9 & 20.5 & 36.2 & 59.1 & 49.7 \\
Steering only & 69.7 & 66.7 & 16.2 & 35.2 & 55.4 & 48.6 \\
\cleanturn & 72.8 & 69.2 & 19.4 & 35.6 & 60.7 & 51.6 \\
Other-case steering & 52.7 & 48.1 & 15.7 & 32.0 & 48.7 & 39.4 \\
Mask + other-case steering & 70.0 & 62.2 & 22.0 & 35.2 & 59.4 & 49.8 \\
Steering on \full & 71.7 & 64.2 & 19.4 & 36.4 & 59.1 & 50.1 \\
\bottomrule
\end{tabular}}
\end{minipage}
\end{table*}
\begin{table*}[!htbp]
\centering
\begin{minipage}[t]{0.488\textwidth}
\vspace{0pt}
\caption{\textbf{InternVL3.5-14B: repair accuracy (\%).} All 3,218 evaluation cases.}
\label{tab:repair_tasks_internvl3.5-14b}
\centering\scriptsize\setlength{\tabcolsep}{2.0pt}
\renewcommand{\arraystretch}{1.12}
\resizebox{\linewidth}{!}{\begin{tabular}{@{}lcccccc@{}}
\toprule
Condition & C. obj. & C. time & Ch. ground & Ch. obj. & P. obj. & Mean \\
\midrule
\full & 85.8 & 88.8 & 20.7 & 28.2 & 61.4 & 57.0 \\
\sharded & 63.5 & 63.5 & 15.7 & 23.0 & 35.2 & 40.2 \\
Recap & 72.5 & 85.0 & 16.2 & 25.3 & 49.0 & 49.6 \\
Clean-context mix & 82.8 & 83.0 & 19.4 & 27.2 & 56.4 & 53.7 \\
Reply Mask only & 81.0 & 78.9 & 16.7 & 30.1 & 57.4 & 52.8 \\
Steering only & 76.2 & 74.6 & 16.2 & 27.2 & 46.6 & 48.2 \\
\cleanturn & 80.4 & 80.3 & 16.2 & 30.1 & 56.0 & 52.6 \\
Other-case steering & 65.4 & 62.8 & 15.4 & 24.3 & 36.6 & 40.9 \\
Mask + other-case steering & 78.7 & 78.8 & 15.2 & 28.9 & 57.0 & 51.7 \\
Steering on \full & 85.1 & 88.5 & 21.5 & 27.0 & 58.7 & 56.2 \\
\bottomrule
\end{tabular}}
\end{minipage}\hfill
\begin{minipage}[t]{0.488\textwidth}
\vspace{0pt}
\caption{\textbf{InternVL3.5-8B: repair accuracy (\%).} All 3,218 evaluation cases.}
\label{tab:repair_tasks_internvl3.5-8b}
\centering\scriptsize\setlength{\tabcolsep}{2.0pt}
\renewcommand{\arraystretch}{1.12}
\resizebox{\linewidth}{!}{\begin{tabular}{@{}lcccccc@{}}
\toprule
Condition & C. obj. & C. time & Ch. ground & Ch. obj. & P. obj. & Mean \\
\midrule
\full & 86.1 & 73.5 & 14.4 & 25.1 & 52.0 & 50.2 \\
\sharded & 64.0 & 39.0 & 11.1 & 24.1 & 34.9 & 34.6 \\
Recap & 78.9 & 65.4 & 15.4 & 28.4 & 52.3 & 48.1 \\
Clean-context mix & 80.2 & 66.3 & 12.9 & 24.5 & 46.3 & 46.0 \\
Reply Mask only & 73.2 & 58.2 & 14.4 & 24.3 & 48.0 & 43.6 \\
Steering only & 62.9 & 39.6 & 10.9 & 24.3 & 38.6 & 35.3 \\
\cleanturn & 76.1 & 57.3 & 14.6 & 23.9 & 49.0 & 44.2 \\
Other-case steering & 63.4 & 38.4 & 11.4 & 24.1 & 39.3 & 35.3 \\
Mask + other-case steering & 73.9 & 58.0 & 15.2 & 23.6 & 47.7 & 43.7 \\
Steering on \full & 76.3 & 45.5 & 6.1 & 23.9 & 47.7 & 39.9 \\
\bottomrule
\end{tabular}}
\end{minipage}
\end{table*}